\documentclass{article}

\PassOptionsToPackage{numbers, compress}{natbib}
 \usepackage[preprint]{neurips_2026}

\usepackage[utf8]{inputenc} 
\usepackage[T1]{fontenc}    
\usepackage{url}            
\usepackage{booktabs}       
\usepackage{amsfonts}       
\usepackage{nicefrac}       
\usepackage{microtype}      
\usepackage{xcolor}         

\usepackage{graphicx}       
\usepackage{subcaption}     
\usepackage{bbding}         
\usepackage{colortbl}       
\usepackage{caption}        
\usepackage{multirow}       
\usepackage{array}          
\usepackage{tabularx}       
\usepackage{wrapfig}        
\usepackage{enumitem}       
\usepackage{makecell}       
\usepackage{amsmath}        

\newcommand{\eg}{\emph{e.g.}}

\newcommand{\etc}{\emph{etc.}}

\usepackage{hyperref}       
\usepackage{titletoc}
\usepackage{tocloft}

\title{\includegraphics[height=1em]{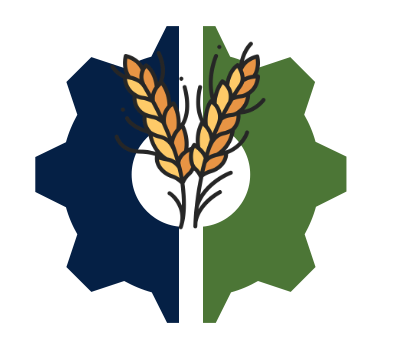}AgroCoT: A Chain-of-Thought Benchmark \\for Evaluating Reasoning in Vision-Language Models \\for Agriculture}

\author{
\normalfont
Yibin Wen$^{1,*}$,
Qingmei Li$^{2,*}$,
Zi Ye$^{1,*}$,
Jiarui Zhang$^{1,*}$,\\
Xiaoya Fan$^{3}$,
Zurong Mai$^{1}$,
Jing Wu$^{1}$,
Shuohong Lou$^{1}$,
Yuhang Chen$^{1}$,\\
Henglian Huang$^{1}$,
Yang Zhang$^{1}$,
Defeng Gu$^{1}$,
Lingyuan Zhao$^{4}$,\\
Yutong Lu$^{1,7}$,
Haohuan Fu$^{2,7}$,
Jianxi Huang$^{5,6}$,
Juepeng Zheng$^{1,7,\dagger}$
\\[2mm]
$^{1}$Sun Yat-sen University \quad
$^{2}$Tsinghua University
\\
$^{3}$Southwest University \quad
$^{4}$HuanTian Wisdom Technology Co., Ltd. \\
$^{5}$China Agricultural University \quad
$^{6}$Southwest Jiaotong University \\
$^{7}$National Supercomputing Center in Shenzhen 
\\[2mm]
$^{*}$Equal contribution \quad
$^{\dagger}$Corresponding author
}

\begin{document}

\maketitle
\begin{abstract}
Recent advancements in Vision-Language Models (VLMs) have significantly impacted various industries. In agriculture, these multimodal capabilities hold great promise for applications such as precision farming, crop monitoring, pest detection, and environmental sustainability. However, while several Visual Question Answering (VQA) datasets and benchmarks have been developed to assess VLM performance, they often fail to effectively evaluate the critical reasoning and problem-solving skills needed in complex agricultural contexts. To address this gap, we introduce AgroCoT, a VQA dataset that integrates Chain-of-Thought (CoT) reasoning, specifically designed to evaluate the reasoning capabilities of VLMs. With 4,759 carefully curated samples, AgroCoT provides a comprehensive and robust evaluation of reasoning abilities, particularly in zero-shot scenarios, focusing on the models' ability to engage in logical reasoning and effective problem-solving. Our evaluation of 30 representative VLMs, including both proprietary and open-source models, reveals a gap in their reasoning capabilities, which underscores the importance of incorporating CoT for assessments. Our dataset is available at  \url{https://huggingface.co/datasets/AgroCoT/AgroCoT}.
\end{abstract}
\section{Introduction}
\label{sec:intro}
Agriculture, the cornerstone of global food security and economic stability, underpins human livelihoods and plays a central role in achieving Sustainable Development Goal 2 (Zero Hunger)~\cite{shaikh_role_2025}. However, it now faces unprecedented challenges from climate extremes and land degradation. Recent advancements in Artificial Intelligence, particularly Large Language Models (LLMs) and Vision-Language Models (VLMs), offer promising solutions to agricultural challenges, including crop disease diagnosis~\cite{zhou2024few, zhang2024visual, cao2023cucumber}, pest management~\cite{zhao2024implementation, zhang2023multimodal, zhang2025few}, and precision farming~\cite{11136213, wu2025farmseg_vlm}. Although existing VLMs (such as GPT-5~\cite{leon2025gpt}, Qwen2.5-VL~\cite{bai2025qwen2}, etc.) demonstrate exceptional generalization across diverse visual and textual tasks and show significant promise in comprehensive agricultural applications, the effectiveness of these VLMs in agricultural scenarios requires careful evaluation to help researchers select the most appropriate models for purposes.

\begin{figure}[htbp]
    \centering
    \includegraphics[width=\linewidth]{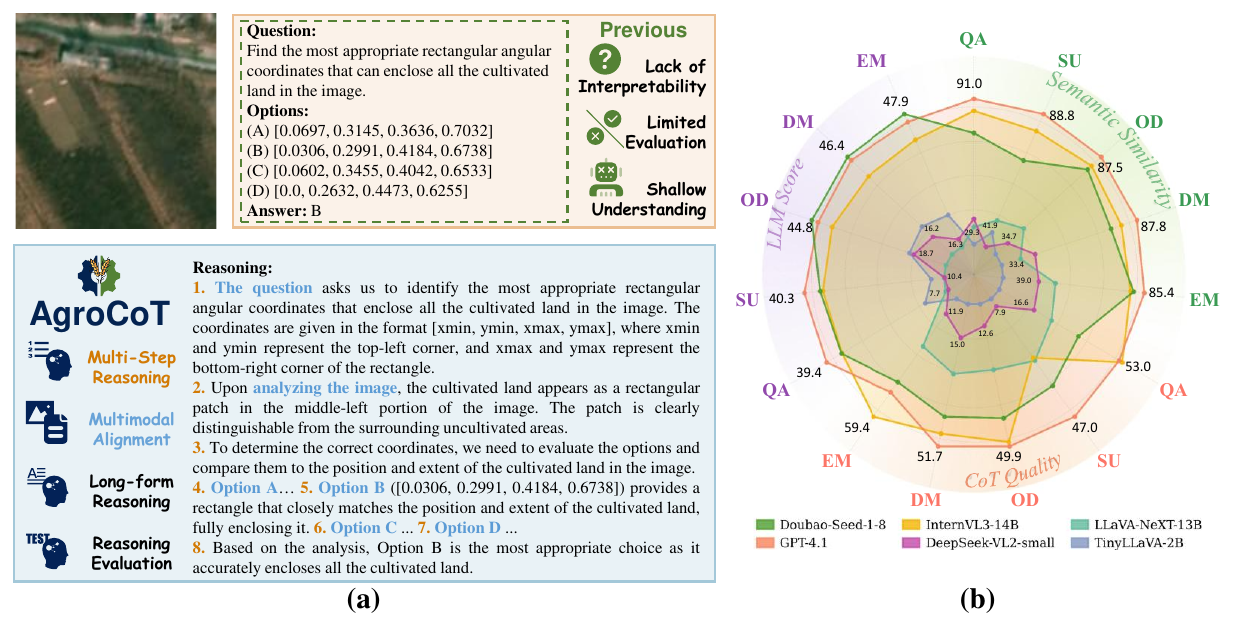}
    \caption{Benchmark comparison and model performance. (a) illustrates the advantages of AgroCoT over existing benchmarks, and (b) presents VLM performance across evaluation dimensions.}
    \label{fig:innovation}
\end{figure}

    

Recently, several multimodal benchmarks have been developed for agricultural applications. These benchmarks~\cite{gaubaagmmu, dongre2025mirage, shinoda2025agrobench} incorporate images from the Internet or existing agriculture dataset, and texts from user-expert interactions or synthetic data. For example, CDDM~\cite{liu2024multimodal} creates a benchmark for crop disease diagnosis and AgroMind~\cite{li2025can} focuses on agricultural remote sensing scenarios. To provide a straightforward assessment of how well VLMs can answer agricultural questions, existing benchmarks typically adopt various question-answering formats only with accuracy-based metrics. 

However, existing agricultural benchmarks only offer a limited assessment of capability of VLM (see Fig.~\ref{fig:innovation} (a)), resulting in poor interpretability, limited evaluation, and shallow understanding. This motivates a more comprehensive benchmark that evaluates internal abilities to analyze problems, understand scenarios, and perform multi-step reasoning. By breaking down complex multimodal tasks into clear steps, Chain-of-Thought (CoT) reasoning~\cite{wei2022chain, zhang2025enhancing, gao2025benchmarking, hao2025can} has been shown to improve the interpretability, depth of reasoning, and generalization capabilities of VLMs. At the same time, it mitigates the black-box nature of these models~\cite{lu2022learn}, reduces the likelihood of hallucinations, and aligns the reasoning process with expert decision-making workflows. 

Considering the limitations of the aforementioned benchmarks in agricultural scenarios, we propose a new benchmark for multi-step multimodal CoT in agriculture, called AgroCoT, which is the first comprehensive VQA benchmark for agriculture with explicit CoT reasoning. As shown in Fig.~\ref{fig:innovation}(a), AgroCoT introduces four key advantages: multi-step reasoning, multimodal alignment, long-form reasoning and reasoning evaluation, which is designed to deeply assess the problem-solving and reasoning capabilities of existing representative VLMs. 
Our evaluation results indicate that existing VLMs may fall short in their reasoning capabilities (see Fig.~\ref{fig:innovation}(b)), highlighting their limitations towards agricultural understanding and decision-making with multi-step and long-form reasoning. 

Overall, the primary contributions of this work can be summarized as follows:
\begin{itemize}[leftmargin=*]
    \item  We propose AgroCoT, a comprehensive agricultural multi-step, multimodal CoT dataset comprising $4,759$ VQA pairs. Each pair includes human-refined CoT reasoning, which allows for a comprehensive evaluation of understanding abilities of model across various agricultural contexts.
    \item As a problem-oriented benchmark, AgroCoT addresses five key dimensions (object detection, quantitative analysis, disease monitoring, spatial understanding, and environmental management) designed to tackle critical agricultural challenges with a total of 15 task types. 
    \item We conduct a thorough evaluation of 25 open-source and 5 proprietary VLMs, focusing on their accuracy in answering questions as well as their reasoning abilities. Additionally, we analyze how the performance of models varied across tasks with different levels of CoT reasoning steps.
\end{itemize}
\section{Related Work}
\label{sec:related_work}

\subsection{Large Vision-Language Models and Chain-of-Thought}

Vision-Language Models (VLMs) couple visual encoders with large language models to enable multimodal understanding~\cite{li2025benchmark}. CLIP~\cite{pmlr-v139-radford21a} established scalable vision-language representations, while models such as GPT-4o~\cite{achiam2023gpt}, Gemini~\cite{team2023gemini}, and DeepSeek~\cite{liu2024deepseek} further advance multimodal reasoning. As multimodal tasks increasingly require models to move beyond recognition toward evidence-grounded reasoning, Chain-of-Thought (CoT) prompting has become an important mechanism for decomposing complex questions into intermediate steps~\cite{wei2022chain,kojima2022large,wang2025multimodal}. 
In multimodal tasks, rationales must align with visual evidence~\cite{singh2023assessing}; representative efforts include ScienceQA~\cite{lu2022learn}, scene-graph-based compositional CoT~\cite{mitra2024compositional}, and region-guided interleaved-modal CoT~\cite{gao2025interleaved}.

Adapting VLMs to agriculture has shown promise through AgroGPT~\cite{awais2025agrogpt}, Agri-LLaVA~\cite{wang2024agri}, and some vision-language pipelines for crop health monitoring and cropland management~\cite{zhang2024visual,wu2025farmseg_vlm}. However, these efforts are often trained and evaluated on crop- or symptom-specific datasets with limited scenes and sensors~\cite{liu2024multimodal,zhang2024visual}. Meanwhile, existing multimodal CoT benchmarks~\cite{schwenk2022okvqa,lu2022learn,lu2023mathvista,yue2024mmmu,chen2024m3cot,jiang2025mme} either target general domains or lack detailed agricultural reasoning annotations. As a result, current evaluations rarely assess practical agricultural skills such as counting, planting recommendation, environmental management, or evidence-grounded intermediate reasoning. This motivates a broader agricultural multimodal benchmark with explicit, structured CoT, enabling faithful evaluation of both task performance and the reasoning process behind model decisions in real-world field conditions.

\begin{table}[t]
    \centering
    \caption{Comparison of AgroCoT with existing agriculture benchmarks. Abbreviations adopted: MCQ for Multiple Choice Questions; TFQ for True-or-False Questions; OEQ for Open-Ended Questions; LFQ for Long-Form Question-answering; Conv for conversational multi-turn interactions.}
    \label{tab:benchmark_comparison}
    \small
        \begin{tabular}{cccccc}
            \Xhline{1.2pt}
            \textbf{Dataset} & \textbf{Size} & \textbf{Answer Type} & \textbf{Multimodal} & \textbf{Multi-view} & \textbf{CoT} \\
            \Xhline{1.2pt}
            CROP~\cite{zhang2024empowering} & 5,045 & MCQ & \XSolidBrush & \XSolidBrush & \XSolidBrush \\
            CDDM~\cite{liu2024multimodal} & 1 million & TFQ, OEQ & \Checkmark & \XSolidBrush & \XSolidBrush \\
            AgroInstruct~\cite{awais2025agrogpt} & 70K & TFQ, OEQ & \Checkmark & \XSolidBrush & \XSolidBrush \\
            MIRAGE~\cite{dongre2025mirage} & 41K & LFQ, Conv & \Checkmark & \XSolidBrush & \XSolidBrush \\
            AGMMU~\cite{gaubaagmmu} & 1,492 & MCQ, OEQ & \Checkmark & \XSolidBrush & \XSolidBrush \\
            AgroBench~\cite{shinoda2025agrobench} & 4,342 & MCQ & \Checkmark & \XSolidBrush & \XSolidBrush \\
            AgriBench-VL-4K~\cite{yang2025agrigpt} & 3,876 & MCQ, OEQ & \Checkmark & \XSolidBrush & \XSolidBrush \\
            AgroMind~\cite{li2025can} & 28,482 & MCQ, TFQ, OEQ & \Checkmark & \Checkmark & \XSolidBrush \\
            \hline
            \rowcolor{green!20}
            AgroCoT (Ours) & 4,759 & MCQ, TFQ, OEQ, LFQ & \Checkmark & \Checkmark & \Checkmark \\
            \Xhline{1.2pt}
        \end{tabular}
\end{table}

\subsection{Agriculture Benchmark for VLMs}
Table~\ref{tab:benchmark_comparison} summarizes the variety of existing benchmarks developed in recent years to evaluate VLM capabilities in agricultural domain. Some datasets such as AGMMU~\cite{gaubaagmmu} and AgroBench~\cite{shinoda2025agrobench} focus on general VQA to assess the comprehension of agricultural scenes. More recently, AgroMind~\cite{li2025can} introduces a comprehensive benchmark with multi-sensor imagery to evaluate models on complex tasks like spatial perception and scene reasoning. AgriBench-VL-4K~\cite{yang2025agrigpt} is a comprehensive and challenging benchmark, featuring multi-metric evaluation and a dual evaluation framework. Complementing these general evaluations, other datasets address more specific applications. For instance, AgroInstruct~\cite{awais2025agrogpt} facilitates expert-level instruction tuning, MIRAGE~\cite{dongre2025mirage} explores multi-turn conversational interactions like expert consultations, and CDDM~\cite{liu2024multimodal} provides a large-scale multimodal dataset for crop disease diagnosis. In contrast, CROP~\cite{zhang2024empowering} have focused on enhancing language models for crop science but are text-based and do not incorporate visual input.

Despite these advancements, existing agricultural benchmarks primarily evaluate models based on final answer accuracy with limited question types (see Table~\ref{tab:benchmark_comparison}). None of these benchmarks incorporate explicit CoT reasoning. To address these gaps, we present AgroCoT, a multimodal, multi-view benchmark that encompasses diverse agricultural scenarios. Each item includes a human-refined, multi-step CoT, enabling joint evaluation of answer correctness and reasoning quality.

\section{AgroCoT}
\label{sec:AgroCoT}
AgroCoT is a multimodal benchmark designed for agriculture. It is a VQA dataset enriched with CoT reasoning and includes 4,759 VQA pairs, whose scale is consistent with modern multimodal benchmarks (\eg, MME-CoT~\cite{jiang2025mme}: 1,130 samples; CoMT~\cite{cheng2025comt}: 3,853 samples). As illustrated in Fig.~\ref{fig:innovation} (b), AgroCoT allows for the assessment of reasoning abilities by evaluating not only the accuracy of the final answers but also the depth and coherence of the reasoning processes involved, ensuring a thorough analysis of their agricultural reasoning capabilities.

\subsection{Problem-Oriented}
\label{subsec:problem}
\begin{figure*}[t]
    \centering
    \includegraphics[width=\textwidth]{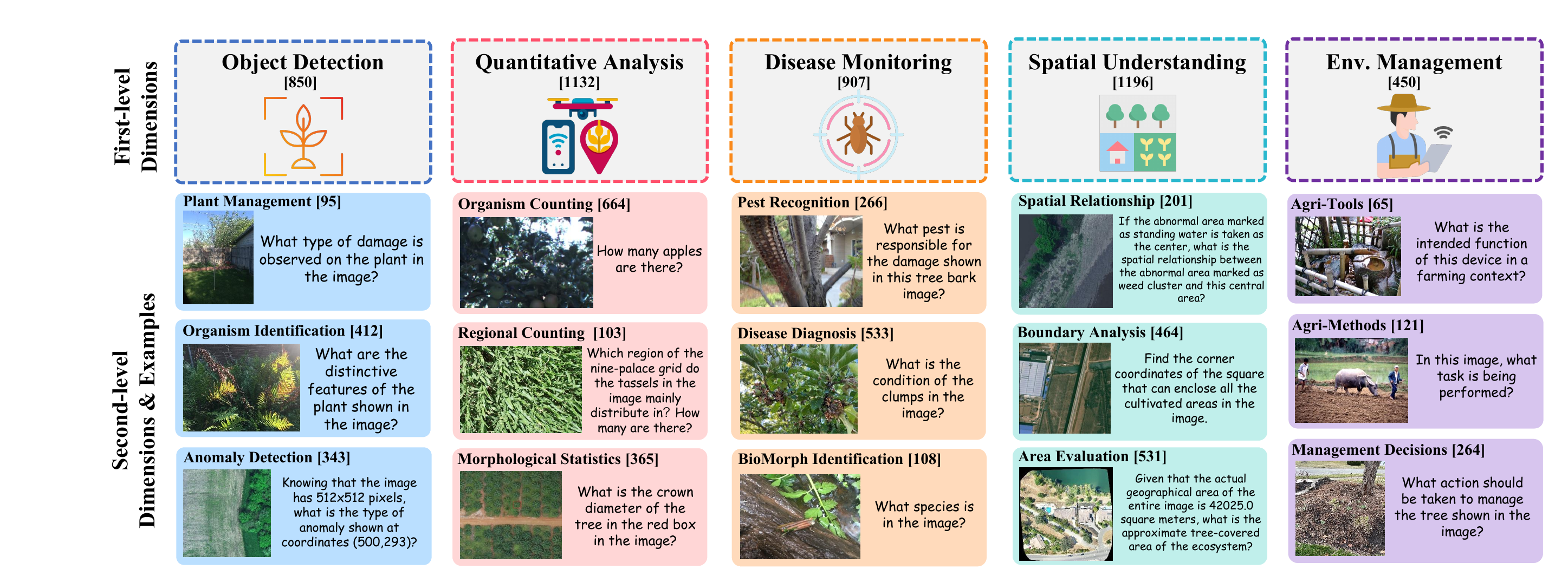}
    \caption{Hierarchical task system of AgroCoT. Based on the progressive cognitive pipeline in agricultural intelligence, AgroCoT constructs five evaluation dimensions (such as object detection, quantitative analysis, disease monitoring, spatial understanding and environmental management), covering 15 different and diverse task types. Numbers in brackets denote the QA pair count per task.}
    \vspace{-2em}
    \label{fig:overview}
\end{figure*}

As a problem-driven multimodal benchmark specifically designed for agricultural applications (see Fig.~\ref{fig:overview}), AgroCoT is organized around five key real-world problem types: Object Detection (\textbf{OD}), Quantitative Analysis (\textbf{QA}), Disease Monitoring (\textbf{DM}), Spatial Understanding (\textbf{SU}), and Environmental Management (\textbf{EM}), commencing with foundational identification, advancing through quantitative assessment and diagnostic scrutiny, and culminating in sophisticated spatial reasoning and decision-making support. These five areas form a systematic hierarchy of agricultural cognition, progressing from fundamental visual perception (OD, SU) to domain-specific analytical reasoning (QA, DM) and ultimately to high-level decision-making (EM).

{\bf Object Detection} forms the foundation of agricultural visual understanding by identifying and localizing key entities and anomalies in field images. Key tasks at this level include Plant Management, covering plant management by providing control and optimization strategies; Organism Identification, focusing on identifying organisms such as crops, plants, leaves, fruits, \etc, addressing core questions like ``what is it?''; and Anomaly Detection, detecting anomalies in planting and environmental issues (\eg, waterlogging), and analyzing spatial relationships between anomalies.

{\bf Quantitative Analysis} targets precise measurement and enumeration to support data-driven agronomic decisions. This includes Organism Counting, addressing counting organisms like insects and individual crops to obtain population data; Regional Counting, targeting in specific areas such as nine-grid divisions or corners; Morphological Statistics, measuring parameters (\eg, crown diameter), counting fruit sizes, and quantifying spatial units in parcels. 

{\bf Disease Monitoring} goes beyond detection to diagnose health status and recommend interventions based on visual symptoms. Typical tasks include Pest Recognition, which involves identifying pests and assessing their impact on crops or other organisms; BioMorph Identification, which focuses on identifying insects or plants, emphasizing species or traits without diagnosing diseases; and Disease Diagnosis, which centers on diagnosing crop health (\eg leaves), identifying disease types with diseases, symptoms, and management solutions.

{\bf Spatial Understanding} represents the geometric reasoning layer, enabling boundary definition, area quantification, and spatial relationship. Tasks in this dimension include Boundary Analysis, which focuses on analyzing the boundaries of targets such as farmland, including coordinate matching and bounding box solutions; Area Evaluation, which encompasses area estimates, ratios, coverage assessments, and level descriptions; and Spatial Relationship, which involves tasks such as analyzing spatial relationships between anomalous regions and developing resource optimization strategies based on weed or crop coverage.

{\bf Environmental Management} integrates contextual knowledge of tools, practices, and terrain to guide adaptive farm operations. This includes Agri-Tools, identifying agricultural tools, equipment, or structures, clarifying their names, functions, and attributes; Agri-Methods, revolving around operational processes, traditional practices, and addressing types; Management Decisions involving determining environmental attributes (\eg, climate zones or terrain) and formulating management decisions for crops, weeds, or trees. 

\subsection{CoT Reasoning} 
\label{subsec:cot_reason}

Compared to traditional VQA benchmarks in the agriculture, AgroCoT introduces a significant innovation of the CoT reasoning process. For each VQA sample (see Fig.~\ref{fig:innovation} (a)), the reasoning steps follow a structured sequence: (1) understanding the question, (2) describing the image, (3) retrieving relevant knowledge to connect the question and image, (4) logically reasoning toward an answer, and (5) providing the final response. The length and complexity of the reasoning chain vary depending on the task. For instance, in text-based multiple-choice questions, the model explains each option, while for image-based questions, the reasoning process involves a detailed description of each image. This flexibility ensures the reasoning chain is comprehensive and adaptable to different question types. 

The incorporation of CoT reasoning addresses a key limitation of traditional agricultural VQA datasets, which typically focus on end-to-end evaluations without explicitly considering the reasoning process. Such datasets primarily assess the ability to retrieve relevant knowledge rather than evaluate the depth of their reasoning. In contrast, AgroCoT advances the evaluation process by providing deeper insights into the reasoning capability, showcasing how well it understands and interprets agricultural scenarios. This evaluation highlights specific weaknesses in the reasoning process, allowing researchers to enhance VLMs for more accurate and reliable applications in agriculture.

\section{Dataset}
\label{sec:dataset}
\subsection{Data Construction}
\label{subsec:construction}
\begin{figure*}[t]
    \centering
    \includegraphics[width=\textwidth]{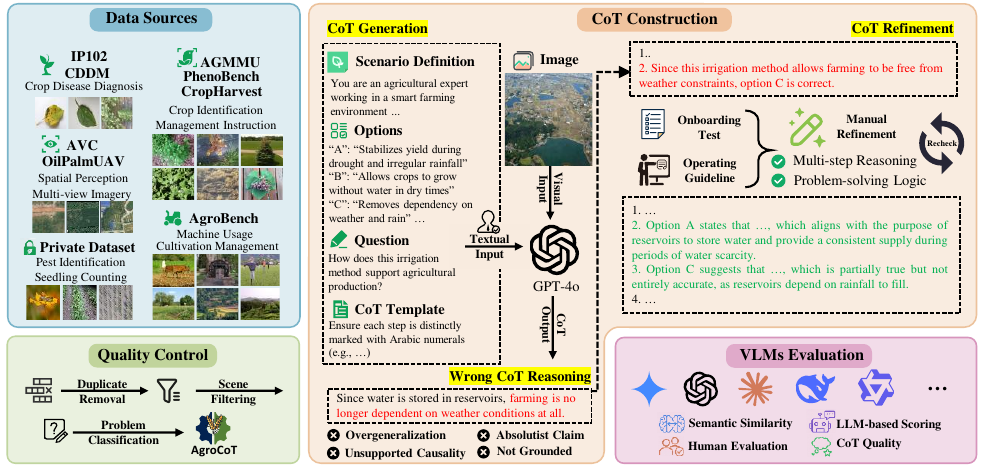}
    \vspace{-1.5em}
    \caption{The construction of AgroCoT benchmark with four steps: collecting samples from data sources, ensuring the quality of the samples, generating a CoT for each QA pair, and conducting a comprehensive evaluation of the representative VLMs.}
    \vspace{-2em}
    \label{fig:curaton}
\end{figure*}

{\bf Data Sources.} 
To address the persistent challenges in agricultural scenarios, we construct \textbf{AgroCoT}, a comprehensive benchmark for evaluating agricultural multimodal reasoning in VLMs. As shown in Fig.~\ref{fig:curaton}, AgroCoT integrates diverse public datasets, including OilPalmUAV~\cite{zheng2021growing}, PhenoBench~\cite{weyler2024phenobench}, AVC~\cite{chiu2020agriculture}, IP102~\cite{wu2019ip102}, CropHarvest~\cite{tseng2021cropharvest}, CDDM~\cite{liu2024multimodal}, AGMMU~\cite{gaubaagmmu}, and AgroBench~\cite{shinoda2025agrobench}, together with a proprietary VQA dataset developed in this work. These sources provide complementary coverage of UAV-based monitoring, fine-grained field understanding, aerial and remote-sensing crop analysis, pest and disease diagnosis, knowledge-intensive reasoning, and expert-oriented vision-language evaluation. By leveraging these heterogeneous resources, AgroCoT supports diverse tasks such as crop recognition, pest and disease diagnosis, field-scene understanding, remote-sensing monitoring, and agricultural decision support. More details are provided in Appendix~\ref{app:data_collection}.

In addition, our proprietary dataset supplements fine-grained tasks such as pest identification and seedling counting, which remain underrepresented in existing benchmarks. Collectively, these datasets expand scenario coverage and task diversity, providing a comprehensive evaluation platform for assessing model generalization under real-world agricultural production conditions.

{\bf Quality Control.} 
To ensure both diversity and relevance, we remove duplicates and apply scenario filtering to prioritize more complex reasoning tasks, while retaining a subset of simpler knowledge-based questions to maintain balance. We further structure the collected dataset around the practical demands of current agricultural tasks, thereby constructing a problem-driven benchmark tailored for rigorous evaluation of VLMs in agricultural applications.

{\bf CoT Generation.} 
Building upon the curated VQA dataset, we adopt a template-guided CoT construction pipeline. Specifically, agricultural experts first design task-specific reasoning templates that define the expected reasoning structure, key agronomic evidence, and response format for different agricultural scenarios. GPT-4o~\cite{hurst2024gpt} is then used to generate an initial CoT for each sample based on the image, question, scenario definition, and expert-designed CoT template. This process provides a scalable way to produce structured reasoning annotations while ensuring that the generated rationales follow agriculturally meaningful reasoning patterns. The initial CoTs are not directly used as final annotations; instead, they are further passed to the subsequent verification and quality-control pipeline. Appendix~\ref{app:prompt_design} presents the prompts and templates used for guidance. 

{\bf Wrong CoT Reasoning.} 
However, due to the limited agricultural knowledge and occasional misinterpretation of the format, some erroneous outputs were generated. We identify CoTs with incorrect answers, those with correct answers but insufficient text length, as well as CoTs exhibiting issues such as inadequate depth of reasoning, format inconsistencies, missing image elements, and logical errors. These problematic CoTs are then subjected to a thorough manual review process for further refinement. More details on the filtering of wrong CoTs can be found in Appendix~\ref{app:wrong_cot_filtering}.

{\bf CoT Refinement.} 
To ensure the quality of the CoTs, we select reviewers with a strong foundation in agricultural knowledge, complemented by excellent logical reasoning and textual expression skills, following a rigorous onboarding test. To ensure consistency and quality, reviewers refine the CoTs based on a standardized set of operating guidelines that outline explicit criteria for issue detection and required modifications. Through this structured refinement process, each CoT is thoroughly reviewed and enhanced to meet the standards of multi-step, logical problem-solving, ensuring alignment with the expectations of the agricultural benchmark. Over 70\% of the initial CoTs were manually revised, ensuring that the final annotations are not simple GPT-4o outputs and providing a more reliable basis for evaluating diverse multimodal models. More details can be found in Appendix~\ref{app:human_refinement}.

{\bf VLMs Evaluation.} Finally, we conduct a comprehensive evaluation on AgroCoT using a diverse set of representative VLMs, covering both proprietary and open-source models, such as GPT~\cite{achiam2023gpt}, Gemini~\cite{team2023gemini}, 
DeepSeek-VL~\cite{lu2024deepseek}, \etc. To assess their performance, we employ various evaluation metrics. Besides traditional answer-level accuracy metrics, we adopt a multi-dimensional evaluation framework that includes text semantic similarity measures, LLM-based scoring, human evaluation, and CoT quality assessment to systematically examine reasoning depth and semantic alignment.

\subsection{Statistics}
\label{subsec:statis}
\begin{figure*}[t]
    \centering
    \includegraphics[width=\textwidth]{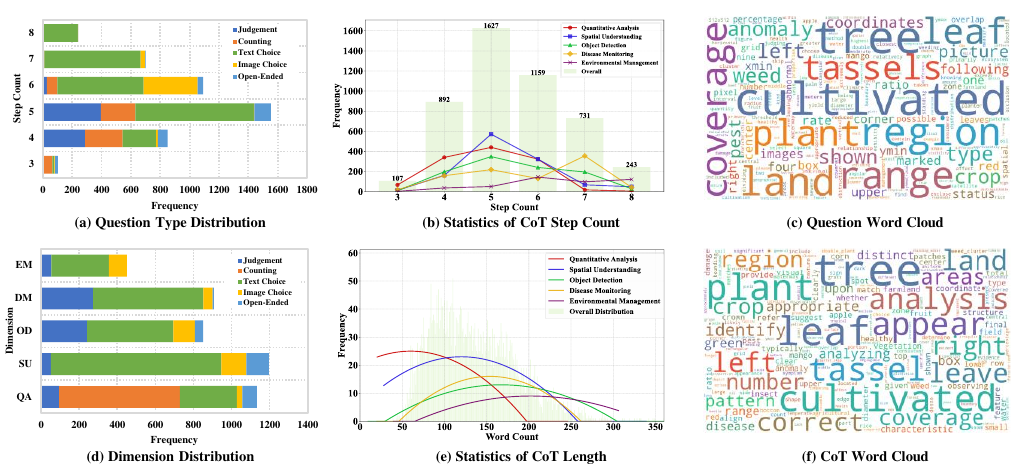}
    \vspace{-1.5em}
    \caption{Statistics of AgroCoT, from three perspectives: the distribution of question types across steps and dimensions, the number of steps and length distribution of CoTs, and word cloud analysis of both questions and CoTs.}
    \vspace{-1.5em}
    \label{fig:statistics}
\end{figure*}
{\bf Distribution of Question Types.}
AgroCoT covers diverse formats, including judgment, counting, text/image choice, and open-ended questions. As shown in Fig.~\ref{fig:statistics}(a), text-choice questions dominate across most reasoning depths, especially at 5--6 steps, while judgment questions are more common in shorter reasoning chains. Image-choice and counting questions appear more frequently in deeper 6--7 step cases. Fig.~\ref{fig:statistics}(d) further shows that text choice is the primary evaluation format, while judgment and counting are more prominent in dimensions such as OD and QA.

{\bf Reasoning Steps and Length.}
Fig.~\ref{fig:statistics}(b) shows that most samples concentrate around 5-step reasoning, suggesting a moderate level of reasoning complexity. SU and OD mainly involve 4--5 steps, whereas DM and EM include more 6--7 step cases, reflecting deeper decision-making and environmental reasoning. Fig.~\ref{fig:statistics}(e) shows that most CoTs contain 50--150 words. SU and OD tend to have shorter reasoning chains, while EM generally requires longer explanations. These distributions indicate AgroCoT covers simple perception-oriented tasks and more complex agricultural reasoning scenarios.

{\bf High-frequency Words.}
Fig.~\ref{fig:statistics}(c) and (f) visualize high-frequency words in questions and CoTs. Terms such as ``tree'', ``leaf'', ``plant'', ``cultivated'' and ``region'' ominate the question texts, while the CoTs also frequently include agricultural concepts such as ``tree'', ``plant'', ``cultivated'', ``land'', and ``tassel''. Words such as ``analyze'', ``identify'', ``correct'', and ``pattern'' indicate that AgroCoT emphasizes agricultural problem-solving and evidence-based reasoning. Overall, the word distributions confirm that both questions and reasoning chains are grounded in agricultural scenarios.
\section{Evaluation Protocol}
\label{sec:metric}
We evaluate AgroCoT from two complementary perspectives: final-answer correctness and reasoning quality. Beyond accuracy, ROUGE, and BERTScore, which are detailed in Appendix~\ref{app:more_evaluation_protocols}, we introduce four reasoning-oriented metrics to evaluate the quality of model-generated CoTs.

{\bf Semantic Similarity.} We use the all-MiniLM-L6-v2 model from Sentence-Transformers to encode both model-generated CoTs and reference CoTs into 384-dimensional embeddings, and compute cosine similarity to measure their semantic consistency.

{\bf CoT Quality.} Inspired by MME-CoT~\cite{jiang2025mme}, we evaluate whether the model-generated reasoning covers the key steps in the reference CoT. Given the number of key steps in the generated CoT $S_{out}$, the reference CoT $S_{ref}$, and the matched steps $S_{match}$, we compute:
\begin{align}
    \text{Recall} = \frac{S_{match}}{S_{ref}}, \quad
    \text{Precision} = \frac{S_{match}}{S_{out}}, \quad
    \text{F1} = \frac{2 \times \text{Precision} \times \text{Recall}}{\text{Precision} + \text{Recall}}.
\end{align}
The F1 score is used as the final score, reflecting the coverage and validity of the reasoning steps.

{\bf LLM Score.} To reduce the bias of a single evaluator, we employ three open-source LLMs, including Qwen3-8B, DeepSeek-R1-Distill-Qwen-7B, and GLM-Z1-9B-0414, to score each reasoning chain. The evaluation is conducted along five dimensions: accuracy, logical coherence, effectiveness, clarity, and completeness. Each dimension is scored from 1 to 10, and the final LLM score is obtained by averaging the scores across evaluators. The evaluator prompts are provided in Appendix~\ref{app:llm_evaluator}.

{\bf Human Evaluation.} We further conduct human evaluation with 20 trained volunteers with relevant backgrounds. They independently score model-generated reasoning chains using the same criteria as the LLM evaluators, and the final human score is computed as the average score across volunteers. Details are provided in Appendix~\ref{app:human_evaluation}.

Overall, these metrics provide a comprehensive evaluation of both answer correctness and reasoning quality. Semantic similarity and CoT quality measure consistency with reference reasoning, while LLM-based and human evaluations assess reasoning from broader dimensions such as correctness, logic, clarity, and completeness. This design reduces over-reliance on a single reference CoT and allows more flexible evaluation of diverse but valid reasoning paths in agricultural scenarios.

\begin{table*}[t]
\caption{Evaluation results of representative VLMs. SS denotes Semantic Similarity, and Quality refers to the CoT quality metric. \textbf{Boldface} indicates the best result across all models. \colorbox[rgb]{1.0, 0.894, 0.812}{Peach} and \colorbox[rgb]{0.996, 1.0, 0.835}{Lemon} highlight the best and second-best results among open-source models, respectively. All results are reported as mean (standard deviation) over three independent runs.}
\vspace{-0.5em}
\label{tab:eval_results}
\centering
\Huge
\resizebox{\textwidth}{!}{%
\begin{tabular}{c|cc|cc|cc|cc|cc|cc}
\Xhline{1.5pt}
\cellcolor[HTML]{FFFFFF}
& \multicolumn{2}{c}{\cellcolor[HTML]{FFFFFF}\textbf{QA}}
& \multicolumn{2}{c}{\cellcolor[HTML]{FFFFFF}\textbf{SU}}
& \multicolumn{2}{c}{\cellcolor[HTML]{FFFFFF}\textbf{OD}}
& \multicolumn{2}{c}{\cellcolor[HTML]{FFFFFF}\textbf{DM}}
& \multicolumn{2}{c}{\cellcolor[HTML]{FFFFFF}\textbf{EM}}
& \multicolumn{2}{c}{\cellcolor[HTML]{FFFFFF}\textbf{Overall}} \\ \cline{2-13}
\multirow{-2}{*}{\cellcolor[HTML]{FFFFFF}\textbf{Model}}
& \textbf{SS} & \textbf{Quality}
& \textbf{SS} & \textbf{Quality}
& \textbf{SS} & \textbf{Quality}
& \textbf{SS} & \textbf{Quality}
& \textbf{SS} & \textbf{Quality}
& \textbf{SS} & \textbf{Quality} \\
\Xhline{1.2pt}

\multicolumn{13}{c}{\cellcolor[rgb]{0.906, 0.898, 0.902}\textbf{Proprietary Large Vision-Language Models}} \\

Claude-Sonnet-4.5
& $89.93\,(0.53)$ & $48.13\,(0.14)$
& $87.16\,(0.34)$ & $42.87\,(0.01)$
& $\mathbf{87.49\,(0.12)}$ & $47.66\,(0.15)$
& $87.32\,(0.32)$ & $47.91\,(0.08)$
& $84.81\,(0.57)$ & $46.45\,(0.14)$
& $87.81\,(0.47)$ & $46.44\,(0.06)$ \\

Gemini-2.5-Pro~\cite{team2023gemini}
& $81.35\,(0.87)$ & $42.20\,(0.13)$
& $79.43\,(1.12)$ & $33.64\,(0.19)$
& $80.63\,(0.78)$ & $40.09\,(0.04)$
& $80.75\,(0.39)$ & $34.50\,(0.13)$
& $78.45\,(0.59)$ & $37.36\,(0.08)$
& $80.30\,(0.72)$ & $38.64\,(0.11)$\\

GPT-4.1~\cite{achiam2023gpt}
& $\mathbf{91.05\,(0.13)}$ & $51.78\,(0.15)$
& $\mathbf{88.82\,(0.18)}$ & $\mathbf{46.97\,(0.24)}$
& $87.45\,(0.45)$ & $\mathbf{49.93\,(0.13)}$
& $\mathbf{87.84\,(0.76)}$ & $\mathbf{51.68\,(0.30)}$
& $\mathbf{85.40\,(0.51)}$ & $48.12\,(0.11)$
& $\mathbf{88.59\,(0.38)}$ & $\mathbf{49.78\,(0.06)}$\\

GPT-5~\cite{leon2025gpt}
& $76.53\,(0.72)$ & $31.36\,(0.13)$
& $74.83\,(0.63)$ & $28.10\,(0.24)$
& $74.96\,(0.43)$ & $29.33\,(0.11)$
& $74.81\,(0.65)$ & $30.56\,(0.04)$
& $73.16\,(0.43)$ & $29.59\,(0.13)$
& $75.11\,(0.12)$ & $29.78\,(0.24)$\\

Doubao-Seed-1.8~\cite{seedseed1}
& $74.58\,(0.13)$ & $34.93\,(0.34)$
& $70.42\,(0.09)$ & $36.04\,(0.16)$
& $79.75\,(0.06)$ & $41.23\,(0.11)$
& $75.77\,(0.21)$ & $41.67\,(0.03)$
& $80.69\,(0.24)$ & $43.43\,(0.15)$
& $74.66\,(0.05)$ & $38.38\,(0.07)$\\

\Xhline{1.2pt}
\multicolumn{13}{c}{\cellcolor[rgb]{0.906, 0.898, 0.902}\textbf{Open-Source Large Vision-Language Models}} \\

DeepSeek-VL2-small~\cite{lu2024deepseek}
& $33.16\,(0.38)$ & $16.62\,(0.11)$
& $36.22\,(0.26)$ & $7.89\,(0.69)$
& $34.73\,(0.35)$ & $12.64\,(0.76)$
& $40.23\,(0.13)$ & $15.04\,(0.67)$
& $39.02\,(0.76)$ & $11.89\,(0.34)$
& $36.26\,(0.36)$ & $12.78\,(0.56)$\\

InternVL2-8B~\cite{chen2024expanding}
& $80.60\,(0.37)$ & $39.43\,(0.67)$
& $81.95\,(0.96)$ & \cellcolor[rgb]{0.996,1.0,0.835}$39.01\,(0.04)$
& $78.66\,(0.75)$ & $38.42\,(0.81)$
& $75.55\,(0.38)$ & $39.46\,(0.18)$
& $75.85\,(0.12)$ & $39.25\,(0.26)$
& $79.11\,(0.18)$ & $39.12\,(0.46)$\\

InternVL3-14B~\cite{chen2024expanding}
& \cellcolor[rgb]{0.996,1.0,0.835}$85.27\,(0.75)$ & \cellcolor[rgb]{1.0,0.894,0.812}$\mathbf{53.04\,(0.23)}$
& \cellcolor[rgb]{0.996,1.0,0.835}$82.13\,(0.19)$ & $26.10\,(0.14)$
& \cellcolor[rgb]{0.996,1.0,0.835}$82.01\,(0.64)$ & \cellcolor[rgb]{1.0,0.894,0.812}$48.59\,(0.19)$
& \cellcolor[rgb]{0.996,1.0,0.835}$80.65\,(0.45)$ & \cellcolor[rgb]{1.0,0.894,0.812}$47.40\,(0.17)$
& $79.41\,(0.18)$ & \cellcolor[rgb]{1.0,0.894,0.812}$\mathbf{59.41\,(0.46)}$
& \cellcolor[rgb]{0.996,1.0,0.835}$82.32\,(0.37)$ & \cellcolor[rgb]{1.0,0.894,0.812}$44.12\,(0.34)$\\

InternVL3.5-38B~\cite{chen2024expanding}
& $75.17\,(0.08)$ & \cellcolor[rgb]{0.996,1.0,0.835}$46.95\,(0.24)$
& $64.99\,(0.18)$ & $26.57\,(0.16)$
& $70.43\,(0.32)$ & \cellcolor[rgb]{0.996,1.0,0.835}$44.06\,(0.37)$
& $66.21\,(0.24)$ & \cellcolor[rgb]{0.996,1.0,0.835}$47.04\,(0.48)$
& $73.05\,(0.23)$ & $49.94\,(0.39)$
& $69.04\,(0.05)$ & $40.66\,(0.47)$\\

Llama-4-Scout~\cite{touvron2023llama}
& \cellcolor[rgb]{1.0,0.894,0.812}$86.19\,(0.16)$ & $46.10\,(0.44)$
& \cellcolor[rgb]{1.0,0.894,0.812}$84.12\,(0.23)$ & \cellcolor[rgb]{1.0,0.894,0.812}$39.95\,(0.73)$
& \cellcolor[rgb]{1.0,0.894,0.812}$84.36\,(0.21)$ & $43.41\,(0.34)$
& \cellcolor[rgb]{1.0,0.894,0.812}$84.59\,(0.14)$ & $45.75\,(0.03)$
& \cellcolor[rgb]{1.0,0.894,0.812}$81.69\,(0.13)$ & $43.41\,(0.17)$
& \cellcolor[rgb]{1.0,0.894,0.812}$84.53\,(0.28)$ & \cellcolor[rgb]{0.996,1.0,0.835}$43.63\,(0.18)$\\

LLaVA-NeXT-13B~\cite{liu2024improved}
& $29.30\,(1.08)$ & $23.89\,(9.75)$
& $46.65\,(0.76)$ & $27.13\,(6.05)$
& $43.79\,(0.35)$ & $26.23\,(1.08)$
& $33.38\,(0.14)$ & $27.24\,(0.54)$
& $46.38\,(0.17)$ & $26.91\,(2.77)$
& $39.11\,(0.64)$ & $26.15\,(3.99)$\\

Qwen2.5-VL-7B~\cite{bai2023qwen}
& $78.10\,(0.91)$ & $38.13\,(0.01)$
& $79.55\,(0.62)$ & $35.01\,(0.07)$
& $74.64\,(0.14)$ & $32.78\,(0.12)$
& $75.26\,(0.67)$ & $34.53\,(0.31)$
& $71.49\,(0.94)$ & $32.75\,(0.17)$
& $76.61\,(0.39)$ & $35.05\,(0.02)$\\

Qwen3-VL-8B~\cite{bai2023qwen}
& $77.77\,(0.17)$ & $42.72\,(0.17)$
& $68.83\,(0.13)$ & $38.65\,(0.05)$
& $80.14\,(0.42)$ & $41.31\,(0.09)$
& $75.53\,(0.02)$ & $43.23\,(0.23)$
& \cellcolor[rgb]{0.996,1.0,0.835} $80.46\,(0.04)$ & $46.39\,(0.43)$
& $74.83\,(0.04)$ & $41.17\,(0.04)$\\

TinyLLaVA-2B~\cite{zhou2024tinyllava}
& $20.93\,(0.13)$ & $2.40\,(0.02)$
& $41.91\,(0.23)$ & $5.30\,(0.00)$
& $28.06\,(0.14)$ & $5.93\,(0.01)$
& $24.76\,(0.34)$ & $3.65\,(0.01)$
& $23.39\,(0.16)$ & $5.04\,(0.05)$
& $28.81\,(0.37)$ & $4.34\,(0.01)$\\

GLM-4.6V~\cite{hong2025glm}
& $77.10\,(0.19)$ & $43.18\,(0.15)$
& $69.89\,(0.20)$ & $36.80\,(0.14)$
& $79.78\,(0.13)$ & $41.67\,(0.07)$
& $75.00\,(0.09)$ & $44.40\,(0.18)$
& $80.45\,(0.08)$ & \cellcolor[rgb]{0.996,1.0,0.835}$51.13\,(0.23)$
& $74.90\,(0.09)$ & $41.09\,(0.32)$\\

\Xhline{1.5pt}
\end{tabular}%
}
\end{table*}

\section{Experiments}
\label{sec:eval}
\subsection{Experimental Settings}
\label{subsec:setup}
We evaluate a broad set of multimodal models, including five proprietary models, \textit{i.e.}, Claude-Sonnet-4.5, Gemini-2.5-Pro~\cite{team2023gemini}, GPT-4.1~\cite{achiam2023gpt}, GPT-5~\cite{leon2025gpt}, and Doubao-Seed-1.8~\cite{seedseed1}, as well as 25 representative open-source models from the DeepSeek-VL2~\cite{lu2024deepseek}, InternVL~\cite{chen2024expanding}, LLaVA-NeXT~\cite{liu2024improved}, Qwen-VL~\cite{bai2023qwen}, GLM-4.6V~\cite{hong2025glm}, TinyLLaVA~\cite{zhou2024tinyllava}, and Llama-4-Scout~\cite{touvron2023llama} families. Closed-source models are evaluated through official APIs, while open-source models are locally deployed with public weights and standardized configurations. We focus on zero-shot evaluation to assess intrinsic agricultural reasoning ability without task-specific fine-tuning. All local experiments are conducted on NVIDIA A800 GPUs, with detailed hyperparameter settings provided in Appendix~\ref{app:evaluation_models}.

\subsection{Main Results}
\label{subsec:result}
\begin{figure*}[t]
    \centering
    \includegraphics[width=\linewidth]{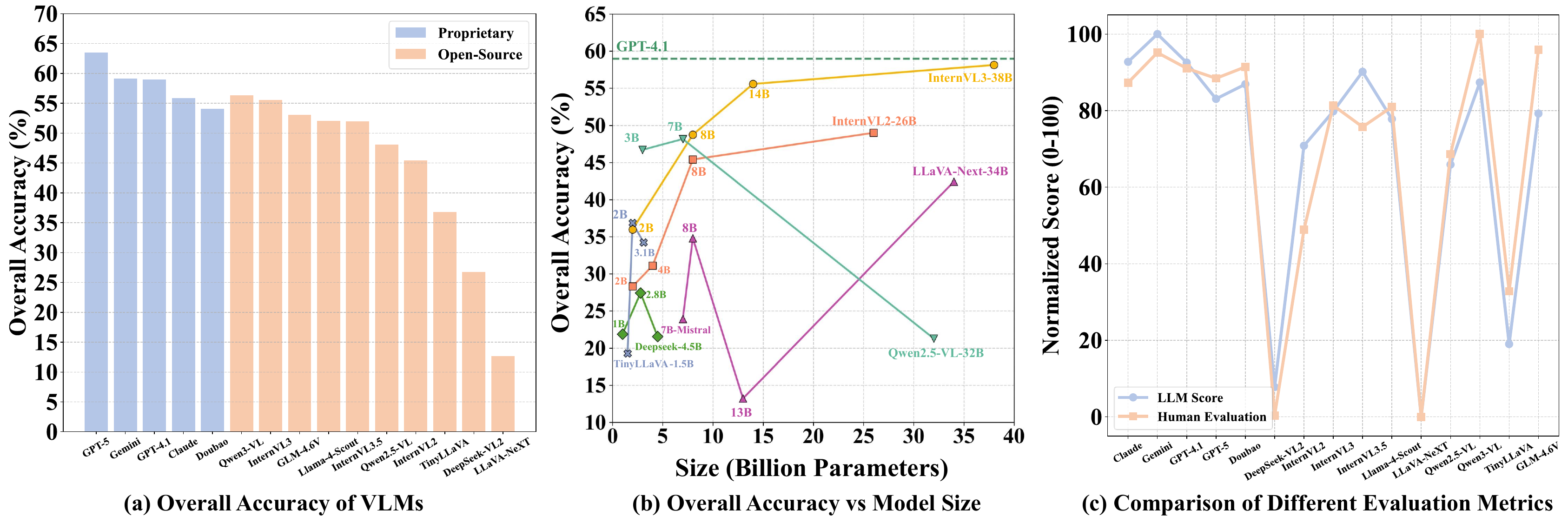}
    \vspace{-1.5em}
    \caption{Performance and analysis of various VLMs across different perspectives.}
    \vspace{-1.5em}
    \label{fig:experiment}
\end{figure*}

We evaluate VLMs on AgroCoT from both final-answer accuracy and reasoning quality. As shown in Fig.~\ref{fig:experiment}(a), proprietary models generally outperform open-source models in accuracy, with GPT-5 achieving the best result above 60\%, followed by Gemini-2.5-Pro and GPT-4.1. Among open-source models, Qwen3-VL-8B and InternVL3-14B are competitive, both exceeding 55\%. However, no model surpasses 70\%, indicating that AgroCoT remains challenging for current VLMs and that agricultural-domain reasoning is still under-optimized in mainstream model development. Table~\ref{tab:eval_results} further reports semantic similarity and key-step matching results for CoT evaluation. GPT-4.1 achieves the highest overall semantic similarity and CoT quality, while Llama-4-Scout, InternVL3-14B, and Qwen3-VL-8B show strong performance among open-source models. Table~\ref{tab:eval_results2} presents LLM-based and human evaluations, where Gemini-2.5-Pro obtains the best LLM score among proprietary models, and Qwen3-VL-8B ranks highest among open-source models in human evaluation. The consistency between automatic and human evaluation suggests that AgroCoT provides a reliable and multi-dimensional assessment of both answer correctness and agricultural reasoning quality. A more detailed analysis of the results can be found in Appendix~\ref{app:main_result}, ROUGE and BERTScore results are provided in Appendix~\ref{app:experimental_results}, and representative case studies are provided in Appendix~\ref{app:dimension_details}, further demonstrating AgroCoT's ability to evaluate model applicability in practical agricultural tasks.

\begin{table*}[t]
\caption{LLM denotes the LLM-based evaluator (maximum score: 50), while Human represents human evaluation scores on a 1–5 scale. \textbf{Boldface} indicates the best overall performance. \colorbox[rgb]{1.0, 0.894, 0.812}{Peach} and \colorbox[rgb]{0.996, 1.0, 0.835}{Lemon} denote the best and second-best results among open-source models, respectively. All results are reported as mean (standard deviation) over three independent runs.}
\vspace{-0.5em}
\label{tab:eval_results2}
\centering
\Huge
\resizebox{\textwidth}{!}{%
\begin{tabular}{ccc|cc|cc|cc|cc|cc}
\Xhline{1.5pt}
\cellcolor[HTML]{FFFFFF}
& \multicolumn{2}{c}{\cellcolor[HTML]{FFFFFF}\textbf{QA}}
& \multicolumn{2}{c}{\cellcolor[HTML]{FFFFFF}\textbf{SU}}
& \multicolumn{2}{c}{\cellcolor[HTML]{FFFFFF}\textbf{OD}}
& \multicolumn{2}{c}{\cellcolor[HTML]{FFFFFF}\textbf{DM}}
& \multicolumn{2}{c}{\cellcolor[HTML]{FFFFFF}\textbf{EM}}
& \multicolumn{2}{c}{\cellcolor[HTML]{FFFFFF}\textbf{Overall}} \\ \cline{2-13}
\multirow{-2}{*}{\cellcolor[HTML]{FFFFFF}\textbf{Model}}
& \textbf{LLM} & \textbf{Human}
& \textbf{LLM} & \textbf{Human}
& \textbf{LLM} & \textbf{Human}
& \textbf{LLM} & \textbf{Human}
& \textbf{LLM} & \textbf{Human}
& \textbf{LLM} & \textbf{Human} \\
\Xhline{1.2pt}
\multicolumn{13}{c}{\cellcolor[rgb]{0.906, 0.898, 0.902}\textbf{Proprietary Large Vision-Language Models}} \\

Claude-Sonnet-4.5
& $40.26(1.40)$ & $1.80(0.28)$
& $40.49(1.86)$ & $2.71(0.10)$
& $43.68(1.54)$ & $\mathbf{4.76(0.25)}$
& $44.24(1.35)$ & $3.39(0.26)$
& $44.68(1.17)$ & $4.11(0.30)$
& $42.25(1.43)$ & $3.60(0.18)$ \\

Gemini-2.5-Pro~\cite{team2023gemini}
& $\mathbf{41.90(0.81)}$ & $2.25(0.07)$
& $\mathbf{43.59(0.88)}$ & $3.61(0.25)$
& $\mathbf{45.65(1.24)}$ & $4.03(0.21)$
& $\mathbf{47.29(0.22)}$ & $3.84(0.15)$
& $47.70(0.31)$ & $4.32(0.30)$
& $\mathbf{44.77(0.56)}$ & $3.81(0.22)$\\

GPT-4.1~\cite{achiam2023gpt}
& $39.41(1.40)$ & $2.50(0.57)$
& $40.26(1.61)$ & $3.36(0.30)$
& $43.29(1.84)$ & $4.24(0.33)$
& $45.11(1.27)$ & $3.47(0.15)$
& $45.85(1.12)$ & $4.12(0.38)$
& $42.18(1.34)$ & $3.70(0.33)$\\

GPT-5~\cite{leon2025gpt}
& $35.17(1.43)$ & $2.60(0.14)$
& $38.08(1.57)$ & $3.46(0.76)$
& $39.70(2.05)$ & $3.94(0.58)$
& $42.08(1.40)$ & $3.47(0.52)$
& $42.40(1.84)$ & $4.00(0.51)$
& $38.89(1.61)$ & $3.63(0.49)$\\

Doubao-Seed-1.8~\cite{seedseed1}
& $35.47(1.85)$ & $2.00(0.47)$
& $36.95(1.41)$ & $3.00(0.35)$
& $44.84(0.21)$ & $4.08(0.35)$
& $46.36(0.42)$ & $3.50(0.71)$
& $\mathbf{47.87(0.47)}$ & $\mathbf{4.64(0.51)}$
& $40.22(1.22)$ & $3.71(0.24)$\\

\Xhline{1.2pt}
\multicolumn{13}{c}{\cellcolor[rgb]{0.906, 0.898, 0.902}\textbf{Open-Source Large Vision-Language Models}} \\

DeepSeek-VL2-small~\cite{lu2024deepseek}
& $7.00(3.59)$ & $1.15(0.07)$
& $10.26(5.16)$ & $1.18(0.05)$
& $18.73(7.54)$ & $1.41(0.08)$
& $16.24(6.71)$ & $1.18(0.04)$
& $16.33(6.45)$ & $1.34(0.03)$
& $12.75(5.65)$ & $1.27(0.02)$\\

InternVL2-8B~\cite{chen2024expanding}
& $31.43(3.63)$ & $1.70(0.14)$
& $33.39(1.54)$ & $2.36(0.20)$
& $37.54(2.54)$ & $2.85(0.46)$
& $36.11(3.82)$ & $2.37(0.22)$
& $37.59(3.65)$ & $2.96(0.45)$
& $34.64(2.94)$ & $2.57(0.33)$\\

InternVL3-14B~\cite{chen2024expanding}
& $35.31(1.98)$ & $2.05(0.07)$
& \cellcolor[rgb]{1.0,0.894,0.812}$36.37(2.80)$ & $2.93(0.20)$
& $39.74(2.55)$ & \cellcolor[rgb]{1.0,0.894,0.812}$4.26(0.12)$
& $39.03(2.59)$ & $3.45(0.04)$
& $41.39(2.28)$ & $3.70(0.08)$
& $37.76(2.45)$ & $3.44(0.10)$\\

InternVL3.5-38B~\cite{chen2024expanding}
& $34.63(1.49)$ & $1.50(0.71)$
& $32.02(3.51)$ & $1.62(0.18)$
& $41.73(2.48)$ & \cellcolor[rgb]{0.996,1.0,0.835}$4.17(0.00)$
& $42.04(2.86)$ & \cellcolor[rgb]{0.996,1.0,0.835}$4.00(0.35)$
& $45.74(2.72)$ & $3.86(0.81)$
& $36.95(2.72)$ & $3.29(0.41)$\\

Llama-4-Scout~\cite{touvron2023llama}
& $35.70(1.82)$ & $1.85(0.07)$
& $33.58(2.54)$ & \cellcolor[rgb]{0.996,1.0,0.835}$3.32(0.25)$
& $37.90(2.31)$ & $3.53(0.50)$
& $40.76(1.87)$ & $3.34(0.11)$
& $41.21(2.08)$ & \cellcolor[rgb]{0.996,1.0,0.835}$4.05(0.33)$
& $37.06(2.10)$ & $3.43(0.27)$\\

LLaVA-NeXT-13B~\cite{liu2024improved}
& $7.71(2.02)$ & $1.05(0.07)$
& $10.39(1.05)$ & $1.07(0.00)$
& $10.63(2.56)$ & $1.15(0.12)$
& $9.68(1.61)$ & $1.37(0.07)$
& $14.41(2.44)$ & $1.43(0.30)$
& $10.04(1.10)$ & $1.26(0.05)$\\

Qwen2.5-VL-7B~\cite{bai2023qwen}
& $30.13(1.88)$ & $2.15(0.21)$
& $30.27(3.12)$ & $2.46(0.35)$
& $36.78(3.34)$ & $3.44(0.29)$
& $35.28(3.99)$ & $3.26(0.45)$
& $35.18(3.48)$ & $3.43(0.20)$
& $32.93(3.00)$ & $3.10(0.30)$\\

Qwen3-VL-8B~\cite{bai2023qwen}
& \cellcolor[rgb]{1.0,0.894,0.812}$40.82(0.43)$ & \cellcolor[rgb]{0.996,1.0,0.835}$3.83(0.24)$
& \cellcolor[rgb]{0.996,1.0,0.835}$34.94(1.99)$ & $2.88(0.18)$
& \cellcolor[rgb]{0.996,1.0,0.835}$44.05(0.67)$ & $4.00(0.24)$
& \cellcolor[rgb]{1.0,0.894,0.812}$45.89(0.20)$ & \cellcolor[rgb]{1.0,0.894,0.812}$\mathbf{4.62(0.53)}$
& \cellcolor[rgb]{0.996,1.0,0.835}$47.19(0.12)$ & \cellcolor[rgb]{1.0,0.894,0.812}$4.14(0.40)$
& \cellcolor[rgb]{1.0,0.894,0.812}$40.38(1.18)$ & \cellcolor[rgb]{1.0,0.894,0.812}$\mathbf{3.94(0.32)}$\\

TinyLLaVA-2B~\cite{zhou2024tinyllava}
& $13.07(4.22)$ & $1.60(0.00)$
& $13.08(3.06)$ & $2.14(0.10)$
& $19.84(1.16)$ & $2.53(0.17)$
& $20.11(3.33)$ & $2.37(0.37)$
& $22.44(2.80)$ & $1.93(0.25)$
& $16.65(3.06)$ & $2.14(0.21)$\\

GLM-4.6V~\cite{hong2025glm}
& \cellcolor[rgb]{0.996,1.0,0.835}$38.54(1.85)$ & \cellcolor[rgb]{1.0,0.894,0.812}$\mathbf{4.00(0.47)}$
& $33.97(1.85)$ & \cellcolor[rgb]{1.0,0.894,0.812}$\mathbf{3.75(0.00)}$
& \cellcolor[rgb]{1.0,0.894,0.812}$45.25(1.74)$ & $4.00(0.47)$
& \cellcolor[rgb]{0.996,1.0,0.835}$45.15(1.61)$ & $3.38(0.53)$
& \cellcolor[rgb]{1.0,0.894,0.812}$47.85(1.25)$ & $3.93(0.51)$
& \cellcolor[rgb]{0.996,1.0,0.835}$39.82(1.71)$ & \cellcolor[rgb]{0.996,1.0,0.835}$3.83(0.41)$\\

\Xhline{1.5pt}
\end{tabular}%
}
\end{table*}

\subsection{Discussion}
\textbf{Does model performance scale with the size?}
To enable a more comprehensive evaluation across model families and scales, we test multiple variants with different parameter sizes from several representative VLM families. Fig.~\ref{fig:experiment} (b) illustrates how the accuracy of models varies with model size on AgroCoT. Although the accuracy of most models improves as the model size increases, there are still some models whose performance actually decrease, such as DeepSeek-VL2 and Qwen2.5-VL. This indicates that increasing the size of model does not enhance its capabilities in agriculture all the time. In Appendix~\ref{app:reasoning_ability}, we provide an analysis of how reasoning capabilities change with model size. 

\textbf{Do LLMs and humans share similar evaluation tendencies for reasoning?} Using LLMs as automatic evaluators is increasingly common, but their consistency with human reasoning preferences in agricultural scenarios remains important to verify. As shown in Fig.~\ref{fig:experiment}(c), we compare normalized LLM scores with human scores. Although the two protocols select different top models, with LLM scoring favoring Gemini-2.5-Pro and human scoring favoring Qwen3-VL-8B, their overall trends across models are highly consistent. This suggests that, with carefully designed prompts, LLM-based evaluation can reasonably reflect human preferences in agricultural reasoning tasks. Therefore, our LLM scoring provides a reliable and cost-effective complement to human evaluation, especially for large-scale benchmark assessment. 


\textbf{Are these models good at multi-step reasoning?}
AgroCoT provides reasoning chains with varying lengths, enabling analysis of model reasoning depth. As shown in Fig.~\ref{fig:score_by_para} of Appendix~\ref{app:vlms_deep_reasoning}, model performance varies across reasoning steps: VLMs struggle with simple 3-step tasks and medium-complexity 5--6-step tasks, but obtain higher LLM scores on deeper reasoning tasks. This suggests that current VLMs may have difficulty producing concise and precise reasoning for simpler cases, while they can generate more structured explanations when explicitly required to reason deeply. However, higher LLM scores may also reflect redundant reasoning rather than true efficiency. Additional ROUGE and BERTScore analyses in Appendix~\ref{app:experimental_results} further examine this trade-off.


\section{Conclusion}
\label{sec:conclu&lim}

We introduce AgroCoT, a benchmark designed to evaluate vision-language models in the agricultural domain, consisting of 4,759 high-quality VQA pairs sourced from both public and private agricultural datasets. To address the limitations of existing benchmarks, we expanded the dataset by annotating each VQA pair with a CoT, enabling a more comprehensive evaluation of model capabilities in both answer generation and complex reasoning. Additionally, we introduce multi-dimensional evaluation metrics from four distinct perspectives to establish a systematic evaluation framework. Through the evaluations of 30 models, we find that both proprietary and open-source VLMs encounter significant challenges in agricultural reasoning, without notable improvements observed in version updates. We hope AgroCoT will offer valuable guidance for the development of agriculture-specific models and foster the integration of domain knowledge with reasoning capabilities.

\newpage

{
\small
\bibliographystyle{unsrtnat}
\bibliography{ref}
}


\appendix

\definecolor{mybrown}{RGB}{191, 144, 0}
\definecolor{mygreen}{RGB}{84, 130, 53}

\newpage
\appendix
\begin{center}
    \Large
    \textbf{\includegraphics[height=1em]{figures/logo.png}AgroCoT: A Chain-of-Thought Benchmark for Evaluating Reasoning in Vision-Language Models for Agriculture \\(Supplementary material)} 
    \vspace{-1.0em} 
\end{center}


\setlength{\cftbeforesecskip}{1pt}
\setlength{\cftbeforesubsecskip}{0pt}
\setlength{\cftbeforesubsubsecskip}{0pt}
\setlength{\cftaftertoctitleskip}{1pt}

\startcontents[sections]
\printcontents[sections]{}{1}{\section*{Table of Contents in Appendix}\setcounter{tocdepth}{3}}

\newpage

\newpage
\section{More Related Works}
\label{app:related_work}
\subsection{Large Vision-Language Models}
Vision-Language Models (VLMs) that couple visual encoders with large language models have rapidly advanced general-purpose multimodal understanding~\cite{li2025benchmark}, with CLIP~\cite{pmlr-v139-radford21a} establishing scalable vision-language representations for strong zero-shot transfer. Proprietary models like GPT‑4o~\cite{achiam2023gpt} and Gemini~\cite{team2023gemini} push multimodal reasoning at scale, while open-weight models such as DeepSeek~\cite{liu2024deepseek} and Qwen‑VL~\cite{bai2023qwen} lower the barrier for domain adaptation in specialized applications.

Adapting VLMs to agriculture shows promise but also reveals gaps. Domain-specific initiatives include AgroGPT~\cite{awais2025agrogpt} for agricultural VQA, Agri-LLaVA~\cite{wang2024agri} for crop disease diagnosis, and vision-language pipelines for crop health monitoring, disease recognition~\cite{zhang2024visual} or parcel segmentation~\cite{wu2025farmseg_vlm}. 
However, these models are often trained and evaluated on crop‑ or symptom‑specific datasets with limited scenes and sensors~\cite{liu2024multimodal, zhang2024visual}, and current evaluations rarely cover practical skills such as counting, planting recommendations or environmental management. Therefore, there is an urgent need for a benchmark that spans a broader range of agricultural scenes across multiple sensing platforms, so that the capabilities of models can be assessed more faithfully in real‑world field conditions.

\subsection{Chain-of-Thought}
Chain-of-Thought (CoT)~\cite{wang2025multimodal} helps large models solve complex problems by breaking them into intermediate steps. Simple prompts can elicit stepwise reasoning and improve performance on arithmetic, commonsense, and logic tasks~\cite{wei2022chain, kojima2022large, jiang2025mme}. Reliability is further enhanced through self-consistency voting~\cite{wang2022self} and systematic search over intermediate states~\cite{yao2023tree}. In multimodal settings, textual rationales should align with visual evidence~\cite{singh2023assessing}, as in ScienceQA~\cite{lu2022learn}, which pairs images, questions, and human-written rationales to supervise step-by-step reasoning. Compositional CoT~\cite{mitra2024compositional} uses scene-graph intermediates, and Interleaved-modal CoT~\cite{gao2025interleaved} inserts attention-selected regions. 

Despite this progress, multimodal CoT benchmarks~\cite{qian2025prism, li2025think, wang2025chain, zhang2025video, guo2025rbench, jiang2025mme, gao2025benchmarking} remain limited in reasoning depth, image diversity, and domain coverage. ScienceQA~\cite{lu2022learn} targets K--12; MMMU~\cite{yue2024mmmu} and MathVista~\cite{lu2023mathvista} emphasize cross-disciplinary and visual mathematics without explicit long-chain annotations; A-OKVQA~\cite{schwenk2022okvqa} offers training or multiple-choice rationales but does not evaluate generated intermediate steps at scale; M$^{3}$CoT~\cite{chen2024m3cot} and MME-CoT~\cite{jiang2025mme} enforce multi-domain, multi-step, multimodal reasoning but do not cover agriculture. This reveals a core limitation for agricultural multimodal reasoning: the absence of a benchmark with detailed CoT for qualitative assessment. Motivated by this, we introduce explicit, structured CoT for agricultural scenarios, which enables evaluation of the CoT generated by models and provides deeper insights into how well models understand and analyze complex agricultural problems.

\section{Data Collection}
\label{app:data_collection}
\subsection{Data Sources Selection}
\label{app:data_sources_selection}
AgroCoT is constructed based on four state-of-the-art agricultural vision-language benchmarks (see Table~\ref{tab:dataset_summary}), each offering unique characteristics and specialized annotations for comprehensive evaluation of VLMs in agriculture.

\textbf{IP102~\cite{wu2019ip102}} is a large-scale benchmark dataset for insect pest recognition. It contains diverse pest categories with fine-grained visual differences, making it useful for evaluating agricultural vision models on pest identification and long-tailed category recognition. In AgroCoT, IP102 mainly contributes to pest-related visual understanding, including insect recognition, pest category discrimination, and reasoning over visually similar agricultural threats.

\textbf{Agriculture-Vision Corpus (AVC)~\cite{chiu2020agriculture}} is a large aerial image dataset designed for agricultural pattern analysis. It provides high-resolution field imagery and supports the recognition of farmland patterns, crop conditions, and abnormal agricultural regions. In AgroCoT, AVC enriches aerial-view agricultural understanding and contributes to tasks related to field-scene interpretation, anomaly identification, and remote observation of farmland conditions.

\textbf{OilPalmUAV~\cite{zheng2021growing}} focuses on growth status observation of oil palm trees using UAV imagery. It provides plantation-scale visual data captured from low-altitude aerial platforms, enabling tree-level monitoring and growth assessment. In AgroCoT, this dataset supports UAV-based crop monitoring, plantation management, and reasoning about crop growth status from aerial agricultural images.

\textbf{PhenoBench~\cite{weyler2024phenobench}} is an agricultural benchmark for semantic image interpretation in field environments. It provides fine-grained annotations for understanding crops, weeds, plants, and leaves, which are essential for phenotyping and field-level agricultural perception. In AgroCoT, PhenoBench contributes to detailed crop-weed discrimination, plant structure understanding, and fine-grained visual reasoning in real agricultural scenes.

\textbf{CropHarvest~\cite{tseng2021cropharvest}} is a global dataset for crop-type classification based on satellite observations. It covers diverse geographic regions and crop categories, supporting large-scale crop monitoring and remote-sensing-based agricultural analysis. In AgroCoT, CropHarvest expands the benchmark toward global crop-type understanding, remote sensing interpretation, and reasoning over large-scale agricultural land-cover patterns.

\textbf{CDDM~\cite{liu2024multimodal}} enhances our dataset with \textbf{large-scale crop disease coverage}, comprising 137,000 disease images and 1 million question-answer pairs spanning 16 crop types and 60 disease categories. The dataset combines web-collected imagery with field survey photographs, ensuring diverse representation of agricultural scenarios. A distinctive aspect is the inclusion of GPT-4 generated instruction-following data with carefully designed negative response examples, addressing the critical challenge of model over-confidence in disease diagnosis.

\textbf{AGMMU~\cite{gaubaagmmu}} contributes \textbf{real-world conversational data} derived from 116,231 authentic dialogues between farmers and USDA-authorized agricultural experts. This dataset uniquely provides both multiple-choice questions (MCQs) and open-ended questions (OEQs), with 746 instances of each format, all rigorously verified by human annotators. The accompanying AGBASE development corpus contains 57,079 multimodal facts, facilitating comprehensive model training and evaluation across five critical agricultural knowledge types: insect identification, species recognition, disease categorization, symptom description, and management instruction.

\begin{table*}[t]
    \centering
    \caption{More details of data sources used in AgriCoT.} 
    \label{tab:dataset_summary}
    \scriptsize
    \begin{tabularx}{\textwidth}{
        >{\centering\arraybackslash}m{2.2cm}   
        !{\vrule width 0.5pt}
        >{\centering\arraybackslash}m{1.0cm}   
        !{\vrule width 0.5pt}
        >{\raggedright\arraybackslash}X
    }
        \toprule 
        \textbf{Dataset} & \textbf{Year} & \textbf{Key Features}  \\
        \midrule

        IP102~\cite{wu2019ip102} & 
        2019 & 
        Pest recognition benchmark covering diverse insect categories and fine-grained pest appearances \\

        \midrule

        AVC~\cite{chiu2020agriculture} & 
        2020 & 
        Aerial imagery dataset for agricultural pattern analysis under diverse field conditions \\

        \midrule

        OilPalmUAV~\cite{zheng2021growing} & 
        2021 & 
        UAV imagery dataset for oil palm tree monitoring and growth status assessment \\

        \midrule

        PhenoBench~\cite{weyler2024phenobench} & 
        2024 & 
        Field-level agricultural vision benchmark for crop, weed, plant, and leaf understanding \\

        \midrule

        CropHarvest~\cite{tseng2021cropharvest} & 
        2021 & 
        Global remote-sensing dataset for crop-type classification across diverse geographic regions \\

        \midrule
        
        CDDM~\cite{liu2024multimodal} & 
        2024 & 
        137,000 images of various crop diseases; 1 million question-answer pairs; for the crop and disease categories, detailed disease insights, and prevention and control strategies \\
        \midrule
        
        AGMMU~\cite{gaubaagmmu} & 
        2025 & 
        746 multiple-choice questions and 746 open-ended questions; from real-world conversations between users and USDA-funded experts; human verification; five major agricultural knowledge types \\
        \midrule

        AgroBench~\cite{shinoda2025agrobench} & 
        2025 & 
        4,342 QA pairs, 203 crop categories and 682 disease categories; 98 machine categories and 77 traditional management methods; multiple-choice format; annotated by human agronomist experts  \\
        
        \bottomrule
    \end{tabularx}
\end{table*}

\begin{figure}[t]
    \centering
    \includegraphics[width=0.95\linewidth]{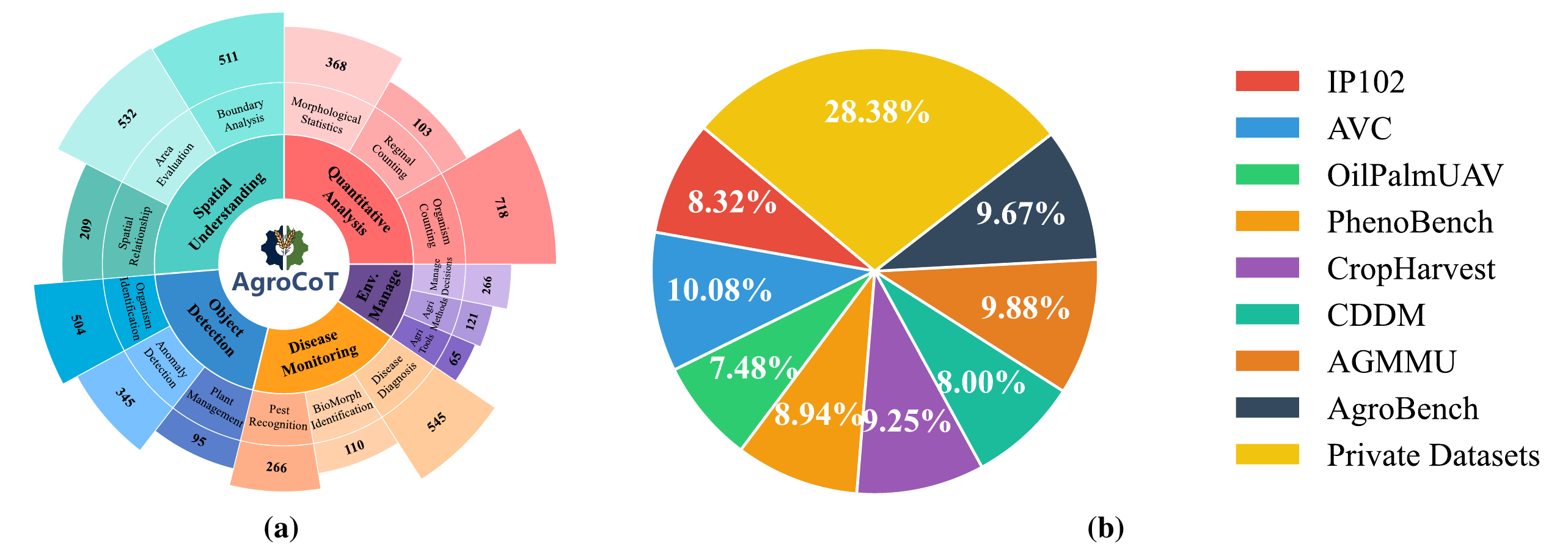}
    \caption{Overview of AgroCoT data composition. (a) Hierarchical task distribution of AgroCoT. (b) Proportion of samples from different data sources.}
    \label{fig:PercentageOfDatasets}
\end{figure}

\textbf{AgroBench~\cite{shinoda2025agrobench}} serves as a foundational component of our dataset, distinguished by its \textbf{expert annotations from professional agronomists}. This benchmark covers 7 distinct agricultural tasks including disease identification, pest recognition, weed detection, crop management, disease management, machinery usage, and traditional farming methods. With an extensive coverage of 203 crop categories, 682 disease types, 134 pest species, and 108 weed varieties, it provides 4,342 carefully curated question-answer pairs supported by 3,745 high-quality images captured in real farm environments. A notable feature is the inclusion of bounding box annotations for precise weed localization, enabling fine-grained visual understanding tasks.

\subsection{Task Distribution}
\label{app:task_distribution}
Figure~\ref{fig:PercentageOfDatasets} (a) presents the hierarchical task organization of AgroCoT. The benchmark is structured around five major agricultural reasoning dimensions: Spatial Understanding, Quantitative Analysis, Environment Management, Disease Monitoring, and Object Detection. Each dimension is further decomposed into fine-grained task types. For example, Spatial Understanding includes tasks such as boundary analysis, area evaluation, and spatial relationship reasoning, which assess a model's ability to understand field layout and spatial structures. Quantitative Analysis covers counting- and measurement-related tasks, including morphological statistics, regional counting, and organism counting. Disease Monitoring focuses on agricultural health-related reasoning, such as pest recognition, biological morphology identification, and disease diagnosis. Object Detection emphasizes the recognition and localization of agricultural objects, anomalies, and plant-related targets. Environment Management further introduces decision-oriented reasoning scenarios related to agricultural risks and management actions. This hierarchical design enables AgroCoT to evaluate VLMs beyond simple visual recognition, covering multi-level agricultural reasoning from perception and counting to diagnosis, spatial analysis, and practical decision support.

\subsection{Data Filtering}
\label{app:data_filtering}
To build a comprehensive and balanced agricultural benchmark, we carefully curate and filter samples from multiple public datasets and our self-constructed private dataset. As shown in Figure~\ref{fig:PercentageOfDatasets} (b), AgroCoT contains 4,759 samples from nine data sources: IP102~\cite{wu2019ip102}, AVC~\cite{chiu2020agriculture}, OilPalmUAV~\cite{zheng2021growing}, PhenoBench~\cite{weyler2024phenobench}, CropHarvest~\cite{tseng2021cropharvest}, CDDM~\cite{liu2024multimodal}, AGMMU~\cite{gaubaagmmu}, AgroBench~\cite{shinoda2025agrobench}, and private datasets. Specifically, IP102~\cite{wu2019ip102} contributes 1,351 samples, providing the largest share and strengthening pest-related recognition and reasoning. AVC~\cite{chiu2020agriculture}, OilPalmUAV~\cite{zheng2021growing}, PhenoBench~\cite{weyler2024phenobench}, and CropHarvest~\cite{tseng2021cropharvest} contribute 480, 356, 425, and 440 samples, respectively, enriching AgroCoT with aerial observation, UAV-based monitoring, field-level semantic understanding, and remote-sensing-based crop classification. CDDM~\cite{liu2024multimodal}, AGMMU~\cite{gaubaagmmu}, and AgroBench~\cite{shinoda2025agrobench} contribute 381, 470, and 460 samples, respectively, supporting multimodal disease diagnosis, agricultural knowledge reasoning, and expert-oriented vision-language evaluation. In addition, 396 private samples are introduced to improve scenario diversity and complement underrepresented agricultural tasks. Overall, this integration strategy enables AgroCoT to synthesize the complementary strengths of different datasets and construct a unified evaluation framework that spans:
\begin{itemize}
\item \textbf{Multiple agricultural domains}: From fine-grained disease identification to practical management recommendations.
\item \textbf{Diverse annotation formats}: Including multiple-choice, open-ended, and judgement tasks.
\item \textbf{Real-world relevance}: Incorporating authentic farmer queries and expert-verified responses.
\item \textbf{Comprehensive coverage}: Encompassing crops, diseases, pests, weeds, machinery, and traditional practices.
\item \textbf{Multiview image sources}: Including images captured by remote sensing satellites, low-altitude drone imagery, and ground-based handheld cameras.
\item \textbf{Multi-dimensional evaluation}: Supporting assessment across object detection, quantitative analysis, disease monitoring, spatial understanding, and environmental management.
\end{itemize}

This comprehensive integration enables robust assessment of VLMs in addressing practical agricultural challenges while maintaining scientific rigor through expert validation and diverse task formulations.

\subsection{Geospatial Distribution}
\label{app:geospatial_distribution}
\begin{figure*}[t]
    \centering
    \includegraphics[width=\textwidth]{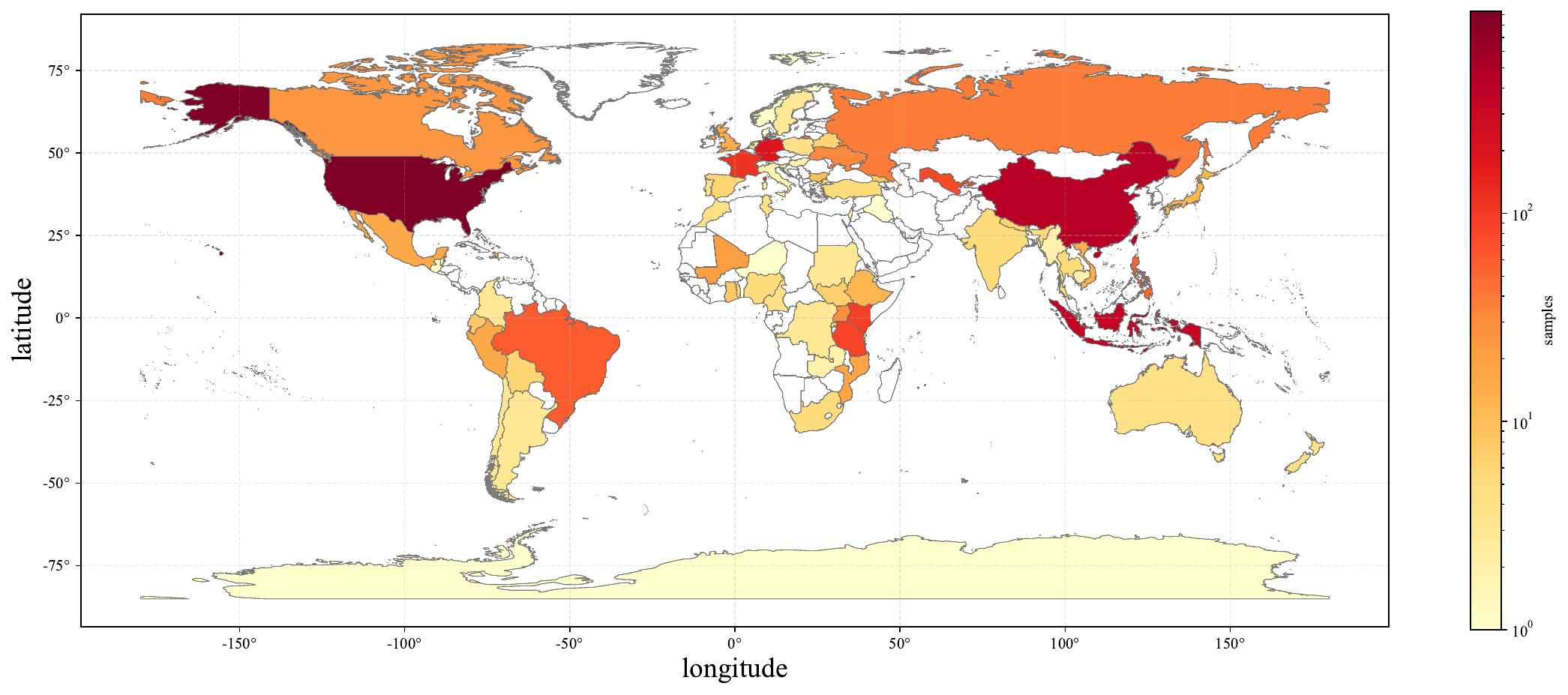}
    \caption{Worldwide Distribution of AgroCoT.}
    \label{fig:world}
\end{figure*}
AgroCoT retains substantial geographical diversity even after sampling, ensuring that the dataset continues to represent a wide spectrum of agricultural environments (see Figure~\ref{fig:world}). Moreover, the incorporation of additional private datasets further expands the spatial coverage, enriching the diversity of regions and strengthening the global representativeness of AgroCoT.

Specifically, the dataset covers a broad range of climatic and agricultural zones:
\begin{itemize}
    \item \textbf{Climate diversity:} The samples span tropical regions (\eg, Indonesia), temperate zones (\eg, European countries, Japan), and continental climates (\eg, Ukraine, Russia), ensuring representation across major global climate regimes.
    \item \textbf{Land-use variation:} AgroCoT includes data from highly industrialized monoculture systems (such as the U.S.\ Corn Belt) as well as diverse polyculture farming landscapes (\eg, Tanzania), capturing a wide range of agricultural practices and intensities.
    \item \textbf{Crop phenology:} The dataset reflects growth cycles of major staple crops including maize, rice, and oil palm, providing coverage across key phenological stages in different agricultural regions.
\end{itemize}

This rich and globally distributed dataset enables comprehensive cross-regional generalization analysis, supporting research on robust agricultural decision-making and sustainable global crop management.

\section{CoT Construction Details}
\label{app:cot_construction_details}

\subsection{Prompt Design}
\label{app:prompt_design}
\begin{figure}[t]
    \centering
    \includegraphics[width=\linewidth]{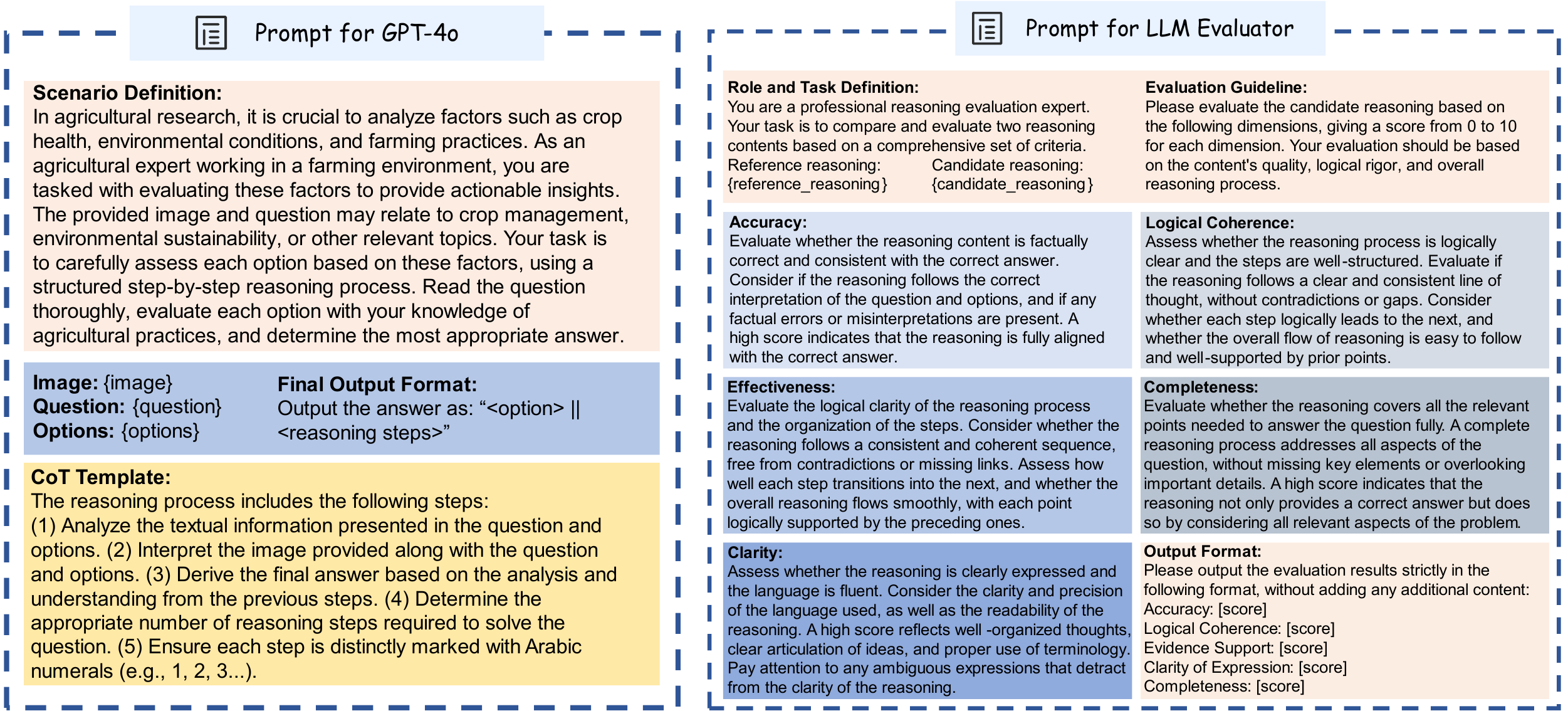}
    \caption{CoT pre-generation prompt (left) and LLM Scoring prompt (right).}
    \vspace{-1.25em}
    \label{fig:prompt}
\end{figure}

To ensure that the GPT-4o-generated CoTs are structured, informative, and aligned with agricultural reasoning practice, we design a prompt template under the guidance of agricultural experts, as shown in Figure~\ref{fig:prompt} (left). In addition to the basic inputs, including the reference image, question, and options, the template explicitly instructs the model to identify the core problem, describe relevant visual evidence, analyze each candidate option, and derive the final answer through a complete reasoning process. This expert-designed template helps standardize the reasoning format while encouraging the model to consider agriculturally meaningful cues, such as crop appearance, plant organs, disease symptoms, field conditions, and management-related context when applicable.

This structured prompting strategy is used to guide multi-step reasoning from visual observation to option analysis and final conclusion, while reducing overly brief, fragmented, or unsupported explanations. To ensure fairness and reproducibility during evaluation, all compared models are evaluated under the same standardized input setting. Specifically, for each test sample, the identical prompt template is provided to every model, so that performance differences mainly reflect the models' reasoning and domain understanding abilities rather than variations in prompt design.

\subsection{Wrong CoT Filtering}
\label{app:wrong_cot_filtering}
GPT-4o’s pre-generated answers and reasoning chains frequently exhibit issues such as incorrect conclusions, formatting errors, missing details, shallow reasoning, and logical inconsistencies. To mitigate this, we adopt a rigorous two-stage screening procedure. First, we discard samples whose final answers are incorrect, as erroneous conclusions usually indicate fundamental flaws in the underlying reasoning. Second, for samples with correct answers, we do not accept them directly; instead, we carefully re-check each chain against criteria on length, formatting, content completeness, and reasoning rigor. This process filters out CoTs that fail to meet our standards of well-structured format, deep reasoning, multi-step derivation, and logical consistency, and flags them for subsequent human refinement.

\subsection{Human Refinement}
\label{app:human_refinement}
To ensure the quality and reliability of CoT annotations in AgroCoT, we established a human-in-the-loop refinement protocol involving 20 trained annotators. The annotation team consisted of 5 undergraduate students and 15 graduate students with backgrounds in agricultural science, computer science, or related fields. Before participating in the annotation process, all annotators received standardized training on agricultural scenarios, CoT annotation principles, and quality-control criteria, and were required to pass a qualification test to ensure a consistent understanding of the task requirements.

During annotation, GPT-4o-generated CoTs were treated only as initial drafts rather than final references. Annotators were instructed to carefully verify each reasoning chain against the image, question, and answer, and to revise the content when necessary. The refinement guideline covered several aspects, including correcting factual or agronomic errors, removing unsupported assumptions, enriching missing visual evidence, improving logical coherence, and standardizing the reasoning format. In particular, annotators were asked to ensure that each CoT was grounded in observable image evidence and consistent with basic agricultural knowledge, rather than merely preserving the language style or reasoning pattern of the initial model-generated output.

To further improve annotation reliability, we adopted multiple rounds of manual review and cross-validation. Ambiguous cases, visually insufficient samples, or reasoning chains containing potential hallucinations were revised or filtered after discussion. This process ensured that the final CoTs satisfied three key requirements: visual grounding, agricultural validity, and logical consistency.

Our annotation statistics show that more than 70\% of the initial CoTs were manually edited, expanded, or restructured. This substantial human intervention leads to clear differences between the final annotations and the original GPT-4o-generated drafts. Therefore, AgroCoT should not be viewed as a direct collection of GPT-4o rationales. Instead, it is a manually refined and quality-controlled agricultural reasoning benchmark. This design helps reduce potential teacher-style bias from GPT-4o and provides a fairer basis for evaluating the reasoning ability of different multimodal models.

\section{Experiment Details}
\label{app:experiment_details}
\subsection{More Evaluation Protocols}
\label{app:more_evaluation_protocols}
We use accuracy as one of the core evaluation metrics, measuring the consistency between the final answer and the standard answer. Given the diverse answer formats in our dataset, we adopt different evaluation strategies tailored to each question type. For multiple-choice and true/false questions, correctness is determined via strict character-level exact matching. For open-ended short responses, we employ a predefined keyword-matching mechanism that allows limited lexical variation while enforcing semantic constraints. For long-form question answering, both model-generated and reference answers are projected into a shared embedding space using a language model, and cosine similarity is computed to quantify semantic alignment.

We supplement the evaluation with ROUGE-1, ROUGE-2, and ROUGE-L, computing precision, recall, and F1 score for each. ROUGE measures n-gram overlap between generated and reference CoTs, reflecting content coverage and sequence structure. Precision indicates how well the model captures relevant information, while recall shows how comprehensively the model covers the reference. The F1 score balances both metrics. We also use BERTScore to evaluate precision and recall in the embedding space, comparing semantic similarity using cosine similarity of word embeddings. Precision in BERTScore reflects how much of the generated reasoning matches the reference semantically, while recall assesses how well the model captures the reference’s semantic content. F1 score for BERTScore balances both, providing a measure of semantic quality and completeness. These metrics allow for a comprehensive evaluation of both surface-level n-gram overlap and deeper semantic alignment.

\subsection{Evaluation Models}
\label{app:evaluation_models}

As shown in Table~\ref{tab:model_comparison_grouped}, we evaluate both proprietary and open-source vision-language models that represent the current state of multimodal reasoning. When a model family provides several parameter scales, we include a small and a large variant in order to study the effect of model size on both answer accuracy and CoT quality. Unless otherwise noted, all inputs contain the same images and textual questions, and outputs are requested in the unified “answer then reasoning” format. We use nucleus sampling with temperature=0.1, top-p=0.7 for all models. For models that expose an explicit “thinking mode” switch via API, we enable thinking mode with a fixed configuration (detailed below). For all other decoding parameters, we use the default values recommended by each provider. All local inference for open-source models was run on machines equipped with NVIDIA A800 GPUs, and proprietary models were accessed via their official APIs.


\textbf{Proprietary Models.} We consider five major proprietary multimodal systems accessed through their official APIs: GPT-5 and GPT-4.1 from OpenAI~\cite{leon2025gpt}, Gemini-2.5-Pro from Google~\cite{comanici2025gemini}, Claude-Sonnet-4.5 from Anthropic, and Doubao-Seed-1.8 from Volcano Engine~\cite{seedseed1}. All of them support joint image and text input and can produce long-form natural-language responses. For each model we use the recommended chat completion interface with a unified decoding configuration, and we supply a short system instruction that asks the model to first provide the final answer and then provide step-by-step reasoning. We do not apply task-specific fine-tuning, so the reported results reflect the intrinsic zero-shot capabilities of these proprietary models on AgroCoT.

For proprietary models that provide explicit thinking-mode controls, we enable thinking mode with fixed settings: GPT-5 is run with reasoning effort set to medium, Claude-Sonnet-4.5 is run with thinking enabled and budget\_tokens=8192, Doubao-Seed-1.8 is run with think-medium. Other proprietary models are queried using their recommended chat completion interfaces under the same sampling configuration, while the models without an explicit thinking-mode switch are used with their default behavior. 

\textbf{DeepSeek-VL2}~\cite{lu2024deepseek}. DeepSeek-VL2 is an open-source VLM family that integrates a vision encoder with a DeepSeek language backbone for general vision-language understanding. We evaluate three parameter scales, DeepSeek-VL2-tiny (1B parameters), DeepSeek-VL2-small (2.8B), and DeepSeek-VL2 (4.5B). All checkpoints are loaded from the official HuggingFace releases, and inference follows the evaluation script.

\begin{table}[t]
    \centering
    \caption{Models grouped by family.} 
    \label{tab:model_comparison_grouped} 
    \scriptsize
    \begin{tabularx}{\linewidth}{
    >{\centering\arraybackslash}m{2cm}
    !{\vrule width 0.5pt}
    >{\centering\arraybackslash}m{2.3cm}
    !{\vrule width 0.5pt}
    >{\centering\arraybackslash}m{1.0cm}
    !{\vrule width 0.5pt}
    >{\centering\arraybackslash}X
}
        \toprule
        \textbf{Model Family} & \textbf{Model Name} & \textbf{Year} & \textbf{Parameters} \\
        \specialrule{\lightrulewidth}{0pt}{0pt}
        \rowcolor[HTML]{FCE4D6}
        \multicolumn{4}{l}{
          \rule{0pt}{3.0ex}\textbf{Proprietary, API}\rule[-1.5ex]{0pt}{3.0ex}
        } \\
        \specialrule{\lightrulewidth}{0pt}{0.6ex}
        
        \multirow[c]{2}{*}{GPT}
          & GPT-5 & 2025 & N/A \\
          & GPT-4.1 & 2025 & N/A \\

        \midrule 
        \multirow[c]{1}{*}{Gemini}
            & Gemini-2.5-Pro & 2025 & N/A \\
            
        \midrule 
        Claude & Claude-Sonnet-4.5 & 2025 & N/A \\

        \midrule 
        Doubao & Doubao-Seed-1.8 & 2025 & N/A \\
        \specialrule{\lightrulewidth}{0pt}{0pt}
        \rowcolor[HTML]{FFF2CC}
        \multicolumn{4}{l}{%
          \rule{0pt}{3.0ex}\textbf{Open-source}\rule[-1.5ex]{0pt}{3.0ex}
        } \\
        \specialrule{\lightrulewidth}{0pt}{0.6ex}
        
        \multirow[c]{5}{*}{DeepSeek-VL2}
            & DeepSeek-VL2-tiny & 2024 & 1B \\
            & DeepSeek-VL2-small & 2024 & 2.8B \\
            & DeepSeek-VL2 & 2024 & 4.5B \\
        \midrule

        \multirow[c]{8}{*}{InternVL} 
          & InternVL2-2B & 2024 & 2B \\
          & InternVL2-4B & 2024 & 4B \\
          & InternVL2-8B & 2024 & 8B \\
          & InternVL2-26B & 2024 & 26B \\
          \cmidrule{2-4}
          & InternVL3-2B & 2025 & 2B \\
          & InternVL3-8B & 2025 & 8B \\
          & InternVL3-14B & 2025 & 14B \\
          & InternVL3-38B & 2025 & 38B \\
          \cmidrule{2-4}
          & InternVL3.5-38B & 2025 & 38B \\
        \midrule 

        \multirow[c]{4}{*}{LLaVA-NeXT} 
          & LLaVA-NeXT-7B-Mistral & 2024 & 7B \\
          & LLaVA-NeXT-8B & 2024 & 8B \\
          & LLaVA-NeXT-13B & 2024 & 13B \\
          & LLaVA-NeXT-34B & 2024 & 34B \\
        \midrule 

        \multirow[c]{7}{*}{Qwen-VL}
          & Qwen2.5-VL-3B-Instruct  & 2025 & 3B  \\
          & Qwen2.5-VL-7B-Instruct  & 2025 & 7B \\
          & Qwen2.5-VL-32B-Instruct & 2025 & 32B \\
          \cmidrule{2-4}
          & Qwen3-VL-8B-Instruct & 2025 & 8B \\
        \midrule

        \multirow[c]{3}{*}{TinyLLaVA}
          & TinyLLaVA-1.5B & 2024 & 1.5B \\
          & TinyLLaVA-2B & 2024 & 2B \\
          & TinyLLaVA-3.1B & 2024 & 3.1B \\
        \midrule

          Meta Llama 4 & Llama-4-Scout-17B-16E-Instruct & 2025 & 17B \\

        \midrule
          GLM-V & GLM-4.6V & 2025 & 106B \\
        \bottomrule
    \end{tabularx}
\end{table}

\textbf{InternVL}~\cite{chen2024expanding}. The InternVL family couples a high-capacity vision encoder with an InternLM language model and is trained on web-scale image-text data. We evaluate two generations, InternVL2.5 and InternVL3, covering a wide range of model sizes: InternVL2.5-2B, 4B, 8B, 26B and InternVL3-2B, 8B, 14B, 38B. In addition, we run InternVL3.5-38B as a newer high-capacity checkpoint in the series that targets stronger multimodal reasoning. For all InternVL models, we adopt the official OpenGVLab checkpoints and inference implementation without modification, provide each image and question in the recommended chat format, and keep the same unified CoT prompt to elicit natural-language reasoning.

\textbf{LLaVA-NeXT}~\cite{liu2024improved}. LLaVA-NeXT extends the LLaVA framework with updated language backbones and visual instruction tuning. We include four LLaVA-NeXT variants: LLaVA-NeXT-7B-Mistral, 8B, 13B, and 34B. All models are evaluated using the official checkpoints, which pair a CLIP-style vision encoder with different language models. 


\textbf{Qwen-VL}~\cite{bai2023qwen}. The Qwen-VL family includes both Qwen2.5-VL and Qwen3-VL, covering instruction-tuned vision-language models at multiple parameter scales. We assess three instruction-tuned Qwen2.5-VL variants: Qwen2.5-VL-3B-Instruct, Qwen2.5-VL-7B-Instruct, and Qwen2.5-VL-32B-Instruct. In addition, we evaluate Qwen3-VL-8B-Instruct as a representative model from the newer Qwen3-VL line. All models are loaded from the official Qwen releases, and inference follows the recommended chat template and image preprocessing pipeline.

\textbf{TinyLLaVA}~\cite{zhou2024tinyllava}. TinyLLaVA is designed as a lightweight alternative to large VLMs by attaching a compact vision encoder to a small language model, targeting resource-constrained scenarios such as edge devices. We test three publicly available versions, TinyLLaVA-1.5B, 2B, and 3.1B, obtained from the official project page. 

\textbf{Llama-4-Scout}~\cite{touvron2023llama}. We include Llama-4-Scout-17B-16E-Instruct as a recent large-scale open-source model with strong general reasoning ability. The Scout variant is a natively multimodal model that processes text and images via an integrated vision encoder and projection module in the official implementation. 

\textbf{GLM-V}~\cite{hong2025glm}. GLM-V is a family of multimodal models released by Z.ai that targets strong multimodal reasoning and long-context visual understanding. We evaluate GLM-4.6V (106B parameters) using the official checkpoint. We enable its thinking mode during inference, while keeping the same unified CoT prompt and the same sampling configuration (temperature=0.1, top-p=0.7) as in other models.


\subsection{LLM Evaluator}
\label{app:llm_evaluator}
To enable the automated and standardized evaluation of model-generated reasoning chains, we designed a structured prompt template, as shown in Figure~\ref{fig:prompt} (right). The prompt specifies:
\begin{itemize}
\item \textbf{Role and task definition}: The large language model acts as a professional evaluator, tasked with objectively scoring the reasoning chain across multiple dimensions.

\item \textbf{Evaluation content}: The subject of evaluation is the reasoning chain output generated by the model.

\item \textbf{Five evaluation dimensions and scoring criteria}: These include accuracy, logical coherence, effectiveness, clarity, and completeness, with detailed scoring guidelines for each dimension.

\item \textbf{Scoring system}: Each dimension is scored out of 10 points, with a total score of 50 points.

\item \textbf{Output format requirements}: The model is required to output the scores in a structured format, facilitating automated extraction and statistical analysis.
\end{itemize}
Through this standardized prompt design, we ensure that each evaluation adheres to a consistent set of standards, while also enabling the automated parsing of dimension scores and the total score from the model’s output.

To ensure the reproducibility and cost-effectiveness of the automatic scoring process, we selected open-source large language models as evaluators, all of which were obtained from the Hugging Face Hub and deployed locally using vLLM. The specific models used include: Qwen/Qwen3-8B, deepseek-ai/DeepSeek-R1-Distill-Qwen-7B, and zai-org/GLM-Z1-9B-0414. The generation parameters for each model are as follows:
\begin{itemize}
\item Qwen3-8B: enable\_thinking=True, temperature=0.6, max\_tokens=2048, \\top\_p=0.95, top\_k=20, min\_p=0.

\item DeepSeek-R1-Distill-Qwen-7B: temperature=0.6, max\_tokens=2048, \\top\_p=0.95.

\item GLM-Z1-9B-0414: temperature=0.6, max\_tokens=2048, top\_p=0.95, \\top\_k=40.
\end{itemize}
These parameter settings are designed to balance the diversity and stability of the generated results while ensuring the efficiency and reproducibility of the scoring process.

\subsection{Human Evaluation}
\label{app:human_evaluation}
To ensure consistency between human evaluations and the refinement of reasoning chains, we invited the same 20 volunteers described in Section~\ref{app:human_refinement} to participate in the manual scoring process. The evaluation criteria follow the five dimensions set in Figure~\ref{fig:prompt} (right). However, to avoid difficulties caused by a large score range, we adjusted the total score from the original 50-point scale to a 5-point scale, distinguishing it from the automatic scores generated by the LLMs. Finally, we calculated the mean and standard deviation of the scores from all volunteers, which served as the results for the human evaluation experiment.

\section{Main Results}
\label{app:main_result}
\subsection{Accuracy of the Final Answer}
We evaluate the final-answer accuracy of VLMs on AgroCoT, as shown in Fig.~\ref{fig:experiment}(a). Proprietary models generally perform better than open-source models, with GPT-5 achieving the highest accuracy above 60\%, followed by Gemini-2.5-Pro and GPT-4.1. Among open-source models, Qwen3-VL-8B and InternVL3-14B are competitive, both exceeding 55\%. Although most models outperform the 45\% baseline, none surpasses 70\%, indicating that AgroCoT remains challenging for current VLMs. The limited improvement of InternVL3.5 over InternVL3 further suggests that agricultural-domain reasoning is still under-optimized in mainstream VLM development. Additional ROUGE and BERTScore results are provided in Appendix~\ref{app:experimental_results}.

\subsection{Semantic Similarity and Key Steps Matching}
We evaluated the reasoning generation capabilities of the models based on semantic similarity and key steps matching, as shown in Table~\ref{tab:eval_results}. Semantic similarity measures the proximity between the model's output and the reference CoT, reflecting its ability to capture shallow semantics. GPT-4.1 outperformed all models with an average similarity score of 88.59, while Llama-4-Scout led the open-source models with a score of 84.53. GPT-4.1 performed best across all dimensions except for OD, where Claude-Sonnet-4.5 was slightly better. Among open-source models, InternVL3-14B and Llama-4-Scout showed comparable results. For key steps matching, the results aligned closely with semantic similarity scores. GPT-4.1 maintained its lead in overall CoT quality. Notably, InternVL3-14B had the strongest key steps coverage in QA and EM, indicating a higher overlap with the vocabulary of the standard answers in these tasks.

\subsection{LLM Scoring and Human Evaluation}
In the LLM Score evaluation presented in Table~\ref{tab:eval_results2}, Gemini-2.5-Pro leads with a score of 44.77, excelling in SU and DM dimensions, though slightly underperforming Doubao-Seed-1.8 in EM. Among open-source models, Qwen3-VL-8B and GLM-4.6V show similar performance, with Qwen3-VL-8B slightly leading overall. InternVL3-14B stands out in the SU dimension, showcasing its potential in spatial understanding for agriculture. Human evaluation results align with LLM scores, with Qwen3-VL-8B ranking highest among open-source models (3.94), followed by GLM-4.6V (3.83). Gemini-2.5-pro leads in closed-source models with 3.81, confirming its strength in deep reasoning. These findings suggest that combining automatic and human evaluation offers a comprehensive view of model performance in complex agricultural reasoning tasks. In addition, we provide representative case studies in the Appendix~\ref{app:dimension_details}, which further demonstrate that AgroCoT can effectively evaluate model applicability in practical agricultural tasks and highlight AgroCoT’s integration of visual information and reasoning challenges.

\section{More Experimental Results}
\label{app:experimental_results}
\subsection{Accuracy, ROUGE-L F1 and BERTScore F1 Evaluation}
\label{app:accuracy_rouge_bert}
\begin{table*}[ht]
\caption{Comprehensive performance evaluation of various VLMs. Acc for \textbf{Accuracy} (\%); R-L-f for \textbf{ROUGE-L F1} (\%); B-f for \textbf{BERTScore F1} (\%). ``-'' indicates that the model is unable to output a normal CoT reasoning, resulting in a negative BERTScore. \textbf{Boldface} indicates the best overall across all models. \colorbox[rgb]{1.0, 0.894, 0.812}{Peach} and \colorbox[rgb]{0.996, 1.0, 0.835}{Lemon} denote the best and the second-best among open-source models, respectively.}
\label{tab:eval_results3}
\centering
\resizebox{\textwidth}{!}{%
\begin{tabular}{cccc|ccc|ccc|ccc|ccc|ccc}
\Xhline{1.5pt}
\cellcolor[HTML]{FFFFFF}  
& \multicolumn{3}{c}{\cellcolor[HTML]{FFFFFF}\textbf{QA}} & \multicolumn{3}{c}{\cellcolor[HTML]{FFFFFF}\textbf{SU}} & \multicolumn{3}{c}{\cellcolor[HTML]{FFFFFF}\textbf{OD}} 
& \multicolumn{3}{c}{\cellcolor[HTML]{FFFFFF}\textbf{DM}} & \multicolumn{3}{c}{\cellcolor[HTML]{FFFFFF}\textbf{EM}} & \multicolumn{3}{c}{\cellcolor[HTML]{FFFFFF}\textbf{Overall}} \\
\multirow{-2}{*}{\cellcolor[HTML]{FFFFFF}\textbf{Model}} & \textbf{Acc} & \textbf{R-L-f} & \textbf{B-f} & \textbf{Acc}  & \textbf{R-L-f} & \textbf{B-f} & \textbf{Acc} & \textbf{R-L-f} & \textbf{B-f} & \textbf{Acc} & \textbf{R-L-f} & \textbf{B-f} & \textbf{Acc} & \textbf{R-L-f} & \textbf{B-f}& \textbf{Acc} & \textbf{R-L-f} & \textbf{B-f} \\ 
\Xhline{1.2pt}
\multicolumn{19}{c}{\cellcolor[rgb]{0.906, 0.898, 0.902}\textbf{Proprietary Large Vision-Language Models}} \\                                                                                                                                              
Claude-Sonnet-4.5                   & 44.22 & 47.09 & 48.44 & 45.20 & 41.70 & 43.89 & 66.34 & 45.22 & 45.85 & 65.82 & 45.95 & 45.35 & 70.09 & 44.77 & 44.15 & 55.83 & 44.92 & 45.74 \\

Gemini-2.5-pro~\cite{team2023gemini}                      & 42.72 & 36.39 & 35.41 & 49.64 & 32.65 & 30.95 & 66.99 & 35.16 & 32.26 & 71.11 & 36.77 & 33.02 & \textbf{84.15} & 36.21 & 32.28 & 59.14 & 35.27 & 32.88 \\

GPT-4.1~\cite{achiam2023gpt}                             & 41.75 & \textbf{50.99} & \textbf{52.11} & 49.02 & \textbf{45.93} & \textbf{48.43} & 69.38 & \textbf{47.25} & \textbf{48.28} & 72.11 & \textbf{49.79} & \textbf{49.11} & 79.46 & \textbf{46.62} & \textbf{45.66} & 58.96 & \textbf{48.31} & \textbf{49.18} \\

GPT-5~\cite{leon2025gpt}                               & \textbf{50.66} & 30.46 & 30.42 & \textbf{54.26} & 27.65 & 29.55 & 70.03 & 27.76 & 26.21 & \textbf{76.51} & 29.27 & 27.92 & 79.68 & 28.46 & 28.83 & \textbf{63.52} & 28.78 & 28.69 \\ 

\Xhline{1.2pt}
\multicolumn{19}{c}{\cellcolor[rgb]{0.906, 0.898, 0.902}\textbf{Open-Source Large Vision-Language Models}} \\

DeepSeek-VL2~\cite{lu2024deepseek}                        & 27.62 & 16.18 & -     & 15.18 &  8.63 & -     & 20.41 & 11.92 & -     & 19.51 & 11.09 & -     & 28.79 &  6.52  & -     & 21.56  & 11.47 & -       \\
DeepSeek-VL2-small~\cite{lu2024deepseek}                  & 34.59 & 16.92 & -     & 16.34 &  7.87 & -     & 30.61 & 12.31 & -     & 27.67 & 14.86 & -     & 30.13 & 11.93  & -     & 27.43  & 12.83 & -       \\
DeepSeek-VL2-tiny~\cite{lu2024deepseek}                   & 18.01 & 16.79 & 7.81  & 15.89 &  7.24 & -     & 23.56 & 12.55 & -     & 27.34 & 10.38 & -     & 32.14 &  5.42  & -     & 21.87  & 11.15 & -       \\

InternVL2-2B~\cite{chen2024expanding}                        & 36.80 & 29.02 & 25.68 & 16.78 & 28.64 & 25.52 & 29.21 & 29.68 & 24.49 & 29.20 & 29.02 & 23.78 & 32.14 & 27.89  & 20.07 & 28.31  & 28.95 & 24.46   \\
InternVL2-4B~\cite{chen2024expanding}                        & 22.86 & 40.25 & 39.35 & 23.09 & 34.98 & 34.63 & 36.59 & 33.95 & 30.67 & 40.13 & 36.71 & 34.07 & 42.41 & 34.62  & 31.66 & 31.09  & 36.40 & 34.60   \\
InternVL2-8B~\cite{chen2024expanding}                        & 38.04 & 38.08 & 38.72 & 39.96 & 38.67 & \cellcolor[rgb]{0.996, 1.0, 0.835}38.04 & 55.37 & 37.03 & 36.41 & 48.51 & 38.45 & 35.95 & 50.89 & 38.35  & \cellcolor[rgb]{0.996, 1.0, 0.835}35.80 & 45.40  & 38.11 & 37.24   \\
InternVL2-26B~\cite{chen2024expanding}                       & 42.81 & 38.62 & 38.74 & 37.21 & \cellcolor[rgb]{1.0, 0.894, 0.812}
40.49 & \cellcolor[rgb]{1.0, 0.894, 0.812}
41.24 & 55.70 & 37.72 & 35.52 & 55.46 & 38.81 & \cellcolor[rgb]{0.996, 1.0, 0.835}37.43 & 67.41 & 37.47  & 35.73 & 49.00  & 38.82 & \cellcolor[rgb]{0.996, 1.0, 0.835}38.15   \\

InternVL3-2B~\cite{chen2024expanding}                        & 34.77 & 39.73 & 37.54 & 33.84 & 33.70 & 25.88 & 41.59 & 34.62 & 26.21 & 30.65 & 34.72 & 28.25 & 43.75 & 37.04  & 30.67 & 35.99  & 35.93 & 29.81   \\
InternVL3-8B~\cite{chen2024expanding}                        & 35.48 & 42.63 & 35.05 & 40.76 & \cellcolor[rgb]{0.996, 1.0, 0.835}39.19 & 27.60 & \cellcolor[rgb]{0.996, 1.0, 0.835}63.08 & 38.81 & 30.94 & 56.01 & 39.46 & 28.35 & 58.26 & 40.71  & 28.83 & 48.75  & 40.18 & 30.41   \\
InternVL3-14B~\cite{chen2024expanding}                       &  \cellcolor[rgb]{0.996, 1.0, 0.835}48.72 & \cellcolor[rgb]{1.0, 0.894, 0.812}
46.23 & 38.44 & \cellcolor[rgb]{0.996, 1.0, 0.835}44.67 & 38.08 & 27.34 & 62.32 & \cellcolor[rgb]{1.0, 0.894, 0.812}
42.01 & 34.51 & 63.84 & \cellcolor[rgb]{0.996, 1.0, 0.835}40.98 & 33.71 & \cellcolor[rgb]{0.996, 1.0, 0.835}69.64 & \cellcolor[rgb]{0.996, 1.0, 0.835}40.82  & 32.86 & \cellcolor[rgb]{0.996, 1.0, 0.835}55.57  & \cellcolor[rgb]{0.996, 1.0, 0.835}41.77 & 33.39   \\

InternVL3-38B~\cite{chen2024expanding}                       & 46.69 & 44.29 & \cellcolor[rgb]{0.996, 1.0, 0.835}40.51 & \cellcolor[rgb]{1.0, 0.894, 0.812}
45.74 & 38.50 & 36.89 & \cellcolor[rgb]{1.0, 0.894, 0.812}
\textbf{72.64} & 39.15 & \cellcolor[rgb]{0.996, 1.0, 0.835}36.50 & \cellcolor[rgb]{1.0, 0.894, 0.812}
66.48 & 39.17 & 35.17 & \cellcolor[rgb]{1.0, 0.894, 0.812}
71.43 & 38.95  & 35.34 & \cellcolor[rgb]{1.0, 0.894, 0.812}
58.13  & 40.25 & 37.22   \\

Llama-4-Scout-17B-16E-Instruct~\cite{touvron2023llama}      & 40.69 & \cellcolor[rgb]{0.996, 1.0, 0.835}44.90 & \cellcolor[rgb]{1.0, 0.894, 0.812}
43.63 & 40.05 & 37.88 & 36.87 & 61.78 & \cellcolor[rgb]{0.996, 1.0, 0.835}41.35 & \cellcolor[rgb]{1.0, 0.894, 0.812}
40.31 & \cellcolor[rgb]{0.996, 1.0, 0.835}65.05 & \cellcolor[rgb]{1.0, 0.894, 0.812}
43.40 & \cellcolor[rgb]{1.0, 0.894, 0.812}
41.18 & 64.73 & \cellcolor[rgb]{1.0, 0.894, 0.812}
41.34  & \cellcolor[rgb]{1.0, 0.894, 0.812}
38.65 & 52.06  & \cellcolor[rgb]{1.0, 0.894, 0.812}
41.78 & \cellcolor[rgb]{1.0, 0.894, 0.812}
40.29   \\

LLaVA-NeXT-7B-Mistral~\cite{liu2024improved}               & 27.54 & 32.36 & 29.52 & 14.56 & 31.89 & 27.74 & 22.58 & 30.79 & 24.81 & 24.81 & 32.87 & 27.85 & 39.29 & 31.05  & 28.29 & 23.93  & 31.90 & 27.67   \\
LLaVA-NeXT-8B~\cite{liu2024improved}                       & 38.57 & 34.56 & 34.54 & 23.53 & 37.15 & 35.48 & 35.72 & 34.05 & 29.77 & 41.35 & 35.91 & 33.81 & 38.39 & 33.81  & 30.99 & 34.80  & 35.29 & 33.31   \\
LLaVA-NeXT-13B~\cite{liu2024improved}                      & 10.06 & 23.45 & 10.55 & 13.85 & 26.92 & 19.86 & 12.05 & 25.79 & 17.14 & 15.21 & 26.95 & 16.62 & 17.86 & 25.99  & 17.45 & 13.21  & 25.74 & 16.09   \\
LLaVA-NeXT-34B~\cite{liu2024improved}                      &  \cellcolor[rgb]{1.0, 0.894, 0.812}49.60 & 32.13 & 26.42 & 29.48 & 34.04 & 32.48 & 50.71 & 33.12 & 31.82 & 39.03 & 33.19 & 29.76 & 46.65 & 33.98  & 32.24 & 42.43  & 33.20 & 30.27   \\

Qwen2.5-VL-3B-Instruct~\cite{bai2023qwen}              & 33.27 & 32.45 & 31.65 & 38.28 & 30.70 & 27.49 & 61.45 & 29.54 & 24.75 & 57.33 & 29.66 & 24.43 & 50.00 & 31.80  & 26.16 & 46.70  & 30.80 & 27.23   \\
Qwen2.5-VL-7B-Instruct~\cite{bai2023qwen}              & 38.75 & 36.28 & 34.09 & 35.70 & 33.38 & 30.63 & 63.08 & 31.89 & 25.82 & 57.99 & 32.91 & 26.76 & 52.68 & 31.30  & 25.94 & 48.16  & 35.50 & 29.28   \\
Qwen2.5-VL-32B-Instruct~\cite{bai2023qwen}             & 33.80 & 36.73 & 33.13 & 15.28 & 33.33 & 30.11 & 16.07 & 33.85 & 28.36 & 17.42 & 35.05 & 27.62 & 23.21 & 36.11  & 30.38 & 21.28  & 34.90 & 29.54   \\
TinyLLaVA-1.5B~\cite{zhou2024tinyllava}                      & 19.51 & 18.29 &  9.28 & 14.56 &  7.72 & -     & 17.59 & 11.16 & -     & 25.58 &  8.60 & -     & 21.21 & 10.32  & -     & 19.27  & 11.49 & -       \\
TinyLLaVA-2B~\cite{zhou2024tinyllava}                        & 34.51 &  3.52 & -     & 22.82 &  5.71 & -     & 45.17 &  5.28 & -     & 46.64 &  3.46 & -     & 41.07 &  4.85  & -     & 36.85  &  4.54 & -       \\
TinyLLaVA-3.1B~\cite{zhou2024tinyllava}                      & 33.45 & 16.06 &  4.76 & 18.12 & 12.65 &  0.93 & 41.91 & 11.58 & -     & 40.35 & 10.40 & -     & 48.66 & 13.51  &  2.41 & 34.24  & 12.92 &  0.37   \\

\Xhline{1.5pt}
\end{tabular}%
}
\end{table*}
As shown in Table~\ref{tab:eval_results3}, proprietary models generally outperform their open-source counterparts in overall accuracy, reflecting their direct question-answering capabilities. GPT-5 achieves the highest accuracy, surpassing Gemini-2.5-pro by 4.38\%, with significant improvements in QA, SU, and DM. InternVL3-38B leads in Object Detection, while Gemini-2.5-pro excels in EU. Among open-source models, the InternVL3 family and Llama-4-Scout stand out as strong performers. Notably, Qwen2.5-VL performs well at 3B and 7B scales but experiences a sharp decline at 32B. Across dimensions, most models notably lag in QA and SU, highlighting the challenges in precision computation and spatial reasoning.

ROUGE-L-f and BERTScore-f assess the reasoning capabilities of models through rule-based and semantic matching, respectively. GPT-4.1 achieves the best overall performance on both metrics, particularly excelling in the Quantitative Analysis dimension. In contrast, despite its high accuracy, GPT-5 demonstrates notably weaker reasoning ability than several other models, including some open-source ones. Although the DeepSeek-VL and TinyLLaVA series show low ROUGE scores, this does not necessarily imply reliable reasoning; their consistently low BERTScores indicate incoherent reasoning outputs and poor semantic alignment. Among open-source models, the InternVL2, InternVL3, and Llama-4-Scout series exhibit the strongest reasoning capabilities, outperforming both Gemini-2.5-pro and GPT-5 by a significant margin.

\subsection{Keyword Extraction Capability}
\label{app:keyword_extraction_capability}
The ROUGE-1 scores presented in Table~\ref{tab:app_rouge1} provide insights into the reasoning capabilities of the evaluated models across various tasks. ROUGE-1, which measures the overlap of unigrams between the generated CoT and the reference, reflects the model's ability to generate keywords in reasoning texts. Across all tasks, proprietary models like Gemini-2.5-pro and GPT-4.1 perform exceptionally well in recall and F1 score, indicating their strong ability to retain key word in the reasoning process. InternVL2-26B achieves the highest overall precision score, indicating that it generates fewer but more accurate keywords compared to other models. However, the scores of these models across different task types showed minimal variation, suggesting that the models demonstrate consistent keyword reasoning capabilities across all task categories. With the exception of Gemini-2.5-pro, nearly all other models exhibit lower recall compared to precision. This suggests that while the models generate accurate reasoning in certain sections, as reflected in their high precision, there are gaps in the depth of reasoning across longer texts, leading to missed or incomplete coverage when compared to the reference text.

\subsection{Coherence and Fluency of Reasoning Texts}
\label{app:coherence_and_fluency}
\begin{table*}[t]
\caption{Additional performance evaluation of VLMs using \textbf{ROUGE-1: Precision, Recall, and F1 Score}. \textbf{Boldface} indicates the best overall across all models. \colorbox[rgb]{1.0, 0.894, 0.812}{Peach} and \colorbox[rgb]{0.996, 1.0, 0.835}{Lemon} denote the best and the second-best among open-source models, respectively.}
\label{tab:app_rouge1}
\centering
\resizebox{\textwidth}{!}{%
\begin{tabular}{c|ccc|ccc|ccc|ccc|ccc|ccc}
\Xhline{1.5pt}
\cellcolor[HTML]{FFFFFF}  
& \multicolumn{3}{c}{\cellcolor[HTML]{FFFFFF}\textbf{QA}} & \multicolumn{3}{c}{\cellcolor[HTML]{FFFFFF}\textbf{SU}} & \multicolumn{3}{c}{\cellcolor[HTML]{FFFFFF}\textbf{OD}} 
& \multicolumn{3}{c}{\cellcolor[HTML]{FFFFFF}\textbf{DM}} & \multicolumn{3}{c}{\cellcolor[HTML]{FFFFFF}\textbf{EM}} & \multicolumn{3}{c}{\cellcolor[HTML]{FFFFFF}\textbf{Overall}} \\
\multirow{-2}{*}{\cellcolor[HTML]{FFFFFF}\textbf{Model}}  & \textbf{Precision} & \textbf{Recall} & \textbf{F1} & \textbf{Precision} & \textbf{Recall} & \textbf{F1} & \textbf{Precision} & \textbf{Recall} & \textbf{F1} & \textbf{Precision} & \textbf{Recall} & \textbf{F1} & \textbf{Precision} & \textbf{Recall} & \textbf{F1} & \textbf{Precision} & \textbf{Recall} & \textbf{F1} \\ 
\Xhline{1.2pt}
\multicolumn{19}{c}{\cellcolor[rgb]{0.906, 0.898, 0.902}\textbf{Proprietary Large Vision-Language Models}} \\                                                                                                                                      
Claude-Sonnet-4.5                          & 49.18 & 52.21 & 50.00 & 44.56 & 46.27 & 44.71 & 49.38 & 48.84 & 48.45 & 50.85 & 48.38 & 49.04 & 49.96 & 46.91 & 47.81 & 48.48 & 48.76 & 47.96 \\

Gemini-2.5-pro~\cite{team2023gemini}       & 30.76 & \textbf{53.70} & 38.58 & 27.40 & \textbf{49.90} & 34.85 & 30.61 & \textbf{50.47} & 37.70 & 32.34 & \textbf{51.80} & 39.34 & 32.39 & \textbf{50.36} & 38.98 & 30.37 & \textbf{51.39} & 37.67 \\

GPT-4.1~\cite{achiam2023gpt}               & 55.93 & 52.91 &\textbf{53.71} & 53.82 & 46.37 & \textbf{49.27} & 56.93 & 46.33 & \textbf{50.39} & 60.48 & 47.82 & \textbf{52.82} & 57.65 & 44.67 & \textbf{49.69} & 56.69 & 48.12 & \textbf{51.36}\\

GPT-5~\cite{leon2025gpt}                   & 50.64 & 25.09 & 32.79 & 47.20 & 22.36 & 29.99 & 50.01 & 22.07 & 30.14 & 52.60 & 22.97 & 31.52 & 54.19 & 21.85 & 30.64 & 50.40 & 23.05 & 31.09 \\ 

\Xhline{1.2pt}
\multicolumn{19}{c}{\cellcolor[rgb]{0.906, 0.898, 0.902}\textbf{Open-Source Large Vision-Language Models}} \\

DeepSeek-VL2~\cite{lu2024deepseek}         & 43.50 & 11.44 & 16.61 & 24.93 & 6.65  & 9.10  & 37.72 & 10.10 & 12.75 & 53.34  & 9.57 & 11.65  & 48.14 & 4.47  & 6.81 & 40.14 & 8.94  & 11.99    \\
DeepSeek-VL2-small~\cite{lu2024deepseek}   & 49.92 & 12.08 & 17.61 & 35.91 & 5.38  & 8.26  & 48.34 & 9.39  & 13.06 & 46.11  & 11.21 & 15.77 & 50.61 & 8.45  & 12.52 & 45.42 & 9.34 & 13.49     \\
DeepSeek-VL2-tiny~\cite{lu2024deepseek}    & 48.17 & 11.81 & 17.59 & 30.25 & 6.04  & 7.91  & 45.47 & 10.02 & 13.68 & 56.96  & 8.51  & 11.11 & 62.15 & 4.31  & 5.89  & 46.31 & 8.61  & 11.94     \\

InternVL2-2B~\cite{chen2024expanding}      & 57.98 & 23.44 & 30.91 & 47.11 & 24.82 & 30.90 & 56.98 & 23.25 & 31.74 & 55.28 & 23.42 & 31.15 & 53.08 & 22.91 & 29.70 & 54.05 & 23.69 & 31.00      \\
InternVL2-4B~\cite{chen2024expanding}      & 55.54 & 36.25 & 42.72 & 55.53 & 29.95 & 37.57 & 54.96 & 28.93 & 36.70 & 57.46 & 31.35 & 39.52 & 55.61 & 29.02 & 37.22 & 55.81 & 31.50 & 39.04       \\
InternVL2-8B~\cite{chen2024expanding}      & 60.50 & 31.78 & 40.43 & 54.15 & 34.82 & 41.32 & 55.33 & 32.08 & 39.68 & 57.88 & 32.75 & 41.11 & 55.79 & 32.89 & 40.53 & 56.88 & 32.90 & 40.65       \\
InternVL2-26B~\cite{chen2024expanding}     & 61.79 & 31.56 & 40.60 & 60.26 & 33.99 & \colorbox[rgb]{1.0, 0.894, 0.812}{42.67} & \colorbox[rgb]{0.996, 1.0, 0.835}{62.04} & 30.80 & 40.03 & \colorbox[rgb]{1.0, 0.894, 0.812}{\textbf{62.90}} & 31.41 & 41.06 & \colorbox[rgb]{0.996, 1.0, 0.835}{63.52} & 29.66 & 39.67 & \colorbox[rgb]{1.0, 0.894, 0.812}{\textbf{61.85}} & 31.79 & 40.99      \\

InternVL3-2B~\cite{chen2024expanding}      & 55.09 & 35.11 & 41.79 & 52.14 & 28.99 & 35.72 & 54.76 & 29.33 & 37.19 & 56.89 & 28.45 & 36.73 & 55.66 & 31.67 & 39.30 & 54.71 & 30.74 & 38.09       \\
InternVL3-8B~\cite{chen2024expanding}      & 56.12 & 38.98 & 45.11 & 51.79 & 35.89 & \colorbox[rgb]{0.996, 1.0, 0.835}{41.51} & 54.79 & 34.80 & 41.61 & 55.91 & 34.83 & 42.10 & 56.02 & 36.20 & 43.28 & 54.72 & 36.26 & 42.72       \\
InternVL3-14B~\cite{chen2024expanding}     & 57.92 & 43.65 & \colorbox[rgb]{1.0, 0.894, 0.812}{48.78} & 53.03 & 34.10 & 40.55 & 57.01 & 38.09 & \colorbox[rgb]{1.0, 0.894, 0.812}{45.00} & 57.03 & 36.45 & \colorbox[rgb]{0.996, 1.0, 0.835}{43.78} & 55.70 & 36.96 & \colorbox[rgb]{0.996, 1.0, 0.835}{43.61} & 56.12 & 38.05 & \colorbox[rgb]{0.996, 1.0, 0.835}{44.46}        \\
InternVL3-38B~\cite{chen2024expanding}     & \colorbox[rgb]{0.996, 1.0, 0.835}{62.27} & 38.40 & 46.60 & 54.76 & 33.83 & 41.15 & 56.68 & 33.93 & 41.83 & 57.36 & 33.75 & 41.88 & 57.55 & 33.53 & 41.74 & 57.82 & 34.95 & 42.85         \\

Llama-4-Scout-17B-16E-Instruct~\cite{touvron2023llama}   & 46.79 & \colorbox[rgb]{1.0, 0.894, 0.812}{51.42} & \colorbox[rgb]{0.996, 1.0, 0.835}{47.52} & 38.98 & \colorbox[rgb]{1.0, 0.894, 0.812}{46.64} & 40.62 & 41.54 & \colorbox[rgb]{1.0, 0.894, 0.812}{50.28} & \colorbox[rgb]{0.996, 1.0, 0.835}{44.50} & 46.44 & \colorbox[rgb]{1.0, 0.894, 0.812}{48.65} & \colorbox[rgb]{1.0, 0.894, 0.812}{46.70} & 45.28 & \colorbox[rgb]{1.0, 0.894, 0.812}{45.87} & \colorbox[rgb]{1.0, 0.894, 0.812}{44.44} & 43.57 & \colorbox[rgb]{1.0, 0.894, 0.812}{48.90} & \colorbox[rgb]{1.0, 0.894, 0.812}{44.73}   \\

LLaVA-NeXT-7B-Mistral~\cite{liu2024improved} & 62.04 & 25.87 & 34.84 & 59.74 & 24.63 & 33.82 & 57.74 & 24.55 & 33.25 & 59.33 & 25.89 & 35.37 & 61.58 & 23.06 & 32.77 & \colorbox[rgb]{0.996, 1.0, 0.835}{60.01} & 25.02 & 34.16              \\
LLaVA-NeXT-8B~\cite{liu2024improved}         & 56.37 & 29.06 & 36.47 & \colorbox[rgb]{0.996, 1.0, 0.835}{61.49} & 28.71 & 38.53 & \colorbox[rgb]{1.0, 0.894, 0.812}{\textbf{62.59}} & 26.05 & 36.18 & 60.40 & 28.74 & 38.06 & 59.29 & 26.18 & 35.85 & 59.99 & 28.01 & 37.18                  \\
LLaVA-NeXT-13B~\cite{liu2024improved}        & 35.86 & 20.71 & 25.31 & 44.28 & 22.04 & 28.92 & 50.02 & 20.97 & 27.54 & 43.29 & 22.34 & 28.39 & 46.32 & 20.75 & 27.72 & 43.34 & 21.42 & 27.51               \\
LLaVA-NeXT-34B~\cite{liu2024improved}        & 49.79 & 27.24 & 34.42 & \colorbox[rgb]{1.0, 0.894, 0.812}{\textbf{62.73}} & 25.75 & 35.89 & 58.26 & 26.25 & 35.50 & \colorbox[rgb]{0.996, 1.0, 0.835}{61.11} & 25.79 & 35.42 & 59.85 & 25.98 & 35.74 & 57.98 & 26.25 & 35.33             \\

Qwen2.5-VL-3B-Instruct~\cite{bai2023qwen}    & 53.68 & 26.48 & 34.41 & 43.91 & 29.11 & 33.37 & 52.56 & 24.40 & 32.37 & 49.69 & 25.15 & 32.52 & 48.57 & 28.53 & 34.64 & 49.72 & 26.65 & 33.38           \\
Qwen2.5-VL-7B-Instruct~\cite{bai2023qwen}    & 54.11 & 31.67 & 38.71 & 48.49 & 32.17 & 36.35 & 54.23 & 27.49 & 34.94 & 54.12 & 28.05 & 35.96 & 56.40 & 26.58 & 33.94 & 52.97 & 29.72 & 36.34          \\
Qwen2.5-VL-32B-Instruct~\cite{bai2023qwen}   & 40.35 & \colorbox[rgb]{0.996, 1.0, 0.835}{44.97} & 39.47 & 32.96 & \colorbox[rgb]{0.996, 1.0, 0.835}{40.88} & 35.69 & 35.94 & \colorbox[rgb]{0.996, 1.0, 0.835}{41.24} & 36.50 & 36.26 & \colorbox[rgb]{0.996, 1.0, 0.835}{43.72} & 37.52 & 36.96 & \colorbox[rgb]{0.996, 1.0, 0.835}{42.28} & 38.54 & 36.46 & \colorbox[rgb]{0.996, 1.0, 0.835}{42.68} & 37.45        \\

TinyLLaVA-1.5B~\cite{zhou2024tinyllava}      & \colorbox[rgb]{1.0, 0.894, 0.812}{\textbf{72.82}} & 11.27 & 18.88 & 41.17 &  5.01 &  7.89 & 56.20 &  7.40 & 11.65 & 55.21 &  5.14 &  8.85 & 63.05 &  6.08 & 10.53 & 57.10 &  7.19 & 11.85          \\
TinyLLaVA-2B~\cite{zhou2024tinyllava}        & 38.74 &  2.08 &  3.57 & 33.76 &  3.52 &  5.96 & 38.47 &  3.52 &  5.76 & 36.69 &  2.06 &  3.57 & \colorbox[rgb]{1.0, 0.894, 0.812}{\textbf{64.23}} &  2.81 &  5.07 & 39.56 &  2.80 & 4.75                    \\
TinyLLaVA-3.1B~\cite{zhou2024tinyllava}      & 61.30 & 10.70 & 16.62 & 48.52 &  8.58 & 13.02 & 48.00 &  7.62 & 12.04 & 50.28 &  6.50 & 10.77 & 60.55 &  8.51 & 13.94 & 53.15 &  8.49 & 13.36                 \\

\Xhline{1.5pt}
\end{tabular}
}
\end{table*}
ROUGE-2 primarily evaluates the overlap of bigrams between the model's generated reasoning and the reference, which is closely tied to the coherence and fluency of the reasoning text. A higher ROUGE-2 score indicates better alignment with the reference in terms of more complex word pairings, reflecting the model’s ability to maintain coherent and logical connections in its reasoning steps. Compared to ROUGE-1 in Table~\ref{tab:app_rouge1}, all ROUGE-2 scores show a noticeable decline, indicating that all models exhibit gaps in reasoning fluency, particularly in terms of maintaining coherent and logically connected reasoning chains. As illstrated in Table~\ref{tab:app_rouge2}, GPT-4.1 exhibits the highest ROUGE-2 scores across most tasks, particularly in DM and EM. This suggests that GPT-4.1 generates more fluent and coherent reasoning chains, with well-structured and logically connected word pairs. It's higher ROUGE-2 scores indicate that reasoning chains tend to follow a more natural flow, with fewer fragmented parts. In contrast, TinyLLaVA-2B shows lower ROUGE-2 scores, indicating that the reasoning produced by it may be less fluent, with poorer coherence between consecutive terms in the generated text. These models with low ROUGE-2 scores appear to generate reasoning chains that lack depth and fluency, often resulting in disjointed or incomplete logical connections. 
\begin{table*}[ht]
\caption{Additional performance evaluation of VLMs using\textbf{ ROUGE-2: Precision, Recall, and F1 Score}. \textbf{Boldface} indicates the best overall across all models. \colorbox[rgb]{1.0, 0.894, 0.812}{Peach} and \colorbox[rgb]{0.996, 1.0, 0.835}{Lemon} denote the best and the second-best among open-source models, respectively.}
\label{tab:app_rouge2}
\centering
\resizebox{\textwidth}{!}{%
\begin{tabular}{c|ccc|ccc|ccc|ccc|ccc|ccc}
\Xhline{1.5pt}
\cellcolor[HTML]{FFFFFF}  
& \multicolumn{3}{c}{\cellcolor[HTML]{FFFFFF}\textbf{QA}} & \multicolumn{3}{c}{\cellcolor[HTML]{FFFFFF}\textbf{SU}} & \multicolumn{3}{c}{\cellcolor[HTML]{FFFFFF}\textbf{OD}} 
& \multicolumn{3}{c}{\cellcolor[HTML]{FFFFFF}\textbf{DM}} & \multicolumn{3}{c}{\cellcolor[HTML]{FFFFFF}\textbf{EM}} & \multicolumn{3}{c}{\cellcolor[HTML]{FFFFFF}\textbf{Overall}} \\
\multirow{-2}{*}{\cellcolor[HTML]{FFFFFF}\textbf{Model}}  & \textbf{Precision} & \textbf{Recall} & \textbf{F1} & \textbf{Precision} & \textbf{Recall} & \textbf{F1} & \textbf{Precision} & \textbf{Recall} & \textbf{F1} & \textbf{Precision} & \textbf{Recall} & \textbf{F1} & \textbf{Precision} & \textbf{Recall} & \textbf{F1} & \textbf{Precision} & \textbf{Recall} & \textbf{F1} \\ 
\Xhline{1.2pt}
\multicolumn{19}{c}{\cellcolor[rgb]{0.906, 0.898, 0.902}\textbf{Proprietary Large Vision-Language Models}} \\                                                                                                                                      
Claude-Sonnet-4.5                                   & 24.64 & 26.11 & 24.94 & 19.37 & 19.67 & 19.11 & 22.89 & 22.20 & 22.13 & 22.91 & 21.17 & 21.68 & 22.19 & 19.76 & 20.54 & 22.39 & 21.10 & 21.83 \\

Gemini-2.5-pro~\cite{team2023gemini}                & 11.85 & 23.31 & 15.45 & 9.13  & 18.51 & 11.99 & 11.19 & 20.22 & 14.20 & 11.73 & 20.22 & 14.62 & 11.78 & 19.31 & 14.40 & 11.01 & 20.48 & 14.07 \\

GPT-4.1~\cite{achiam2023gpt}                        & 31.23 & \textbf{29.40} & \textbf{29.80} & 28.24 & \textbf{23.13} & \textbf{25.05} & 31.12 & \textbf{24.14} & \textbf{26.65} & \textbf{34.49} & \textbf{25.80} & \textbf{29.03} & 30.23 & \textbf{21.64} & \textbf{24.74} & \textbf{31.02} & \textbf{25.29} & \textbf{27.33} \\

GPT-5~\cite{leon2025gpt}                            & 15.39 &  6.26 &  8.64 & 14.55 &  5.54 &  7.91 & 14.50 &  5.14 &  7.44 & 15.58 &  5.38 &  7.85 & 18.59 &  5.60 &  8.43 & 15.35 &  5.61 & 8.04  \\ 

\Xhline{1.2pt}
\multicolumn{19}{c}{\cellcolor[rgb]{0.906, 0.898, 0.902}\textbf{Open-Source Large Vision-Language Models}} \\

DeepSeek-VL2~\cite{lu2024deepseek}                  & 15.79 &  4.68 &  6.78 &  5.73 &  2.40 &  3.17 &  8.74 &  3.50 &  4.23 & 7.79  &  3.36 &  3.93 & 11.75 &  1.32 &  1.88 &  9.86 &  3.28 & 4.31          \\
DeepSeek-VL2-small~\cite{lu2024deepseek}            & 11.48 &  3.86 &  5.43 &  6.58 &  1.62 &  2.37 & 13.82 &  2.85 &  3.91 & 12.16 &  3.52 &  4.86 & 20.39 &  2.63 &  3.89 & 11.75 &  2.91 & 4.09          \\
DeepSeek-VL2-tiny~\cite{lu2024deepseek}             & 21.58 &  4.54 &  7.09 &  8.74 &  1.89 &  2.51 & 10.24 &  2.76 &  3.96 & 11.98 &  2.35 &  3.15 &  2.24 &  1.03 &  1.33 & 12.26 &  2.73 & 3.96        \\

InternVL2-2B~\cite{chen2024expanding}               & 26.95 & 10.45 & 13.51 & 18.05 &  9.33 & 11.47 & 21.91 &  7.65 & 10.71 & 23.17 &  8.58 & 11.57 & 19.79 &  8.34 & 10.86 & 22.25 &  9.02 & 11.78      \\
InternVL2-4B~\cite{chen2024expanding}               & 26.19 & 16.84 & 19.75 & 23.61 & 12.13 & 15.23 & 21.58 & 10.90 & 13.83 & 23.37 & 12.32 & 15.55 & 22.87 & 11.20 & 14.49 & 23.72 & 13.00 & 16.07      \\
InternVL2-8B~\cite{chen2024expanding}               & 28.03 & 13.79 & 17.72 & 22.72 & 14.37 & 16.96 & 22.41 & 12.47 & 15.54 & 24.29 & 12.91 & 16.44 & 24.44 & 13.47 & 16.86 & 24.44 & 13.46 & 16.75       \\
InternVL2-26B~\cite{chen2024expanding}              & 29.72 & 14.70 & 19.05 & 28.24 & 15.22 & \colorbox[rgb]{1.0, 0.894, 0.812}{19.31} & \colorbox[rgb]{0.996, 1.0, 0.835}{26.70} & 12.52 & 16.42 & \colorbox[rgb]{1.0, 0.894, 0.812}{28.91} & 13.45 & 17.87 & \colorbox[rgb]{0.996, 1.0, 0.835}{29.61} & 12.44 & 17.04 & \colorbox[rgb]{1.0, 0.894, 0.812}{28.57} & 13.91 & 18.15  \\

InternVL3-2B~\cite{chen2024expanding}               & 23.56 & 14.47 & 17.37 & 21.24 & 11.28 & 13.88 & 20.13 & 10.06 & 12.92 & 21.70 &  9.90 & 13.01 & 22.37 & 12.29 & 15.27 & 21.80 & 11.65 & 14.52       \\
InternVL3-8B~\cite{chen2024expanding}               & 25.12 & 16.97 & 19.67 & 22.22 & 14.90 & 17.32 & 21.70 & 13.35 & 16.05 & 22.91 & 13.54 & 16.60 & 25.00 & 14.93 & 18.28 & 23.25 & 14.83 & 17.60     \\
InternVL3-14B~\cite{chen2024expanding}              & 29.03 & \colorbox[rgb]{0.996, 1.0, 0.835}{21.43} & \colorbox[rgb]{1.0, 0.894, 0.812}{23.97} & 22.49 & 14.00 & 16.78 & 25.22 & 16.35 & \colorbox[rgb]{1.0, 0.894, 0.812}{19.45} & 25.07 & 14.96 & \colorbox[rgb]{0.996, 1.0, 0.835}{18.36} & 25.24 & \colorbox[rgb]{0.996, 1.0, 0.835}{15.79} & \colorbox[rgb]{1.0, 0.894, 0.812}{18.83} & 25.46 & 16.70 & \colorbox[rgb]{0.996, 1.0, 0.835}{19.64}      \\
InternVL3-38B~\cite{chen2024expanding}              & \colorbox[rgb]{0.996, 1.0, 0.835}{30.26} & 17.61 & 21.73 & 23.12 & 13.44 & 16.66 & 23.61 & 12.94 & 16.40 & 24.21 & 12.88 & 16.48 & 25.18 & 13.03 & 16.82 & 25.42 & 14.23 & 17.85        \\

Llama-4-Scout-17B-16E-Instruct~\cite{touvron2023llama}   & 22.82 & \colorbox[rgb]{1.0, 0.894, 0.812}{25.52} & \colorbox[rgb]{0.996, 1.0, 0.835}{23.09} & 15.94 & \colorbox[rgb]{1.0, 0.894, 0.812}{20.52} & 17.00 & 17.67 & \colorbox[rgb]{1.0, 0.894, 0.812}{22.48} & \colorbox[rgb]{0.996, 1.0, 0.835}{19.24} & 19.71 & \colorbox[rgb]{1.0, 0.894, 0.812}{21.29} & \colorbox[rgb]{1.0, 0.894, 0.812}{20.00} & 18.81 & \colorbox[rgb]{1.0, 0.894, 0.812}{19.53} & \colorbox[rgb]{0.996, 1.0, 0.835}{18.57} & 19.05 & \colorbox[rgb]{1.0, 0.894, 0.812}{22.22} & \colorbox[rgb]{1.0, 0.894, 0.812}{19.73}   \\

LLaVA-NeXT-7B-Mistral~\cite{liu2024improved}        & 26.88 & 10.83 & 14.77 & 27.44 & 10.19 & 14.25 & 22.79 & 8.72 & 12.05 & 21.74 & 8.62 & 12.01 & 26.11 & 8.65 & 12.61 & 25.08 & 9.58 & 13.32       \\
LLaVA-NeXT-8B~\cite{liu2024improved}                & 26.34 & 13.76 & 17.09 & \colorbox[rgb]{1.0, 0.894, 0.812}{\textbf{29.36}} & 12.92 & \colorbox[rgb]{0.996, 1.0, 0.835}{17.62} & 26.44 & 10.09 & 14.24 & 25.03 & 11.08 & 14.91 & 26.05 & 10.08 & 14.27 & 26.82 & 11.91 & 15.93           \\
LLaVA-NeXT-13B~\cite{liu2024improved}               & 9.21 & 4.84 & 6.00 & 14.61 & 7.13 & 9.41 & 13.16 & 5.64 & 7.46 & 13.41 & 6.72 & 8.50 & 15.73 & 6.06 & 8.42 & 12.84 & 6.07 & 7.88        \\
LLaVA-NeXT-34B~\cite{liu2024improved}               & 19.98 & 9.73 & 12.64 & \colorbox[rgb]{0.996, 1.0, 0.835}{28.72} & 11.01 & 15.54 & 22.86 & 9.18 & 12.81 & 22.20 & 8.62 & 12.06 & 25.02 & 9.78 & 13.79 & 23.68 & 9.72 & 13.39      \\

Qwen2.5-VL-3B-Instruct~\cite{bai2023qwen}           & 25.87 & 11.73 & 15.64 & 17.71 & 10.70 & 12.54 & 20.30 & 7.90 & 10.90 & 19.61 & 8.66 & 11.63 & 20.13 & 10.37 & 13.03 & 20.90 & 9.95 & 12.85    \\
Qwen2.5-VL-7B-Instruct~\cite{bai2023qwen}           & 27.13 & 14.58 & 18.20 & 21.17 & 13.11 & 14.84 & 20.89 & 9.46 & 12.26 & 23.01 & 10.46 & 13.86 & 26.58 & 10.49 & 13.81 & 23.51 & 11.95 & 14.86   \\
Qwen2.5-VL-32B-Instruct~\cite{bai2023qwen}          & 16.15 & 19.16 & 15.95 & 13.10 & \colorbox[rgb]{0.996, 1.0, 0.835}{16.56} & 14.15 & 13.69 & \colorbox[rgb]{0.996, 1.0, 0.835}{16.58} & 14.19 & 13.05 & \colorbox[rgb]{0.996, 1.0, 0.835}{16.64} & 13.80 & 14.69 & 16.61 & 15.10 & 14.13 & \colorbox[rgb]{0.996, 1.0, 0.835}{17.23} & 14.63 \\

TinyLLaVA-1.5B~\cite{zhou2024tinyllava}             & \colorbox[rgb]{1.0, 0.894, 0.812}{\textbf{34.19}} & 4.42 & 7.64 & 14.42 & 1.48 & 2.30 & \colorbox[rgb]{1.0, 0.894, 0.812}{\textbf{31.62}} & 2.96 & 4.85 & \colorbox[rgb]{0.996, 1.0, 0.835}{25.20} & 1.63 & 2.90 & \colorbox[rgb]{1.0, 0.894, 0.812}{\textbf{33.64}} & 2.07 & 3.67 & \colorbox[rgb]{0.996, 1.0, 0.835}{26.90} & 2.60 & 4.41   \\
TinyLLaVA-2B~\cite{zhou2024tinyllava}               & 5.29 & 0.54 & 0.89 & 14.30 & 1.11 & 1.92 & 15.21 & 1.31 & 2.25 & 11.05 & 0.64 & 1.15 & 11.58 & 0.88 & 1.62 & 11.32 & 0.89 & 1.55             \\
TinyLLaVA-3.1B~\cite{zhou2024tinyllava}             & 21.94 & 3.67 & 5.81 & 17.01 & 2.53 & 3.90 & 18.68 & 2.12 & 3.45 & 18.76 & 1.81 & 3.08 & 26.40 & 2.80 & 4.72 & 19.86 & 2.61 & 4.20          \\

\Xhline{1.5pt}
\end{tabular}%
}
\end{table*}

Table~\ref{tab:app_rouge-l} complements the ROUGE-L scores by providing a comprehensive evaluation of the models' keyword coverage and reasoning fluency. Gemini-2.5-pro achieves the highest recall rate, indicating its superior ability to capture and reason with complex information. In contrast, InternVL2-26B demonstrates the best precision rate, reflecting its high accuracy in handling smaller, more focused reasoning tasks.
\begin{table*}[ht]
\caption{Additional performance evaluation of VLMs using \textbf{ROUGE-L: Precision and Recall}. \textbf{Boldface} indicates the best overall across all models. \colorbox[rgb]{1.0, 0.894, 0.812}{Peach} and \colorbox[rgb]{0.996, 1.0, 0.835}{Lemon} denote the best and the second-best among open-source models, respectively.}
\label{tab:app_rouge-l}
\centering
\resizebox{\textwidth}{!}{%
\begin{tabular}{c|cc|cc|cc|cc|cc|cc}
\Xhline{1.5pt}
\cellcolor[HTML]{FFFFFF}  
& \multicolumn{2}{c}{\cellcolor[HTML]{FFFFFF}\textbf{QA}} & \multicolumn{2}{c}{\cellcolor[HTML]{FFFFFF}\textbf{SU}} & \multicolumn{2}{c}{\cellcolor[HTML]{FFFFFF}\textbf{OD}} 
& \multicolumn{2}{c}{\cellcolor[HTML]{FFFFFF}\textbf{DM}} & \multicolumn{2}{c}{\cellcolor[HTML]{FFFFFF}\textbf{EM}} & \multicolumn{2}{c}{\cellcolor[HTML]{FFFFFF}\textbf{Overall}} \\
\multirow{-2}{*}{\cellcolor[HTML]{FFFFFF}\textbf{Model}}  & \textbf{Precision} & \textbf{Recall} & \textbf{Precision} & \textbf{Recall} & \textbf{Precision} & \textbf{Recall} & \textbf{Precision} & \textbf{Recall} & \textbf{Precision} & \textbf{Recall} & \textbf{Precision} & \textbf{Recall} \\ 
\Xhline{1.2pt}
\multicolumn{13}{c}{\cellcolor[rgb]{0.906, 0.898, 0.902}\textbf{Proprietary Large Vision-Language Models}} \\                                                                                                                                                                       
Claude-Sonnet-4.5                                         & 46.28 & 49.20 & 41.58 & 43.16 & 46.07 & 45.61 & 47.65 & 45.33 & 46.77 & 43.93 & 45.39 & 45.68     \\

Gemini-2.5-pro~\cite{team2023gemini}                      & 29.01 & \textbf{50.65} & 25.68 & \textbf{46.72} & 28.54 & \textbf{47.07} & 30.22 & \textbf{48.43} & 30.08 & \textbf{46.82} & 28.43 & \textbf{48.13}   \\

GPT-4.1~\cite{achiam2023gpt}                              & 53.09 & 50.25 & 50.18 & 43.23 & 53.35 & 43.47 & 56.97 & 45.12 & 54.06 & 41.94 & 53.29 & 45.28   \\

GPT-5~\cite{leon2025gpt}                                  & 47.11 & 23.32 & 43.61 & 20.62 & 46.09 & 20.34 & 48.88 & 21.32 & 50.49 & 20.27 & 46.72 & 21.34     \\ 

\Xhline{1.2pt}
\multicolumn{13}{c}{\cellcolor[rgb]{0.906, 0.898, 0.902}\textbf{Open-Source Large Vision-Language Models}} \\

DeepSeek-VL2~\cite{lu2024deepseek}                        & 42.72 & 11.13 & 24.15 & 6.30 & 36.55 & 9.47 & 52.57 & 9.09 & 47.66 & 4.48 & 39.32 & 8.53                    \\
DeepSeek-VL2-small~\cite{lu2024deepseek}                  & 48.79 & 11.55 & 35.08 & 5.12 & 46.99 & 8.84 & 44.35 & 10.55 & 49.28 & 8.03 & 44.18 & 8.86                    \\
DeepSeek-VL2-tiny~\cite{lu2024deepseek}                   & 46.43 & 11.25 & 29.24 & 5.45 & 43.35 & 9.20 & 55.58 & 7.92 & 61.51 & 3.93 & 44.85 & 8.00                \\

InternVL2-2B~\cite{chen2024expanding}                     & 54.61 & 22.01 & 43.77 & 23.01 & 53.37 & 21.74 & 51.75 & 21.80 & 50.29 & 21.50 & 50.67 & 22.11               \\
InternVL2-4B~\cite{chen2024expanding}                     & 52.32 & 34.17 & 51.91 & 27.86 & 50.69 & 26.78 & 53.33 & 29.13 & 51.79 & 26.97 & 52.04 & 29.38              \\
InternVL2-8B~\cite{chen2024expanding}                     & 57.07 & 29.92 & 50.72 & 32.58 & 51.62 & 29.94 & 54.09 & 30.65 & 52.77 & 31.14 & 53.36 & 30.85                \\
InternVL2-26B~\cite{chen2024expanding}                    & 58.87 & 30.01 & 57.27 & 32.23 & \colorbox[rgb]{0.996, 1.0, 0.835}{58.38} & 29.04 & \colorbox[rgb]{1.0, 0.894, 0.812}{\textbf{59.42}} & 29.70 & 60.08 & 28.00 & \colorbox[rgb]{1.0, 0.894, 0.812}{\textbf{58.60}} & 30.10            \\

InternVL3-2B~\cite{chen2024expanding}                     & 52.43 & 33.38 & 49.37 & 27.31 & 51.03 & 27.29 & 53.84 & 26.88 & 52.55 & 29.83 & 51.68 & 28.99                \\
InternVL3-8B~\cite{chen2024expanding}                     & 53.05 & 36.86 & 48.95 & 33.86 & 51.04 & 32.48 & 52.40 & 32.64 & 52.69 & 34.04 & 51.46 & 34.10               \\
InternVL3-14B~\cite{chen2024expanding}                    & 54.87 & 41.40 & 49.88 & 31.99 & 53.15 & 35.59 & 53.36 & 34.13 & 52.10 & 34.63 & 52.71 & 35.76               \\
InternVL3-38B~\cite{chen2024expanding}                    & 59.20 & 36.50 & 51.28 & 31.63 & 53.01 & 31.76 & 53.63 & 31.57 & 53.73 & 31.28 & 54.32 & 32.83                 \\

Llama-4-Scout-17B-16E-Instruct~\cite{touvron2023llama}    & 44.25 & \colorbox[rgb]{1.0, 0.894, 0.812}{48.54} & 36.34 & \colorbox[rgb]{1.0, 0.894, 0.812}{43.48} & 38.58 & \colorbox[rgb]{1.0, 0.894, 0.812}{46.76} & 43.17 & \colorbox[rgb]{1.0, 0.894, 0.812}{45.19} & 42.13 & \colorbox[rgb]{1.0, 0.894, 0.812}{42.69} & 40.71 & \colorbox[rgb]{1.0, 0.894, 0.812}{45.68}  \\

LLaVA-NeXT-7B-Mistral~\cite{liu2024improved}              & 57.25 & 24.09 & 56.55 & 23.19 & 53.37 & 22.75 & 55.33 & 24.03 & 58.37 & 21.84 & 56.01 & 23.36           \\
LLaVA-NeXT-8B~\cite{liu2024improved}                      & 52.46 & 27.65 & \colorbox[rgb]{0.996, 1.0, 0.835}{59.28} & 27.68 & \colorbox[rgb]{1.0, 0.894, 0.812}{\textbf{58.87}} & 24.52 & 57.08 & 27.09 & 55.97 & 24.69 & \colorbox[rgb]{0.996, 1.0, 0.835}{56.73} & 26.62                 \\
LLaVA-NeXT-13B~\cite{liu2024improved}                     & 33.40 & 19.12 & 41.30 & 20.49 & 47.24 & 19.63 & 41.13 & 21.22 & 43.49 & 19.46 & 40.71 & 20.02              \\
LLaVA-NeXT-34B~\cite{liu2024improved}                     & 46.64 & 25.39 & \colorbox[rgb]{1.0, 0.894, 0.812}{\textbf{59.55}} & 24.42 & 54.51 & 24.46 & \colorbox[rgb]{0.996, 1.0, 0.835}{57.40} & 24.13 & 56.93 & 24.70 & 54.61 & 24.64            \\

Qwen2.5-VL-3B-Instruct~\cite{bai2023qwen}                 & 50.75 & 24.94 & 40.45 & 26.78 & 47.97 & 22.27 & 45.38 & 22.93 & 44.52 & 26.24 & 45.94 & 24.58         \\
Qwen2.5-VL-7B-Instruct~\cite{bai2023qwen}                 & 50.94 & 29.64 & 44.56 & 29.58 & 49.50 & 25.13 & 49.51 & 25.68 & 52.46 & 24.50 & 48.93 & 27.41         \\
Qwen2.5-VL-32B-Instruct~\cite{bai2023qwen}                & 37.64 & \colorbox[rgb]{0.996, 1.0, 0.835}{41.91} & 30.84 & \colorbox[rgb]{0.996, 1.0, 0.835}{38.14} & 33.27 & \colorbox[rgb]{0.996, 1.0, 0.835}{38.32} & 33.87 & \colorbox[rgb]{0.996, 1.0, 0.835}{40.89} & 34.61 & \colorbox[rgb]{0.996, 1.0, 0.835}{39.62} & 34.01 & \colorbox[rgb]{0.996, 1.0, 0.835}{39.81}      \\

TinyLLaVA-1.5B~\cite{zhou2024tinyllava}                   & \colorbox[rgb]{1.0, 0.894, 0.812}{\textbf{70.75}} & 10.91 & 40.74 & 4.89 & 54.96 & 7.03 & 54.20 & 4.98 & \colorbox[rgb]{0.996, 1.0, 0.835}{62.18} & 5.96 & 55.94 & 6.95         \\
TinyLLaVA-2B~\cite{zhou2024tinyllava}                     & 38.54 & 2.05 & 32.93 & 3.37 & 37.32 & 3.21 & 36.32 & 1.98 & \colorbox[rgb]{1.0, 0.894, 0.812}{\textbf{63.36}} & 2.69 & 38.91 & 2.66                   \\
TinyLLaVA-3.1B~\cite{zhou2024tinyllava}                   & \colorbox[rgb]{0.996, 1.0, 0.835}{59.75} & 10.34 & 47.46 & 8.33 & 46.53 & 7.33 & 48.97 & 6.25 & 59.22 & 8.24 & 51.81 & 8.21                \\

\Xhline{1.5pt}
\end{tabular}%
}
\end{table*}

\subsection{Semantic Evaluation of Reasoning Content}
\label{app:semantic_evaluation}
Table~\ref{tab:app_bertscore} illustrates the reasoning capabilities of the models from the perspective of semantic similarity in their reasoning text. As it directly measures the semantic alignment between the generated reasoning and the reference, BERTScore provides insight into how well the model captures the underlying meaning, even when the specific wording may differ. GPT-4.1 shows strong precision and recall across multiple tasks, reflecting its ability to generate reasoning chains that are both semantically aligned with the reference and provide a detailed analysis. Claude-Sonnet-4.5 has highest recall in OD and EM with greater depth of reasoning ability. In open-source models, Llama-4-Scout-17B-16E-Instruct has the highest recall and InternVL2-26B has the highest precision.
\begin{table*}[ht]
\caption{Additional performance evaluation of various VLMs, including \textbf{BERTScore precision and recall}. ``-'' indicates that the model is unable to output a normal CoT reasoning, resulting in a negative BERTScore. \textbf{Boldface} indicates the best overall across all models. \colorbox[rgb]{1.0, 0.894, 0.812}{Peach} and \colorbox[rgb]{0.996, 1.0, 0.835}{Lemon} denote the best and the second-best among open-source models, respectively.}
\label{tab:app_bertscore}
\centering
\resizebox{\textwidth}{!}{%
\begin{tabular}{c|cc|cc|cc|cc|cc|cc}
\Xhline{1.5pt}
\cellcolor[HTML]{FFFFFF}  
& \multicolumn{2}{c}{\cellcolor[HTML]{FFFFFF}\textbf{QA}} & \multicolumn{2}{c}{\cellcolor[HTML]{FFFFFF}\textbf{SU}} & \multicolumn{2}{c}{\cellcolor[HTML]{FFFFFF}\textbf{OD}} 
& \multicolumn{2}{c}{\cellcolor[HTML]{FFFFFF}\textbf{DM}} & \multicolumn{2}{c}{\cellcolor[HTML]{FFFFFF}\textbf{EM}} & \multicolumn{2}{c}{\cellcolor[HTML]{FFFFFF}\textbf{Overall}} \\
\multirow{-2}{*}{\cellcolor[HTML]{FFFFFF}\textbf{Model}}  & \textbf{Precision} & \textbf{Recall}  & \textbf{Precision} & \textbf{Recall} & \textbf{Precision} & \textbf{Recall} & \textbf{Precision} & \textbf{Recall} & \textbf{Precision} & \textbf{Recall} & \textbf{Precision} & \textbf{Recall} \\ 
\Xhline{1.2pt}
\multicolumn{13}{c}{\cellcolor[rgb]{0.906, 0.898, 0.902}\textbf{Proprietary Large Vision-Language Models}} \\                                                                                                                                       
Claude-Sonnet-4.5                        & 48.38 & 48.48 & 44.11 & 43.67 & 47.26 & \textbf{44.45} & 47.53 & 43.18 & 46.17 & \textbf{42.13} & 46.70 & 44.78 \\

Gemini-2.5-pro~\cite{team2023gemini}     & 27.56 & 43.82 & 24.15 & 38.12 & 26.34 & 38.42 & 26.82 & 39.50 & 27.16 & 37.55 & 26.28 & 39.83  \\

GPT-4.1~\cite{achiam2023gpt}             & \textbf{53.71} & \textbf{50.54} & \textbf{51.13} & \textbf{45.77} & \textbf{52.58} & 44.15 & \textbf{54.12} & \textbf{44.35} & \textbf{50.57} & 40.96 & \textbf{52.61} & \textbf{45.87}  \\

GPT-5~\cite{leon2025gpt}                 & 36.46 & 24.74 & 37.55 & 22.07 & 35.52 & 17.73 & 37.53 & 19.15 & 39.53 & 19.17 & 37.06 & 20.98    \\ 

\Xhline{1.2pt}
\multicolumn{13}{c}{\cellcolor[rgb]{0.906, 0.898, 0.902}\textbf{Open-Source Large Vision-Language Models}} \\

DeepSeek-VL2~\cite{lu2024deepseek}        & 9.01 &     - &  3.31 &     - &  3.66 &     - &     - &     - &  0.74 &     - &  2.36 & -                    \\
DeepSeek-VL2-small~\cite{lu2024deepseek}      -  &     - &  6.28 &     - &  1.34 &     - &     - &     - &  0.17 &     - &    -  & -                    \\
DeepSeek-VL2-tiny~\cite{lu2024deepseek}  & 26.03 &     - & 16.07 &     - & 19.46 &     - & 19.32 &     - & 10.37 &     - & 19.33 & -                    \\

InternVL2-2B~\cite{chen2024expanding}    & 36.16 & 16.53 & 32.56 & 19.14 & 36.03 & 14.29 & 35.00 & 13.86 & 31.98 & 10.01 & 34.60 & 15.55                 \\
InternVL2-4B~\cite{chen2024expanding}    & 44.50 & 34.48 & 42.28 & 27.64 & 39.00 & 23.04 & 43.19 & 25.75 & 39.83 & 24.12 & 42.11 & 27.69                 \\
InternVL2-8B~\cite{chen2024expanding}    & \colorbox[rgb]{0.996, 1.0, 0.835}{47.42} & 30.74 & 43.92 & 32.52 & 44.59 & 28.85 & 45.17 & 27.46 & 43.65 & 28.52 & \colorbox[rgb]{0.996, 1.0, 0.835}{45.15} & 29.92                  \\
InternVL2-26B~\cite{chen2024expanding}   & \colorbox[rgb]{1.0, 0.894, 0.812}{47.58} & 30.60 & \colorbox[rgb]{1.0, 0.894, 0.812}{49.80} & 33.32 & \colorbox[rgb]{1.0, 0.894, 0.812}{47.06} & 25.25 & \colorbox[rgb]{1.0, 0.894, 0.812}{49.13} & 26.92 & \colorbox[rgb]{1.0, 0.894, 0.812}{47.20} & 25.41 & \colorbox[rgb]{1.0, 0.894, 0.812}{48.30} & 28.94                  \\

InternVL3-2B~\cite{chen2024expanding}    & 42.87 & 32.52 & 33.28 & 19.27 & 33.50 & 19.50 & 37.73 & 19.70 & 38.04 & 23.84 & 37.08 & 23.16                   \\
InternVL3-8B~\cite{chen2024expanding}    & 39.34 & 30.93 & 30.28 & 24.96 & 36.27 & 25.87 & 33.64 & 23.26 & 32.93 & 24.81 & 34.70 & 26.28                  \\
InternVL3-14B~\cite{chen2024expanding}   & 41.98 & \colorbox[rgb]{0.996, 1.0, 0.835}{35.01} & 31.70 & 23.23 & 39.15 & 30.02 & 39.27 & 28.37 & 37.25 & 28.65 & 37.84 & 29.11                   \\
InternVL3-38B~\cite{chen2024expanding}   & 47.16 & 34.25 & 43.23 & 30.88 & \colorbox[rgb]{0.996, 1.0, 0.835}{44.66} & 28.92 & 43.65 & 27.30 & 43.36 & 27.87 & 44.60 & 30.31                   \\

Llama-4-Scout-17B-16E-Instruct~\cite{touvron2023llama}   & 43.10 & \colorbox[rgb]{1.0, 0.894, 0.812}{44.26} & 35.64 & \colorbox[rgb]{1.0, 0.894, 0.812}{38.33} & 38.83 & \colorbox[rgb]{1.0, 0.894, 0.812}{41.87} & 41.77 & \colorbox[rgb]{1.0, 0.894, 0.812}{40.61} & 39.10 & \colorbox[rgb]{1.0, 0.894, 0.812}{38.26} & 39.72 & \colorbox[rgb]{1.0, 0.894, 0.812}{40.98}    \\

LLaVA-NeXT-7B-Mistral~\cite{liu2024improved}  & 41.00 & 19.35 & 39.95 & 16.96 & 37.17 & 13.96 & 39.65 & 17.28 & 40.47 & 17.39 & 39.64 & 17.05               \\
LLaVA-NeXT-8B~\cite{liu2024improved}          & 44.01 & 26.03 & \colorbox[rgb]{0.996, 1.0, 0.835}{45.89} & 25.98 & 42.42 & 18.53 & \colorbox[rgb]{0.996, 1.0, 0.835}{45.84} & 23.13 & 41.75 & 21.25 & 44.30 & 23.44                  \\
LLaVA-NeXT-13B~\cite{liu2024improved}         & 15.48 & 5.85  & 28.00 & 12.30 & 28.61 & 7.20  & 26.04 & 8.16  & 26.05 & 9.68  & 24.41 & 8.57               \\
LLaVA-NeXT-34B~\cite{liu2024improved}         & 30.93 & 22.23 & 44.59 & 21.64 & 43.43 & 21.44 & 42.19 & 18.72 & \colorbox[rgb]{0.996, 1.0, 0.835}{43.73} & 21.85 & 40.38 & 21.18            \\

Qwen2.5-VL-3B-Instruct~\cite{bai2023qwen}     & 39.43 & 24.39 & 33.88 & 21.95 & 36.16 & 14.54 & 35.14 & 14.76 & 35.42 & 17.83 & 36.13 & 19.21          \\
Qwen2.5-VL-7B-Instruct~\cite{bai2023qwen}     & 41.17 & 27.44 & 36.56 & 25.40 & 36.45 & 16.36 & 37.85 & 16.81 & 38.18 & 15.36 & 38.11 & 21.36         \\
Qwen2.5-VL-32B-Instruct~\cite{bai2023qwen}    & 27.80 & 34.98 & 25.52 & \colorbox[rgb]{0.996, 1.0, 0.835}{34.87} & 25.34 & \colorbox[rgb]{0.996, 1.0, 0.835}{31.66} & 23.93 & \colorbox[rgb]{0.996, 1.0, 0.835}{31.79} & 26.32 & \colorbox[rgb]{0.996, 1.0, 0.835}{34.59} & 25.82 & \colorbox[rgb]{0.996, 1.0, 0.835}{33.60}       \\

TinyLLaVA-1.5B~\cite{zhou2024tinyllava}       & 31.97 & -     & 19.61 & -     & 26.61 & -     & 22.00 & -     & 23.80 & -     & 25.01 & -         \\
TinyLLaVA-2B~\cite{zhou2024tinyllava}         & 7.50  & -     & 23.55 & -     & 13.27 & -     & 7.88  & -     & 12.81 & -     & 13.25 & -                    \\
TinyLLaVA-3.1B~\cite{zhou2024tinyllava}       & 26.08 & -     & 24.41 & -     & 21.86 & -     & 22.36 & -     & 27.86 & -     & 24.24 & -                \\

\Xhline{1.5pt}
\end{tabular}
}
\end{table*}

\begin{table*}[ht]
\caption{Additional performance evaluation of VLMs using \textbf{accuracy} across different second-level tasks. Abbreviations adopted: OC for Organism Counting; RC for Regional Counting, MS for Morphological Statistics; BA for Boundary Analysis; AE for Area Evaluation; SR for Spatial Relationship; OI for Organism Identification; AD for Anomaly Detection; PM for Plant Management; PR for Pest Recognition; BI for BioMorph Identification; DD for Disease Diagnosis; AT for Agri-Tools; AM for Agri-Methods; MD for Management Decisions. \textbf{Boldface} indicates the best overall across all models. \colorbox[rgb]{1.0, 0.894, 0.812}{Peach} and \colorbox[rgb]{0.996, 1.0, 0.835}{Lemon} denote the best and the second-best among open-source models, respectively.}
\label{tab:acc_sub}
\centering
\resizebox{\textwidth}{!}{%
\begin{tabular}{c|ccc|ccc|ccc|ccc|ccc}
\Xhline{1.5pt}
\cellcolor[HTML]{FFFFFF}  
& \multicolumn{3}{c}{\cellcolor[HTML]{FFFFFF}\textbf{QA}} & \multicolumn{3}{c}{\cellcolor[HTML]{FFFFFF}\textbf{SU}} & \multicolumn{3}{c}{\cellcolor[HTML]{FFFFFF}\textbf{OD}} 
& \multicolumn{3}{c}{\cellcolor[HTML]{FFFFFF}\textbf{DM}} & \multicolumn{3}{c}{\cellcolor[HTML]{FFFFFF}\textbf{EM}} \\
\multirow{-2}{*}{\cellcolor[HTML]{FFFFFF}\textbf{Model}}  & \textbf{OC} & \textbf{RC} & \textbf{MS} & \textbf{BA} & \textbf{AE} & \textbf{SR} & \textbf{OI} & \textbf{AD} & \textbf{PM} & \textbf{PR} & \textbf{BI} & \textbf{DD} & \textbf{AT} & \textbf{AM} & \textbf{MD} \\ 
\Xhline{1.2pt}
\multicolumn{16}{c}{\cellcolor[rgb]{0.906, 0.898, 0.902}\textbf{Proprietary Large Vision-Language Models}} \\                                                                                                                                      
Claude-Sonnet-4.5 & 48.80 & 3.88 & 47.27 & 37.56 & 48.59 & 52.24 & 70.75 & 57.43 & 76.04 & 67.67 & 79.63 & 62.10 & 81.54 & 83.47 & 61.07 \\

Gemini-2.5-pro~\cite{team2023gemini} & 44.28 & 17.48 & 46.99 & 41.37 & 48.78 & 68.16 & 82.78 & 41.40 & 79.17 & 73.31 & 88.89 & 66.42 & 93.85 & \textbf{90.08} & \textbf{79.01} \\

GPT-4.1~\cite{achiam2023gpt} & 43.07 & 18.45 & 45.90 & 39.34 & 52.35 & 59.20 & 80.71 & 51.02 & 78.12 & 71.80 & 87.04 & 69.23 & 90.77 & 87.60 & 72.90 \\

GPT-5~\cite{leon2025gpt} & 52.41 & 17.48 & \textbf{56.83} & \textbf{44.42} & \textbf{57.82} & \textbf{64.18} & \textbf{82.16} & 51.31 & 76.04 & \textbf{75.56} & \textbf{90.74} & \textbf{74.11} & \textbf{96.92} & 87.60 & 71.76 \\ 

\Xhline{1.2pt}
\multicolumn{16}{c}{\cellcolor[rgb]{0.906, 0.898, 0.902}\textbf{Open-Source Large Vision-Language Models}} \\

DeepSeek-VL2~\cite{lu2024deepseek} & 28.92 & 62.14 & 15.57 & 11.68 & 22.03 & 5.97 & 24.48 & 19.24 & 13.54 & 7.52 & 35.19 & 19.70 & 47.69 & 34.71 & 22.52 \\
DeepSeek-VL2-small~\cite{lu2024deepseek} & 33.43 & 30.10 & 35.25 & 9.90 & 22.60 & 8.46 & 28.63 & 37.61 & 12.50 & 6.39 & 42.59 & 33.96 & 49.23 & 39.67 & 19.85 \\
DeepSeek-VL2-tiny~\cite{lu2024deepseek} & 21.84 & 14.56 & 12.02 & 9.14 & 25.61 & 5.97 & 31.95 & 16.62 & 7.29 & 16.17 & 33.33 & 31.71 & 33.85 & 29.75 & 32.82 \\

InternVL2-2B~\cite{chen2024expanding} & 32.08 & 59.22 & 39.07 & 8.88 & 24.11 & 13.43 & 22.82 & 42.57 & 14.58 & 7.52 & 47.22 & 36.21 & 38.46 & 43.80 & 24.81 \\
InternVL2-4B~\cite{chen2024expanding} & 29.67 & 5.83 & 15.30 & 28.43 & 25.42 & 6.47 & 45.85 & 25.36 & 30.21 & 22.93 & 46.30 & 47.47 & 33.85 & 46.28 & 41.98 \\
InternVL2-8B~\cite{chen2024expanding} & 40.51 & 48.54 & 30.60 & 30.96 & 42.37 & 51.24 & 50.41 & 66.18 & 41.67 & 61.28 & 59.26 & 39.96 & 53.85 & 61.98 & 45.04 \\
InternVL2-26B~\cite{chen2024expanding} & 49.55 & 43.69 & 30.33 & 29.19 & 39.17 & 47.76 & 67.01 & 44.61 & 38.54 & 63.53 & \colorbox[rgb]{0.996, 1.0, 0.835}{75.93} & 47.28 & 64.62 & 71.90 & \colorbox[rgb]{0.996, 1.0, 0.835}{66.03} \\

InternVL3-2B~\cite{chen2024expanding} & 34.34 & 69.90 & 27.32 & 16.24 & 45.57 & 39.30 & 36.31 & 51.90 & 26.04 & 9.77 & 55.56 & 36.02 & 58.46 & 58.68 & 30.92 \\
InternVL3-8B~\cite{chen2024expanding} & 43.22 & 22.33 & 25.14 & 33.25 & 41.43 & \colorbox[rgb]{1.0, 0.894, 0.812}{53.73} & 59.13 & \colorbox[rgb]{1.0, 0.894, 0.812}{\textbf{69.97}} & 58.33 & 62.78 & 71.30 & 49.72 & 66.15 & 74.38 & 48.85 \\
InternVL3-14B~\cite{chen2024expanding} & 48.95 & 60.19 & \colorbox[rgb]{1.0, 0.894, 0.812}{45.08} & \colorbox[rgb]{0.996, 1.0, 0.835}{38.07} & \colorbox[rgb]{1.0, 0.894, 0.812}{48.21} & 48.26 & 58.92 & 63.27 & \colorbox[rgb]{0.996, 1.0, 0.835}{76.04} & \colorbox[rgb]{1.0, 0.894, 0.812}{68.42} & 68.52 & 60.60 & \colorbox[rgb]{0.996, 1.0, 0.835}{70.77} & \colorbox[rgb]{0.996, 1.0, 0.835}{80.17} & 64.50 \\
InternVL3-38B~\cite{chen2024expanding} & \colorbox[rgb]{1.0, 0.894, 0.812}{\textbf{55.42}} & 32.04 & 34.97 & \colorbox[rgb]{1.0, 0.894, 0.812}{39.59} & \colorbox[rgb]{0.996, 1.0, 0.835}{47.65} & \colorbox[rgb]{0.996, 1.0, 0.835}{52.74} & \colorbox[rgb]{1.0, 0.894, 0.812}{75.31} & \colorbox[rgb]{0.996, 1.0, 0.835}{66.47} & \colorbox[rgb]{1.0, 0.894, 0.812}{\textbf{81.25}} & 66.92 & 73.15 & \colorbox[rgb]{1.0, 0.894, 0.812}{64.92} & 70.77 & 78.51 & \colorbox[rgb]{1.0, 0.894, 0.812}{68.32} \\

Llama-4-Scout-17B-16E-Instruct~\cite{touvron2023llama} & \colorbox[rgb]{0.996, 1.0, 0.835}{50.15} & 9.71 & 32.24 & 34.77 & 42.75 & 42.29 & \colorbox[rgb]{0.996, 1.0, 0.835}{68.67} & 51.31 & 64.58 & 65.79 & 68.52 & \colorbox[rgb]{0.996, 1.0, 0.835}{64.17} & \colorbox[rgb]{1.0, 0.894, 0.812}{72.31} & \colorbox[rgb]{1.0, 0.894, 0.812}{80.99} & 55.34 \\

LLaVA-NeXT-7B-Mistral~\cite{liu2024improved} & 17.32 & \colorbox[rgb]{1.0, 0.894, 0.812}{\textbf{73.79}} & 33.06 & 7.36 & 20.90 & 12.94 & 31.33 & 9.62 & 21.88 & 31.20 & 37.96 & 19.70 & 44.62 & 50.41 & 32.82 \\
LLaVA-NeXT-8B~\cite{liu2024improved} & 31.93 & \colorbox[rgb]{0.996, 1.0, 0.835}{71.84} & \colorbox[rgb]{0.996, 1.0, 0.835}{41.26} & 31.98 & 16.38 & 25.87 & 43.15 & 23.62 & 41.67 & 59.40 & 31.48 & 34.33 & 40.00 & 47.11 & 33.97 \\
LLaVA-NeXT-13B~\cite{liu2024improved} & 8.89 & 25.24 & 8.47 & 8.12 & 18.08 & 11.44 & 14.32 & 7.29 & 10.42 & 7.14 & 14.81 & 17.07 & 24.62 & 21.49 & 13.36 \\
LLaVA-NeXT-34B~\cite{liu2024improved} & 49.10 & 67.96 & 45.08 & 36.80 & 15.63 & 52.74 & 41.91 & 66.18 & 39.58 & 60.53 & 33.33 & 29.46 & 50.77 & 54.55 & 41.98 \\

Qwen2.5-VL-3B-Instruct~\cite{bai2023qwen}  & 43.52 & 14.56 & 19.95 & 33.76 & 36.35 & 52.24 & 56.64 & 65.01 & 72.92 & 57.52 & 63.89 & 55.91 & 58.46 & 57.85 & 44.27 \\
Qwen2.5-VL-7B-Instruct~\cite{bai2023qwen} & 48.34 & 4.85 & 30.87 & 28.68 & 36.16 & 49.25 & 64.32 & 62.39 & 56.25 & \colorbox[rgb]{0.996, 1.0, 0.835}{67.29} & \colorbox[rgb]{1.0, 0.894, 0.812}{76.85} & 49.34 & 69.23 & 63.64 & 43.51 \\
Qwen2.5-VL-32B-Instruct~\cite{bai2023qwen} & 38.40 & 18.45 & 32.24 & 8.63 & 20.90 & 11.44 & 20.12 & 8.75 & 15.62 & 7.14 & 28.70 & 19.14 & 27.69 & 37.19 & 15.27 \\

TinyLLaVA-1.5B~\cite{zhou2024tinyllava} & 23.64 & 25.24 & 10.38 & 13.45 & 18.64 & 5.97 & 27.18 & 7.29 & 8.33 & 7.89 & 36.11 & 33.02 & 32.31 & 29.75 & 14.50 \\
TinyLLaVA-2B~\cite{zhou2024tinyllava} & 43.07 & 0.00 & 28.69 & 28.17 & 24.86 & 6.97 & 64.11 & 20.70 & 37.50 & 53.01 & 40.74 & 44.47 & 41.54 & 43.80 & 39.69 \\
TinyLLaVA-3.1B~\cite{zhou2024tinyllava} & 38.55 & 13.59 & 30.05 & 23.86 & 19.40 & 8.46 & 58.30 & 12.83 & 62.50 & 42.86 & 45.37 & 38.46 & 47.69 & 61.16 & 39.31 \\

\Xhline{1.5pt}
\end{tabular}%
}
\end{table*}

\subsection{Performance at the Second-level Dimensions}
\label{app:performance_dimensions}
Table~\ref{tab:acc_sub} presents the performance of various VLMs across multiple second-level tasks in agricultural scenarios. Among the proprietary models, GPT-5 achieves highest accuracy in SU and DM of first-level dimensions, demonstrating strong performance in tasks requiring fine-grained analysis. Gemini-2.5-pro also performs well, especially in AM and MD, demonstrating a wealth of agricultural management and decision-making knowledge. Among open-source models, InternVL3 series, Llama-4-Scout-17B-16E-Instruct and LLaVA-NeXT-7B-Mistral stand out, with InternVL3-38B excelling in multiple dimensions like OC and PM, while LLaVA-NeXT-7B-Mistral performs well in RC. TinyLLaVA-1.5B show lower accuracy, especially in tasks like SR and PR, suggesting that smaller models may struggle with more complex tasks. The overall trends indicate that proprietary models tend to perform better across multiple second-level tasks, particularly in tasks requiring complex spatial analysis, such as BA, AE and SR.

As presented in Table~\ref{tab:rouge_sub} and Table~\ref{tab:bf_sub}, ROUGE-L F1 and BERTScore F1 assess the reasoning capability of models across secondary dimensions, with one focusing on content-structural alignment and the other on semantic similarity. As shown in this two tables, the proprietary models, including Claude-Sonnet-4.5 and GPT-4.1, outperform the open-source models across most second-level dimensions. However, several open-source models such as InternVL3 series and Llama-4-Scout-17B-16E-Instruct also exhibit comparable performance in RC, AD and MD.

\begin{table*}[ht]
\caption{Additional performance evaluation of VLMs using \textbf{ROUGE-L F1} across different second-level tasks. Abbreviations adopted: OC for Organism Counting; RC for Regional Counting, MS for Morphological Statistics; BA for Boundary Analysis; AE for Area Evaluation; SR for Spatial Relationship; OI for Organism Identification; AD for Anomaly Detection; PM for Plant Management; PR for Pest Recognition; BI for BioMorph Identification; DD for Disease Diagnosis; AT for Agri-Tools; AM for Agri-Methods; MD for Management Decisions. \textbf{Boldface} indicates the best overall across all models. \colorbox[rgb]{1.0, 0.894, 0.812}{Peach} and \colorbox[rgb]{0.996, 1.0, 0.835}{Lemon} denote the best and the second-best among open-source models, respectively.}
\label{tab:rouge_sub}
\centering
\resizebox{\textwidth}{!}{%
\begin{tabular}{c|ccc|ccc|ccc|ccc|ccc}
\Xhline{1.5pt}
\cellcolor[HTML]{FFFFFF}  
& \multicolumn{3}{c}{\cellcolor[HTML]{FFFFFF}\textbf{QA}} & \multicolumn{3}{c}{\cellcolor[HTML]{FFFFFF}\textbf{SU}} & \multicolumn{3}{c}{\cellcolor[HTML]{FFFFFF}\textbf{OD}} 
& \multicolumn{3}{c}{\cellcolor[HTML]{FFFFFF}\textbf{DM}} & \multicolumn{3}{c}{\cellcolor[HTML]{FFFFFF}\textbf{EM}} \\
\multirow{-2}{*}{\cellcolor[HTML]{FFFFFF}\textbf{Model}}  & \textbf{OC} & \textbf{RC} & \textbf{MS} & \textbf{BA} & \textbf{AE} & \textbf{SR} & \textbf{OI} & \textbf{AD} & \textbf{PM} & \textbf{PR} & \textbf{BI} & \textbf{DD} & \textbf{AT} & \textbf{AM} & \textbf{MD} \\ 
\Xhline{1.2pt}
\multicolumn{16}{c}{\cellcolor[rgb]{0.906, 0.898, 0.902}\textbf{Proprietary Large Vision-Language Models}} \\                                                                                                                                      
Claude-Sonnet-4.5 & 47.71 & 47.47 & 45.86 & 41.91 & 41.08 & 42.63 & 48.08 & 42.67 & 45.24 & 47.73 & 48.09 & 44.63 & 42.73 & 44.11 & 45.56 \\

Gemini-2.5-pro~\cite{team2023gemini} & 37.15 & 32.74 & 36.04 & 33.22 & 31.90 & 34.70 & 36.35 & 32.90 & 36.94 & 38.33 & 37.91 & 35.76 & 36.11 & 36.72 & 36.00 \\

GPT-4.1~\cite{achiam2023gpt} & \textbf{51.31} & \textbf{50.98} & \textbf{50.44} & \textbf{45.25} & \textbf{46.39} & \textbf{44.13} & \textbf{51.10} & \textbf{43.53} & \textbf{49.76} & \textbf{54.09} & \textbf{48.64} & \textbf{47.89} & \textbf{44.50} & \textbf{44.17} & \textbf{48.25} \\

GPT-5~\cite{leon2025gpt} & 29.58 & 31.14 & 31.91 & 26.77 & 28.12 & 27.40 & 29.26 & 26.88 & 27.04 & 31.69 & 29.55 & 28.02 & 27.75 & 27.36 & 29.09 \\ 

\Xhline{1.2pt}
\multicolumn{16}{c}{\cellcolor[rgb]{0.906, 0.898, 0.902}\textbf{Open-Source Large Vision-Language Models}} \\

DeepSeek-VL2~\cite{lu2024deepseek} & 19.13 & 21.94 & 9.23 & 16.01 & 4.63 & 1.76 & 10.68 & 12.45 & 18.81 & 21.91 & 3.75 & 7.18 & 2.79 & 3.93 & 8.61 \\
DeepSeek-VL2-small~\cite{lu2024deepseek} & 19.31 & 16.12 & 12.86 & 12.56 & 5.28 & 5.29 & 15.56 & 9.67 & 8.27 & 20.10 & 7.67 & 13.70 & 6.71 & 8.32 & 14.80 \\
DeepSeek-VL2-tiny~\cite{lu2024deepseek} & 19.19 & 24.69 & 10.24 & 8.23 & 7.20 & 2.97 & 13.43 & 13.71 & 13.07 & 19.24 & 2.09 & 7.65 & 1.59 & 1.32 & 8.21 \\

InternVL2-2B~\cite{chen2024expanding} & 27.64 & 28.34 & 31.71 & 29.38 & 28.01 & 30.15 & 27.34 & 30.44 & 34.53 & 29.00 & 31.15 & 28.60 & 30.91 & 30.30 & 26.05 \\
InternVL2-4B~\cite{chen2024expanding} & 40.03 & 36.72 & 41.70 & 33.55 & 36.80 & 32.74 & 36.42 & 32.09 & 30.80 & 42.99 & 34.07 & 34.12 & 32.66 & 34.24 & 35.24 \\
InternVL2-8B~\cite{chen2024expanding} & 38.04 & 37.81 & 38.27 & 37.85 & 38.99 & 37.65 & 37.50 & 37.73 & 35.78 & 42.41 & 37.30 & 36.70 & 35.18 & 35.22 & 40.51 \\
InternVL2-26B~\cite{chen2024expanding} & 37.97 & 28.10 & 42.82 & 39.05 & \colorbox[rgb]{1.0, 0.894, 0.812}{40.15} & \colorbox[rgb]{1.0, 0.894, 0.812}{39.35} & 40.17 & 38.85 & 32.28 & 41.50 & 37.57 & 37.71 & 33.69 & 34.27 & 39.81 \\

InternVL3-2B~\cite{chen2024expanding} & 40.04 & 36.10 & 40.21 & 30.46 & 37.57 & 33.21 & 36.23 & 32.40 & 31.47 & 37.20 & 35.28 & 33.37 & 36.25 & 37.43 & 37.06 \\
InternVL3-8B~\cite{chen2024expanding} & 42.40 & 39.44 & 43.97 & \colorbox[rgb]{0.996, 1.0, 0.835}{39.56} & \colorbox[rgb]{0.996, 1.0, 0.835}{39.72} & 37.44 & 37.53 & \colorbox[rgb]{0.996, 1.0, 0.835}{40.64} & 36.31 & \colorbox[rgb]{0.996, 1.0, 0.835}{43.36} & 39.76 & 37.45 & 36.50 & \colorbox[rgb]{0.996, 1.0, 0.835}{38.67} & \colorbox[rgb]{0.996, 1.0, 0.835}{42.74} \\
InternVL3-14B~\cite{chen2024expanding} & \colorbox[rgb]{0.996, 1.0, 0.835}{45.86} & \colorbox[rgb]{1.0, 0.894, 0.812}{50.81} & \colorbox[rgb]{1.0, 0.894, 0.812}{45.64} & \colorbox[rgb]{1.0, 0.894, 0.812}{39.58} & 39.62 & 34.90 & \colorbox[rgb]{0.996, 1.0, 0.835}{40.43} & \colorbox[rgb]{1.0, 0.894, 0.812}{41.82} & \colorbox[rgb]{1.0, 0.894, 0.812}{43.31} & \colorbox[rgb]{1.0, 0.894, 0.812}{44.36} & \colorbox[rgb]{0.996, 1.0, 0.835}{42.43} & \colorbox[rgb]{0.996, 1.0, 0.835}{39.00} & \colorbox[rgb]{0.996, 1.0, 0.835}{38.03} & 37.44 & \colorbox[rgb]{1.0, 0.894, 0.812}{43.03} \\
InternVL3-38B~\cite{chen2024expanding} & 44.37 & \colorbox[rgb]{0.996, 1.0, 0.835}{43.40} & \colorbox[rgb]{0.996, 1.0, 0.835}{44.43} & 39.43 & 39.13 & 34.48 & 38.91 & 39.18 & 41.04 & 42.56 & 39.34 & 37.44 & 35.87 & 36.39 & 40.80 \\

Llama-4-Scout-17B-16E-Instruct~\cite{touvron2023llama} & \colorbox[rgb]{1.0, 0.894, 0.812}{46.25} & 40.95 & 43.60 & 38.60 & 38.10 & 38.98 & \colorbox[rgb]{1.0, 0.894, 0.812}{42.88} & 37.99 & \colorbox[rgb]{0.996, 1.0, 0.835}{42.33} & 43.15 & \colorbox[rgb]{1.0, 0.894, 0.812}{42.59} & \colorbox[rgb]{1.0, 0.894, 0.812}{43.68} & \colorbox[rgb]{1.0, 0.894, 0.812}{39.49} & \colorbox[rgb]{1.0, 0.894, 0.812}{41.30} & 41.77 \\

LLaVA-NeXT-7B-Mistral~\cite{liu2024improved} & 28.67 & 24.02 & 41.45 & 30.96 & 32.35 & 33.24 & 30.47 & 29.70 & 34.51 & 36.74 & 28.36 & 31.86 & 29.73 & 32.15 & 30.81 \\
LLaVA-NeXT-8B~\cite{liu2024improved} & 29.49 & 34.42 & 43.83 & 36.78 & 36.72 & \colorbox[rgb]{0.996, 1.0, 0.835}{39.16} & 33.43 & 35.15 & 30.22 & 36.47 & 32.53 & 36.32 & 34.33 & 34.53 & 33.38 \\
LLaVA-NeXT-13B~\cite{liu2024improved} & 22.61 & 26.36 & 24.23 & 27.02 & 25.90 & 27.86 & 28.62 & 25.69 & 16.25 & 29.18 & 23.36 & 26.56 & 24.63 & 26.61 & 25.96 \\
LLaVA-NeXT-34B~\cite{liu2024improved} & 30.24 & 32.93 & 35.38 & 35.34 & 36.97 & 22.05 & 29.78 & 36.09 & 38.84 & 34.86 & 29.40 & 33.13 & 30.35 & 30.86 & 36.30 \\

Qwen2.5-VL-3B-Instruct~\cite{bai2023qwen} & 32.03 & 35.12 & 32.45 & 31.06 & 31.01 & 29.83 & 29.09 & 28.99 & 30.96 & 31.19 & 28.64 & 29.11 & 30.16 & 30.71 & 32.72 \\
Qwen2.5-VL-7B-Instruct~\cite{bai2023qwen} & 36.67 & 30.78 & 37.11 & 31.10 & 35.32 & 34.27 & 32.71 & 29.51 & 34.16 & 35.08 & 32.74 & 31.85 & 31.80 & 33.32 & 30.32 \\
Qwen2.5-VL-32B-Instruct~\cite{bai2023qwen} & 36.99 & 41.60 & 34.89 & 33.12 & 33.34 & 33.89 & 33.21 & 33.38 & 38.47 & 36.72 & 35.93 & 34.05 & 36.10 & 36.28 & 36.00 \\

TinyLLaVA-1.5B~\cite{zhou2024tinyllava} & 18.62 & 23.43 & 16.26 & 10.77 & 7.15 & 1.96 & 9.12 & 14.29 & 11.84 & 13.89 & 5.32 & 6.62 & 7.17 & 8.07 & 12.08 \\
TinyLLaVA-2B~\cite{zhou2024tinyllava} & 1.99 & 1.17 & 6.95 & 7.80 & 5.25 & 2.51 & 1.88 & 8.96 & 5.65 & 0.59 & 1.77 & 5.23 & 2.12 & 1.76 & 6.94 \\
TinyLLaVA-3.1B~\cite{zhou2024tinyllava} & 14.45 & 13.24 & 19.79 & 10.76 & 13.55 & 13.01 & 11.95 & 12.27 & 10.18 & 6.74 & 14.42 & 11.41 & 16.60 & 14.66 & 12.21 \\

\Xhline{1.5pt}
\end{tabular}%
}
\end{table*}
\begin{table*}[ht]
\caption{Additional performance evaluation of VLMs using \textbf{BERTScore F1} across different second-level tasks. ``-'' indicates that the model is unable to output a normal CoT reasoning, resulting in a negative BERTScore. Abbreviations adopted: OC for Organism Counting; RC for Regional Counting, MS for Morphological Statistics; BA for Boundary Analysis; AE for Area Evaluation; SR for Spatial Relationship; OI for Organism Identification; AD for Anomaly Detection; PM for Plant Management; PR for Pest Recognition; BI for BioMorph Identification; DD for Disease Diagnosis; AT for Agri-Tools; AM for Agri-Methods; MD for Management Decisions. \textbf{Boldface} indicates the best overall across all models. \colorbox[rgb]{1.0, 0.894, 0.812}{Peach} and \colorbox[rgb]{0.996, 1.0, 0.835}{Lemon} denote the best and the second-best among open-source models, respectively.}
\label{tab:bf_sub}
\centering
\resizebox{\textwidth}{!}{%
\begin{tabular}{c|ccc|ccc|ccc|ccc|ccc}
\Xhline{1.5pt}
\cellcolor[HTML]{FFFFFF}  
& \multicolumn{3}{c}{\cellcolor[HTML]{FFFFFF}\textbf{QA}} & \multicolumn{3}{c}{\cellcolor[HTML]{FFFFFF}\textbf{SU}} & \multicolumn{3}{c}{\cellcolor[HTML]{FFFFFF}\textbf{OD}} 
& \multicolumn{3}{c}{\cellcolor[HTML]{FFFFFF}\textbf{DM}} & \multicolumn{3}{c}{\cellcolor[HTML]{FFFFFF}\textbf{EM}} \\
\multirow{-2}{*}{\cellcolor[HTML]{FFFFFF}\textbf{Model}}  & \textbf{OC} & \textbf{RC} & \textbf{MS} & \textbf{BA} & \textbf{AE} & \textbf{SR} & \textbf{OI} & \textbf{AD} & \textbf{PM} & \textbf{PR} & \textbf{BI} & \textbf{DD} & \textbf{AT} & \textbf{AM} & \textbf{MD} \\ 
\Xhline{1.2pt}
\multicolumn{16}{c}{\cellcolor[rgb]{0.906, 0.898, 0.902}\textbf{Proprietary Large Vision-Language Models}} \\      

Claude-Sonnet-4.5 & 48.53 & 53.44 & 46.89 & 42.26 & 45.03 & 42.67 & 48.70 & 43.55 & 47.44 & 48.65 & 47.17 & 43.33 & 39.70 & \textbf{40.71} & 46.80  \\

Gemini-2.5-pro~\cite{team2023gemini} & 36.21 & 36.93 & 33.55 & 29.49 & 30.51 & 36.75 & 33.79 & 29.36 & 34.29 & 36.25 & 33.23 & 31.38 & 29.59 & 29.97 & 34.00 \\

GPT-4.1~\cite{achiam2023gpt} & \textbf{52.20} & \textbf{54.85} & \textbf{51.23} & \textbf{45.81} & \textbf{50.03} & \textbf{46.54} & \textbf{51.91} & \textbf{45.15} & \textbf{51.86} & \textbf{54.84} & \textbf{47.76} & \textbf{46.54} & \textbf{42.29} & 40.33 & \textbf{48.86} \\

GPT-5~\cite{leon2025gpt} & 29.08 & 33.37 & 32.07 & 24.88 & 31.45 & 31.25 & 28.52 & 26.47 & 21.51 & 32.83 & 26.55 & 25.76 & 25.31 & 25.93 & 30.96 \\ 

\Xhline{1.2pt}
\multicolumn{16}{c}{\cellcolor[rgb]{0.906, 0.898, 0.902}\textbf{Open-Source Large Vision-Language Models}} \\

DeepSeek-VL2~\cite{lu2024deepseek} & - & 3.05 & - & - & - & - & - & - & - & 2.23 & - & - & - & - & - \\
DeepSeek-VL2-small~\cite{lu2024deepseek} & - & 2.91 & - & - & - & - & - & - & - & - & - & - & - & - & - \\
DeepSeek-VL2-tiny~\cite{lu2024deepseek} & 11.29 & 27.19 & - & - & - & - & - & - & - & 12.73 & - & - & - & - & - \\

InternVL2-2B~\cite{chen2024expanding} & 24.50 & 26.49 & 27.62 & 23.86 & 25.65 & 30.47 & 21.75 & 26.66 & 24.81 & 26.50 & 24.38 & 22.29 & 24.80 & 22.57 & 17.75 \\
InternVL2-4B~\cite{chen2024expanding} & 38.73 & 40.87 & 40.11 & 29.44 & 37.56 & 35.86 & 34.57 & 28.54 & 24.88 & 42.76 & 30.43 & 30.47 & 28.61 & 29.75 & 33.26 \\
InternVL2-8B~\cite{chen2024expanding} & 38.18 & 39.78 & 39.45 & 34.97 & \colorbox[rgb]{0.996, 1.0, 0.835}{40.04} & 37.12 & 37.39 & \colorbox[rgb]{1.0, 0.894, 0.812}{38.15} & 30.47 & \colorbox[rgb]{0.996, 1.0, 0.835}{42.79} & 32.73 & 33.18 & 31.36 & 30.41 & 39.32 \\
InternVL2-26B~\cite{chen2024expanding} & 37.97 & 28.52 & \colorbox[rgb]{1.0, 0.894, 0.812}{43.09} & \colorbox[rgb]{1.0, 0.894, 0.812}{36.84} & \colorbox[rgb]{1.0, 0.894, 0.812}{41.58} & \colorbox[rgb]{1.0, 0.894, 0.812}{42.16} & \colorbox[rgb]{0.996, 1.0, 0.835}{39.15} & 37.58 & 25.79 & \colorbox[rgb]{1.0, 0.894, 0.812}{43.09} & 33.56 & \colorbox[rgb]{0.996, 1.0, 0.835}{35.39} & 31.03 & 30.16 & \colorbox[rgb]{0.996, 1.0, 0.835}{39.36} \\

InternVL3-2B~\cite{chen2024expanding} & 37.95 & 38.41 & 36.57 & 22.90 & 25.67 & 34.92 & 26.52 & 25.31 & 25.17 & 33.39 & 27.08 & 25.93 & 27.69 & 26.11 & 33.41 \\
InternVL3-8B~\cite{chen2024expanding} & 36.49 & 40.84 & 30.87 & 31.36 & 23.16 & 30.40 & 29.57 & 35.10 & 24.87 & 37.18 & 25.82 & 24.46 & 20.03 & 19.85 & 35.05 \\
InternVL3-14B~\cite{chen2024expanding} & 38.33 & \colorbox[rgb]{1.0, 0.894, 0.812}{52.39} & 34.74 & 27.82 & 26.60 & 31.58 & 36.02 & 31.75 & 36.06 & 40.59 & 32.20 & 30.57 & 26.85 & 28.20 & 36.43 \\
InternVL3-38B~\cite{chen2024expanding} & \colorbox[rgb]{0.996, 1.0, 0.835}{41.43} & \colorbox[rgb]{0.996, 1.0, 0.835}{43.00} & 38.18 & \colorbox[rgb]{0.996, 1.0, 0.835}{35.71} & 38.21 & 33.90 & 36.88 & 37.39 & 36.06 & 40.48 & \colorbox[rgb]{0.996, 1.0, 0.835}{34.61} & 32.64 & \colorbox[rgb]{0.996, 1.0, 0.835}{32.38} & \colorbox[rgb]{0.996, 1.0, 0.835}{32.04} & 37.50 \\

Llama-4-Scout-17B-16E-Instruct~\cite{touvron2023llama} & \colorbox[rgb]{1.0, 0.894, 0.812}{44.46} & 42.27 & \colorbox[rgb]{0.996, 1.0, 0.835}{42.52} & 34.93 & 38.72 & \colorbox[rgb]{0.996, 1.0, 0.835}{40.04} & \colorbox[rgb]{1.0, 0.894, 0.812}{42.36} & 36.46 & \colorbox[rgb]{1.0, 0.894, 0.812}{40.26} & 41.04 & \colorbox[rgb]{1.0, 0.894, 0.812}{39.23} & \colorbox[rgb]{1.0, 0.894, 0.812}{41.64} & \colorbox[rgb]{1.0, 0.894, 0.812}{35.98} & \colorbox[rgb]{1.0, 0.894, 0.812}{37.76} & \colorbox[rgb]{1.0, 0.894, 0.812}{39.69} \\

LLaVA-NeXT-7B-Mistral~\cite{liu2024improved} & 27.26 & 15.28 & 37.68 & 25.83 & 26.86 & 31.15 & 26.92 & 22.51 & 28.92 & 33.22 & 20.88 & 26.58 & 24.60 & 26.89 & 29.77 \\
LLaVA-NeXT-8B~\cite{liu2024improved} & 30.05 & 36.39 & 42.22 & 34.00 & 33.72 & 40.00 & 30.83 & 31.49 & 22.10 & 35.40 & 28.38 & 34.12 & 31.50 & 29.41 & 31.58 \\
LLaVA-NeXT-13B~\cite{liu2024improved} & 9.33 & 12.82 & 12.28 & 20.57 & 20.24 & 13.78 & 21.92 & 16.35 & 4.48 & 14.28 & 12.42 & 18.63 & 19.01 & 20.41 & 15.49 \\
LLaVA-NeXT-34B~\cite{liu2024improved} & 23.68 & 32.11 & 29.82 & 32.52 & 37.70 & 14.76 & 27.53 & \colorbox[rgb]{0.996, 1.0, 0.835}{37.69} & \colorbox[rgb]{0.996, 1.0, 0.835}{36.78} & 35.80 & 25.36 & 27.64 & 27.99 & 26.49 & 35.91 \\

Qwen2.5-VL-3B-Instruct~\cite{bai2023qwen} & 31.42 & 38.15 & 30.27 & 27.32 & 29.07 & 24.26 & 23.70 & 25.20 & 24.37 & 28.95 & 21.09 & 22.85 & 23.58 & 23.40 & 28.03 \\
Qwen2.5-VL-7B-Instruct~\cite{bai2023qwen} & 34.43 & 38.22 & 32.30 & 25.35 & 33.20 & 35.48 & 26.48 & 23.86 & 27.62 & 31.75 & 26.18 & 24.40 & 25.43 & 25.73 & 26.19 \\
Qwen2.5-VL-32B-Instruct~\cite{bai2023qwen} & 30.30 & 39.91 & 30.17 & 25.46 & 31.06 & 36.23 & 28.53 & 28.21 & 31.32 & 27.25 & 30.85 & 27.16 & 29.90 & 30.49 & 30.45 \\

TinyLLaVA-1.5B~\cite{zhou2024tinyllava} & 11.87 & 24.27 & 0.40 & - & - & - & - & 2.70 & 0.71 & 11.36 & - & - & - & - & - \\
TinyLLaVA-2B~\cite{zhou2024tinyllava} & - & - & - & - & - & - & - & - & - & - & - & - & - & - & - \\
TinyLLaVA-3.1B~\cite{zhou2024tinyllava} & 3.19 & 5.22 & 7.53 & - & 1.98 & 1.22 & - & - & - & - & 1.12 & 0.17 & 5.80 & 4.20 & 0.70 \\

\Xhline{1.5pt}
\end{tabular}
}
\end{table*}

\section{Further Discussion}
\label{app:further_discussion}
\subsection{Does the reasoning ability of VLMs improve as the parameters increase?}
\label{app:reasoning_ability}
\begin{figure*}[t]
    \centering
    \includegraphics[width=\linewidth]{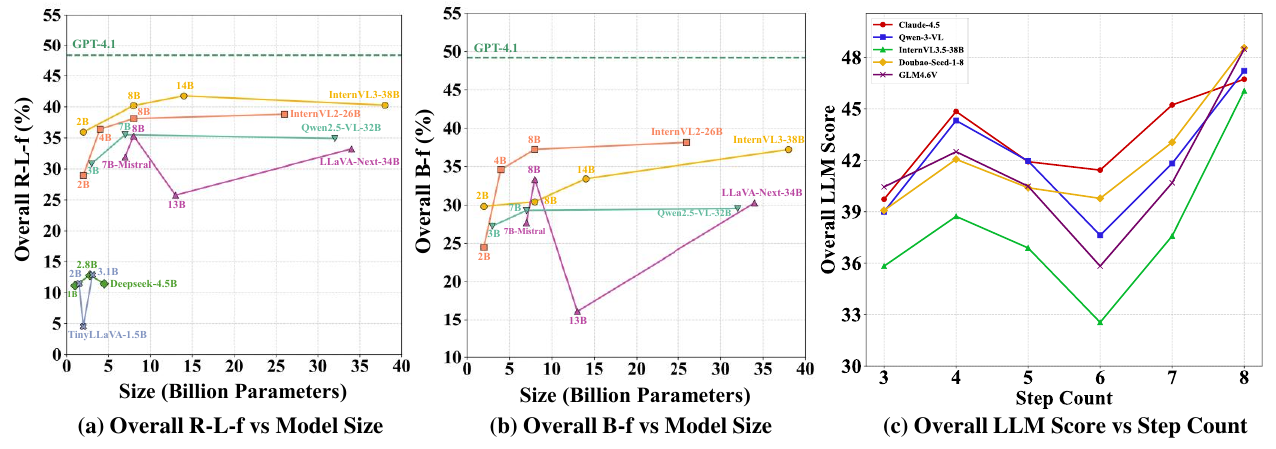}
    \caption{ROUGE-L F1 (a) and BERTScore F1 (b) vary with model parameter size, and LLM scores (c) vary with reasoning step counts.}
    \label{fig:score_by_para}
\end{figure*}

As shown in Figure~\ref{fig:score_by_para} (a) and (b), except for InternVL2\&3 and Qwen2.5-VL, whose reasoning capabilities improve with an increase in parameters when the number of parameters is small, the reasoning capabilities of other models remain relatively stable once the number of parameters reaches a certain threshold. This suggests that increasing the number of parameters primarily enhances the knowledge capacity of models, with limited impact on improving specific reasoning abilities. 

\subsection{Are VLMs capable of handling deep reasoning tasks?}
\label{app:vlms_deep_reasoning}
\begin{figure*}[t]
    \centering
    \includegraphics[width=\linewidth]{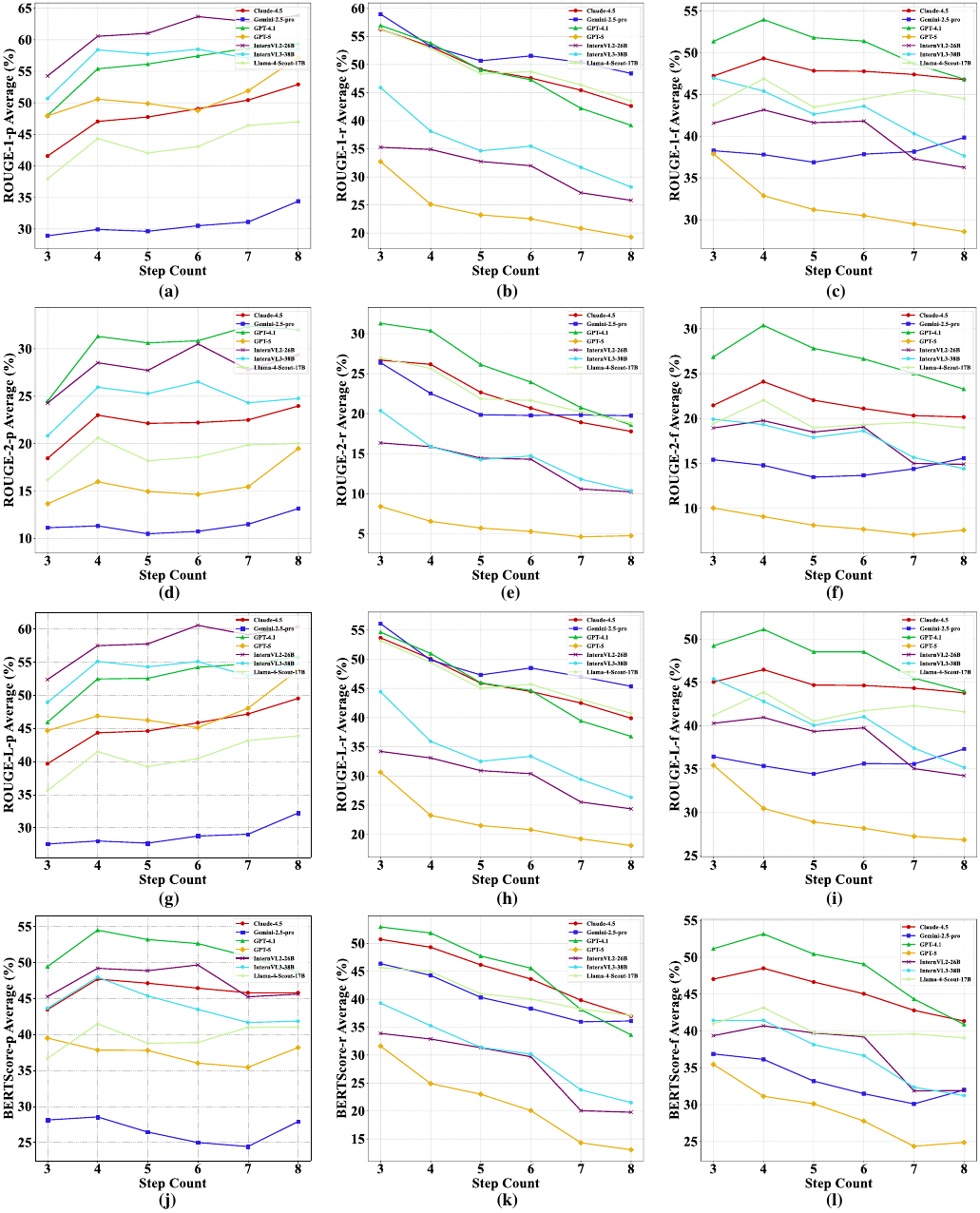}
    \caption{Overall performance of several VLMs using ROUGE-1, ROUGE-2, ROUGE-L and BERTScore across different step counts.}
    \label{fig:score_by_step}
\end{figure*}

As shown in Figure~\ref{fig:score_by_step}, the performance of various models across different ROUGE and BERTScore metrics (precision, recall, and F1 score) demonstrates how reasoning steps impact model performance. The recall scores notably decrease as the number of reasoning steps increases, suggesting that longer reasoning chains lead to more gaps in the model’s coverage. This implies that while the model may accurately capture key information in the initial steps, it struggles to maintain comprehensive reasoning over longer chains. Conversely, precision remains relatively stable, and in some cases, even increases as the number of steps grows. This indicates that while the models are able to generate more specific and accurate responses in later reasoning steps, they may be sacrificing breadth for accuracy, leading to fewer but more focused answers. Overall, these changes result in a decrease in F1 score as the reasoning steps increase. This is because F1 score balances precision and recall, and the decline in recall outweighs the relatively stable performance of precision. This trend underscores a common challenge in reasoning tasks that models may be able to generate precise answers, but as the complexity of the reasoning process increases, they struggle to maintain comprehensive coverage, leading to a trade-off between precision and recall.

\section{Dimension Details and Case Study}
\label{app:dimension_details}
\begin{figure*}[ht]
    \centering
    \includegraphics[width=\linewidth]{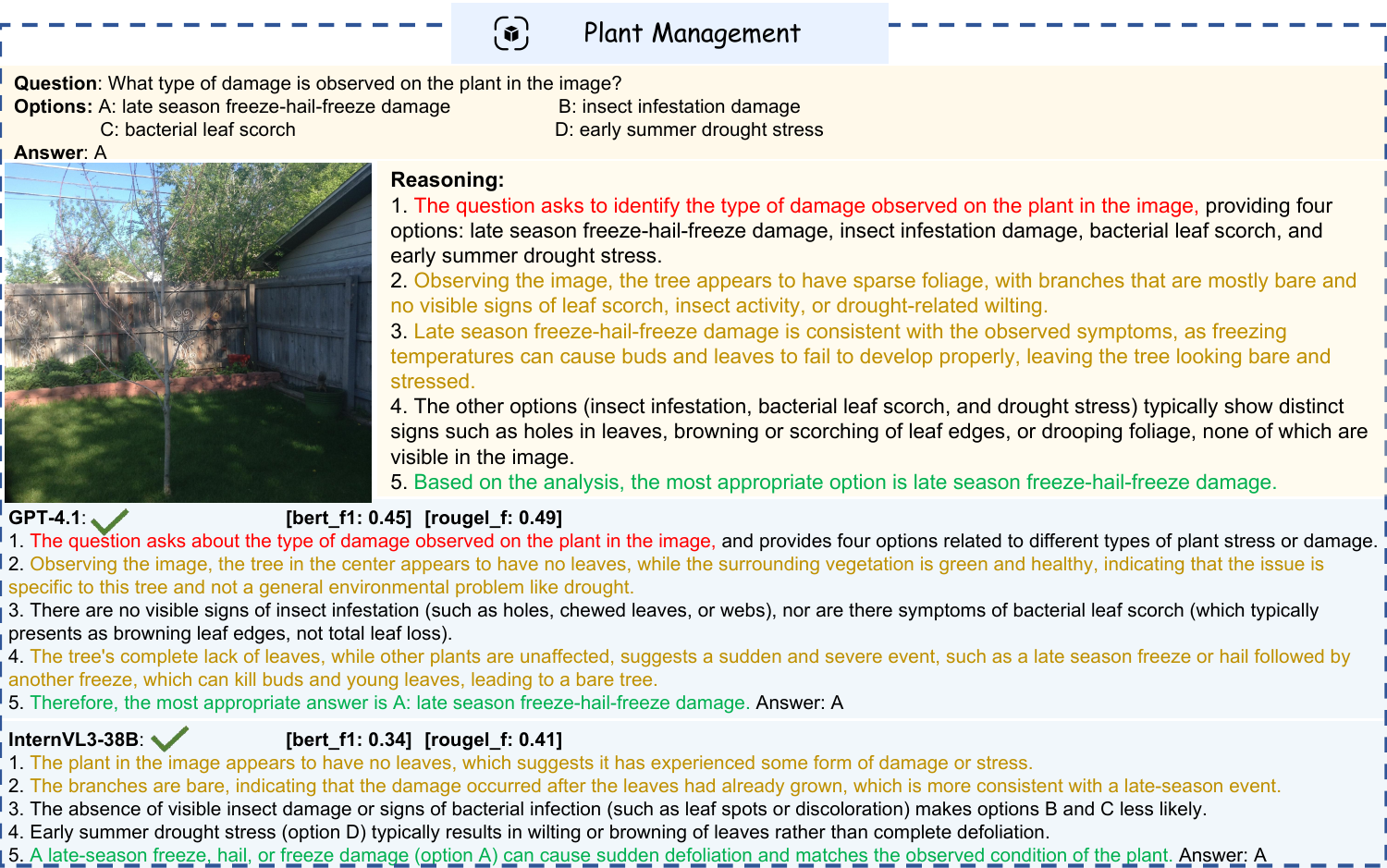}
    \caption{A case of the Plant Management task with responses from GPT-4.1 and InternVL3-38B. \textcolor{red}{Red text} denotes question extraction, \textcolor{mybrown}{brown text} represents key analysis, and \textcolor{mygreen}{green text} indicates answer summary.}
    \label{fig:Plant Management case}
\end{figure*}

\begin{figure*}[t]
    \centering
    \includegraphics[width=\linewidth]{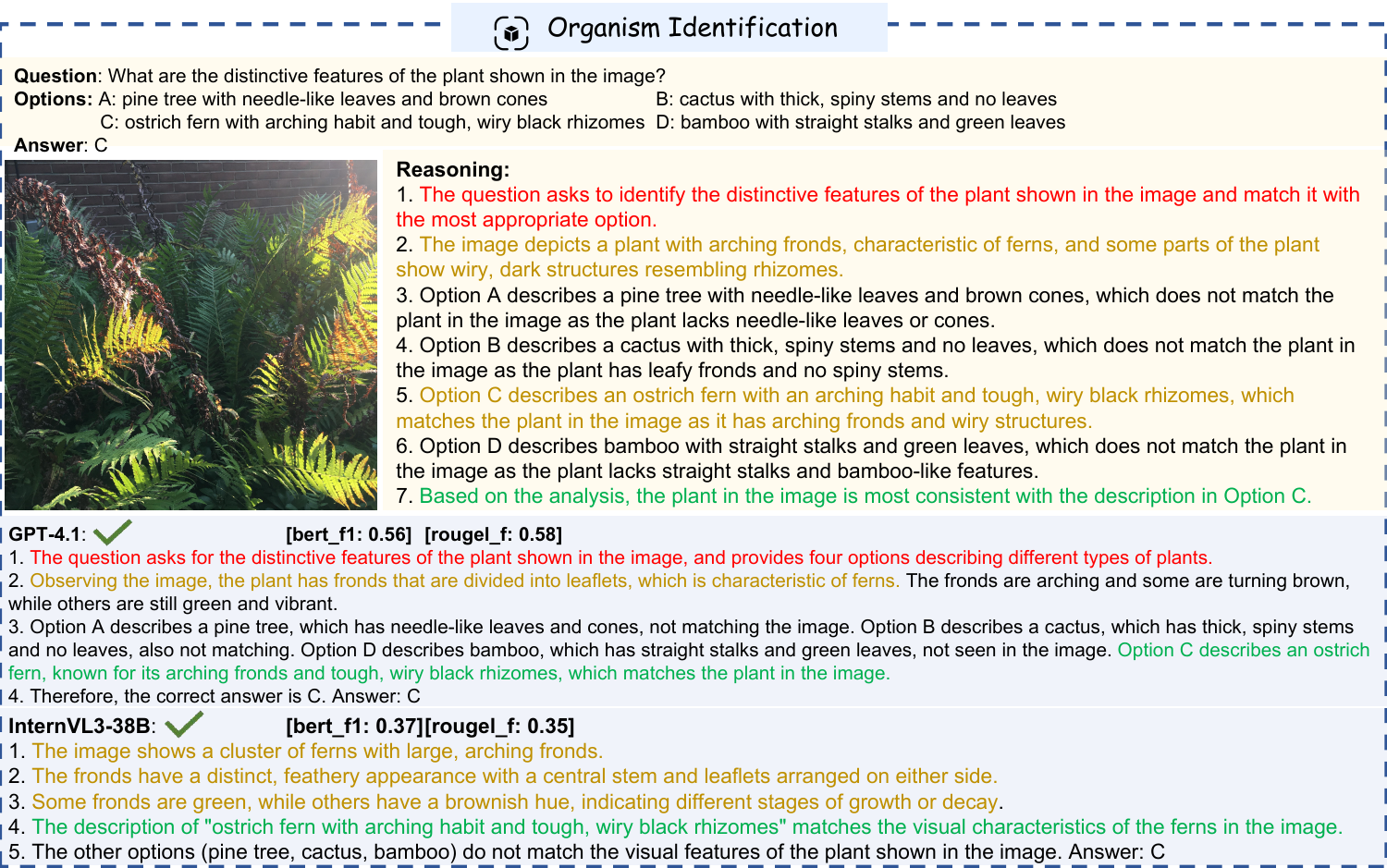}
    \caption{A case of the Organism Identification task with responses from GPT-4.1 and InternVL3-38B. \textcolor{red}{Red text} denotes question extraction, \textcolor{mybrown}{brown text} represents key analysis, and \textcolor{mygreen}{green text} indicates answer summary.
 }
    \label{fig:Organism Identification case}
\end{figure*}

\begin{figure*}[t]
    \centering
    \includegraphics[width=\linewidth]{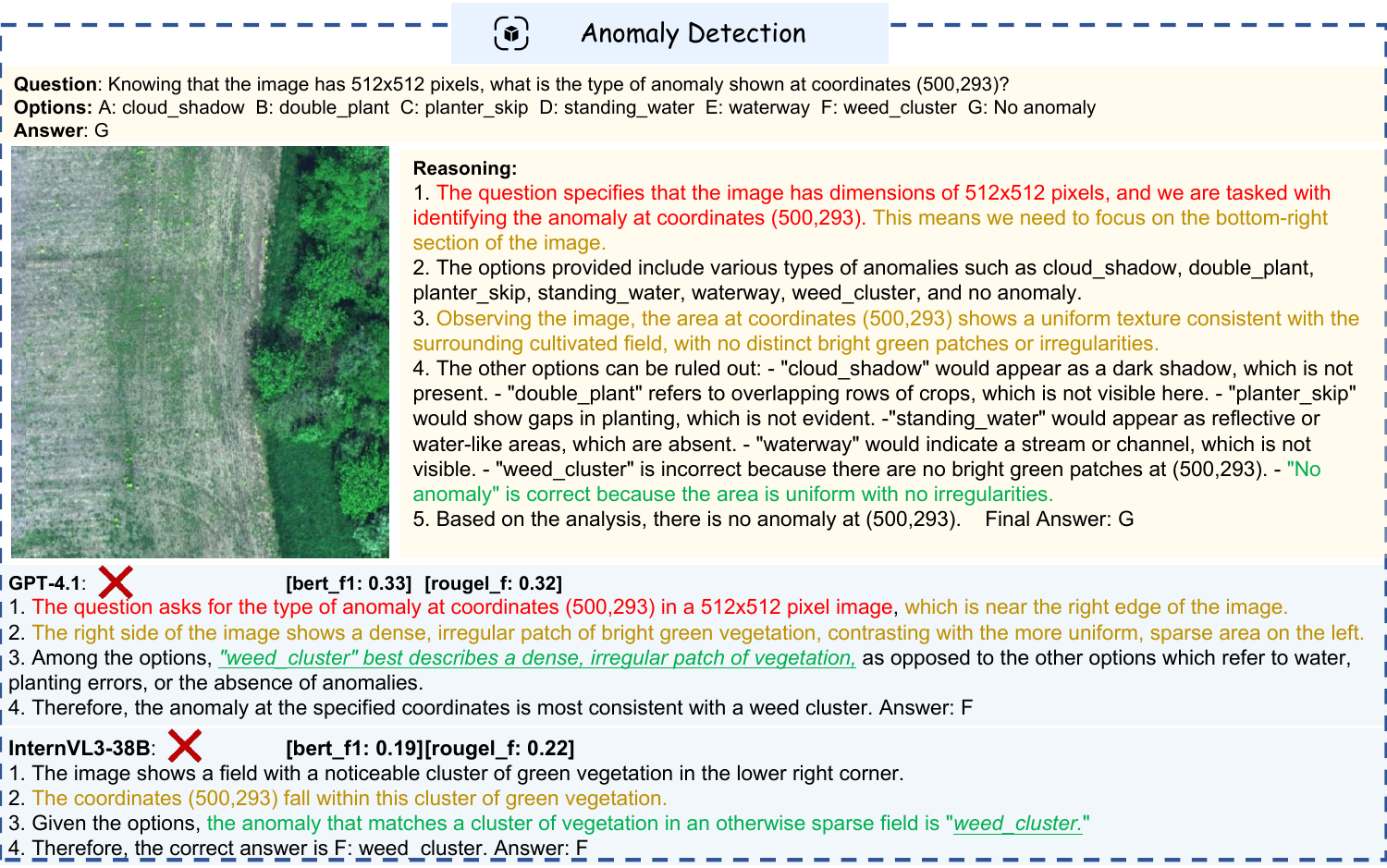}
    \caption{A case of the Anomaly Detection task with responses from GPT-4.1 and InternVL3-38B. \textcolor{red}{Red text} denotes question extraction, \textcolor{mybrown}{brown text} represents key analysis, and \textcolor{mygreen}{green text} indicates answer summary. 
    \underline{\textit{Underlined text}} highlights erroneous reasoning: GPT-4.1 incorrectly associated the dense green vegetation on the right side of the image with the coordinates (500,293), failing to locate the target area and misclassifying the anomaly. InternVL3-38B also misidentified the target coordinates as falling within the green vegetation cluster, leading to the wrong classification.
}
    \label{fig:Anomaly Detection case}
\end{figure*}

\begin{figure*}[t]
    \centering
    \includegraphics[width=\linewidth]{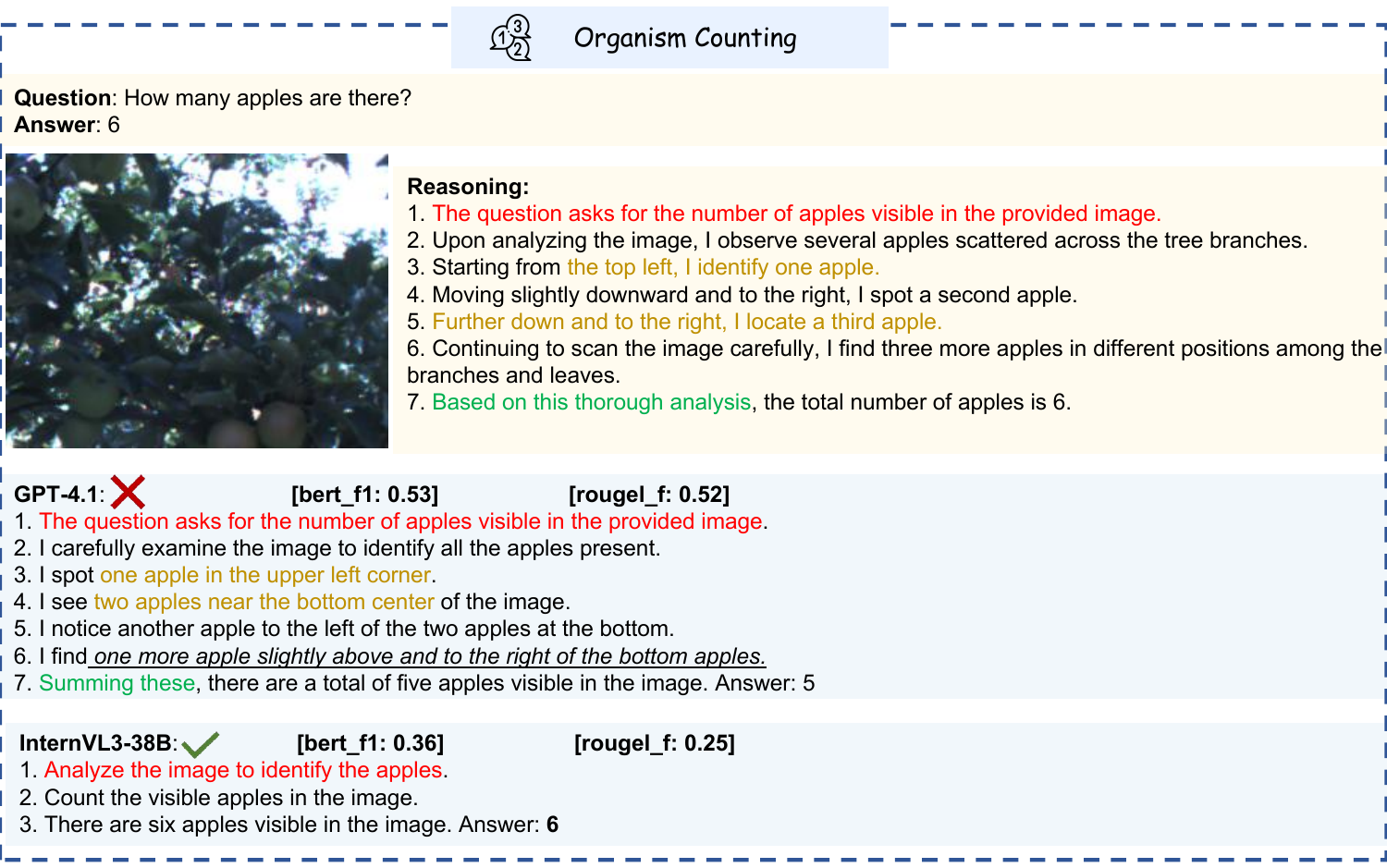}
    \caption{A case of the Organism Counting task with responses from GPT-4.1 and InternVL3-38B. \textcolor{red}{Red text} denotes question extraction, \textcolor{mybrown}{brown text} represents key analysis, and \textcolor{mygreen}{green text} indicates answer summary. 
    \underline{\textit{Underlined text}} highlights erroneous reasoning: GPT-4.1 made an error in counting the apples in the image, resulting in an incorrect answer of 5 instead of the correct count of 6.
}
    \label{fig:Organism Counting case}
\end{figure*}

\begin{figure*}[ht]
    \centering
    \includegraphics[width=\linewidth]{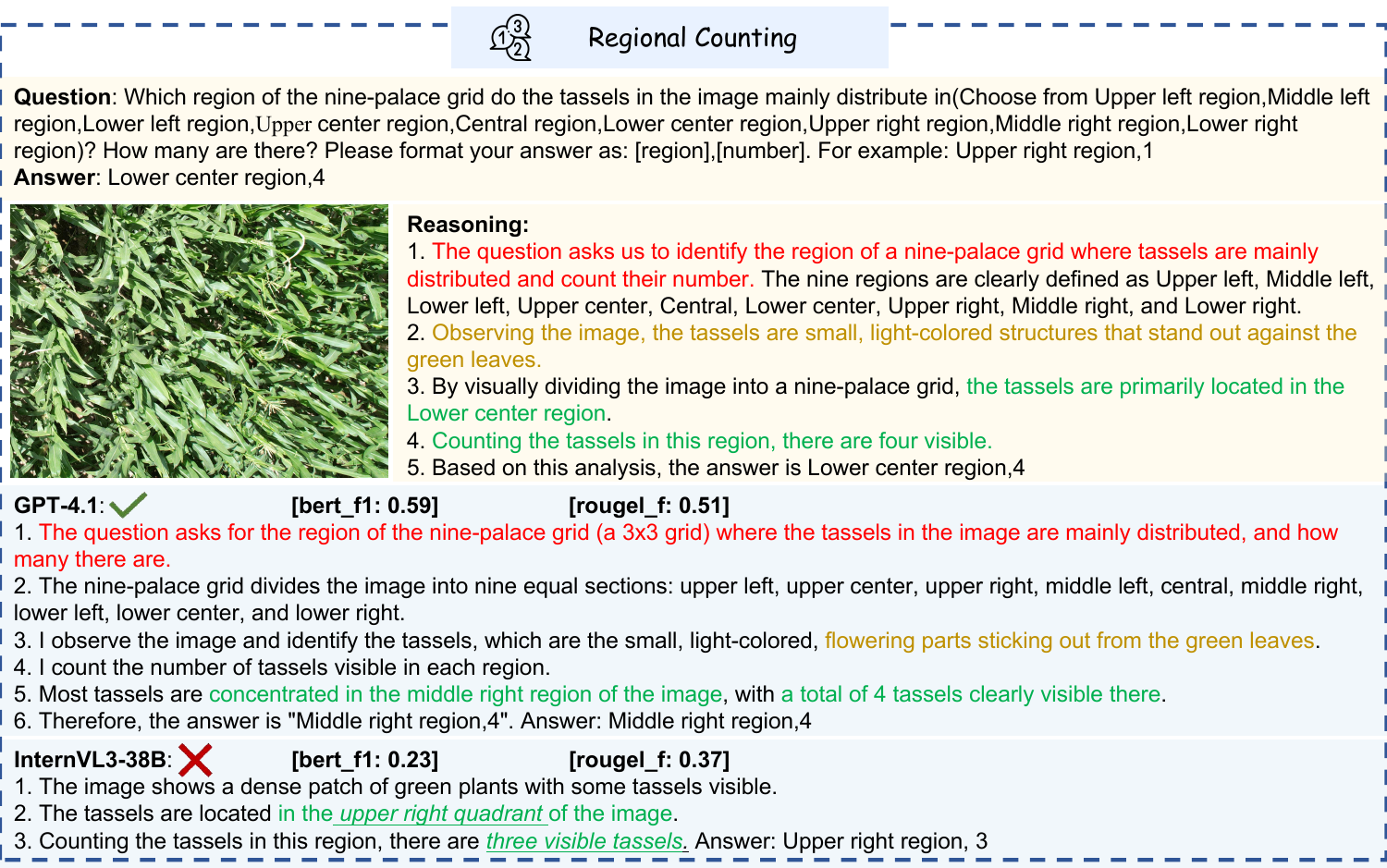}
    \vspace{-1.5em}
    \caption{A case of the Regional Counting task with responses from GPT-4.1 and InternVL3-38B. \textcolor{red}{Red text} denotes question extraction, \textcolor{mybrown}{brown text} represents key analysis, and \textcolor{mygreen}{green text} indicates answer summary.
    \underline{\textit{Underlined text}} highlights erroneous reasoning:  InternVL3-38B made two errors: it misidentified the tassels’ main distribution region as the upper right quadrant and incorrectly counted only three tassels, leading to the wrong answer of “Upper right region, 3”.}
    \vspace{-1.25em}
    \label{fig:Regional Counting case}
\end{figure*}

\begin{figure*}[ht]
    \centering
    \includegraphics[width=\linewidth]{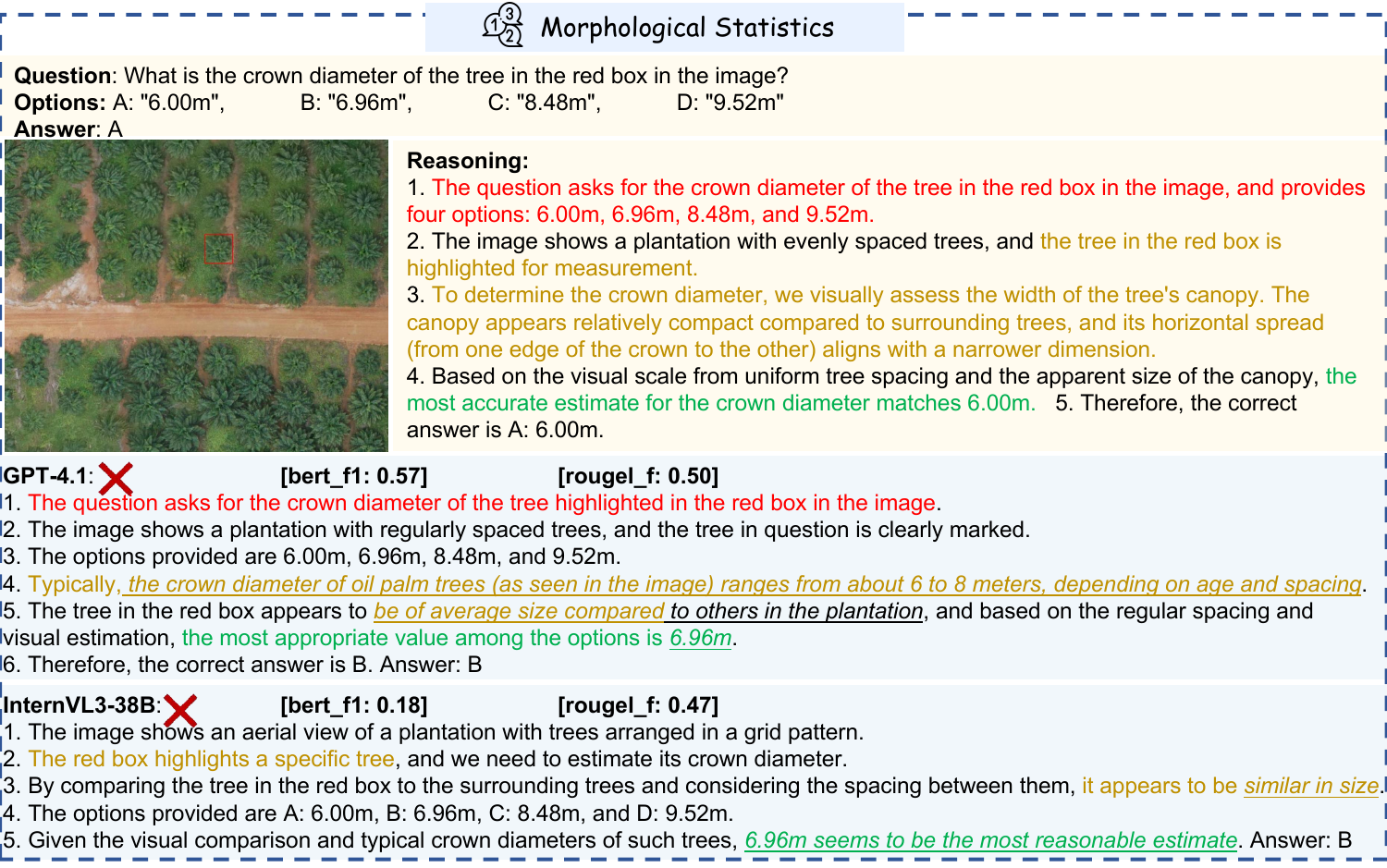}
    \vspace{-1.5em}
    \caption{A case of the Morphological Statistics task with responses from GPT-4.1 and InternVL3-38B. \textcolor{red}{Red text} denotes question extraction, \textcolor{mybrown}{brown text} represents key analysis, and \textcolor{mygreen}{green text} indicates answer summary. \underline{\textit{Underlined text}} highlights erroneous reasoning: both models incorrectly judged the tree in the red box as average-sized, leading to the wrong crown diameter estimate of 6.96m instead of the correct 6.00m.}
    \vspace{-1.25em}
    \label{fig:Morphological Statistics case}
\end{figure*}

\begin{figure*}[ht]
    \centering
    \includegraphics[width=\linewidth]{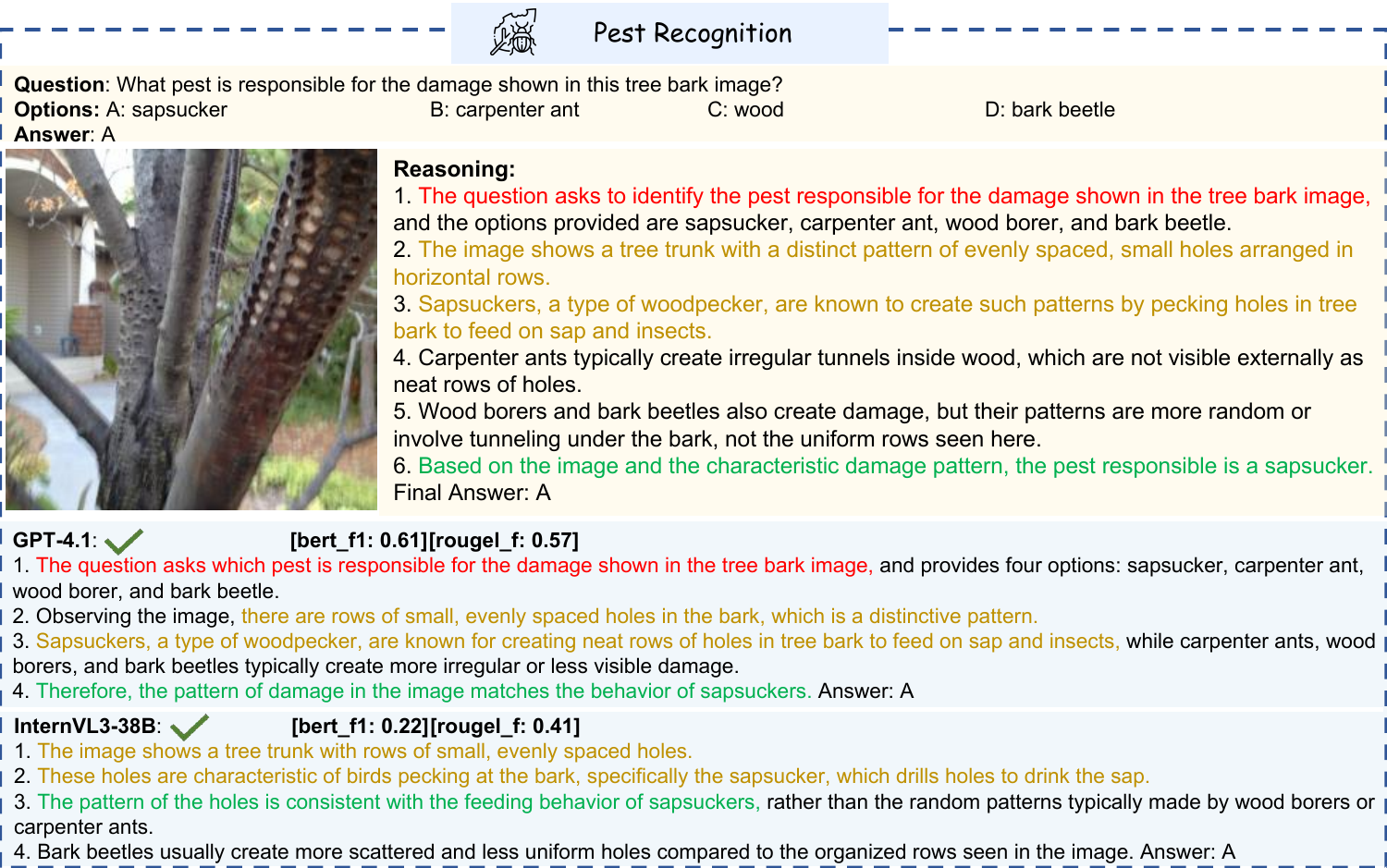}
    \vspace{-1.5em}
    \caption{A case of the Pest Recognition task with responses from GPT-4.1 and InternVL3-38B. \textcolor{red}{Red text} denotes question extraction, \textcolor{mybrown}{brown text} represents key analysis, and \textcolor{mygreen}{green text} indicates answer summary.}
    \vspace{-1.25em}
    \label{fig:Pest Recognition case}
\end{figure*}

\begin{figure*}[ht]
    \centering
    \includegraphics[width=\linewidth]{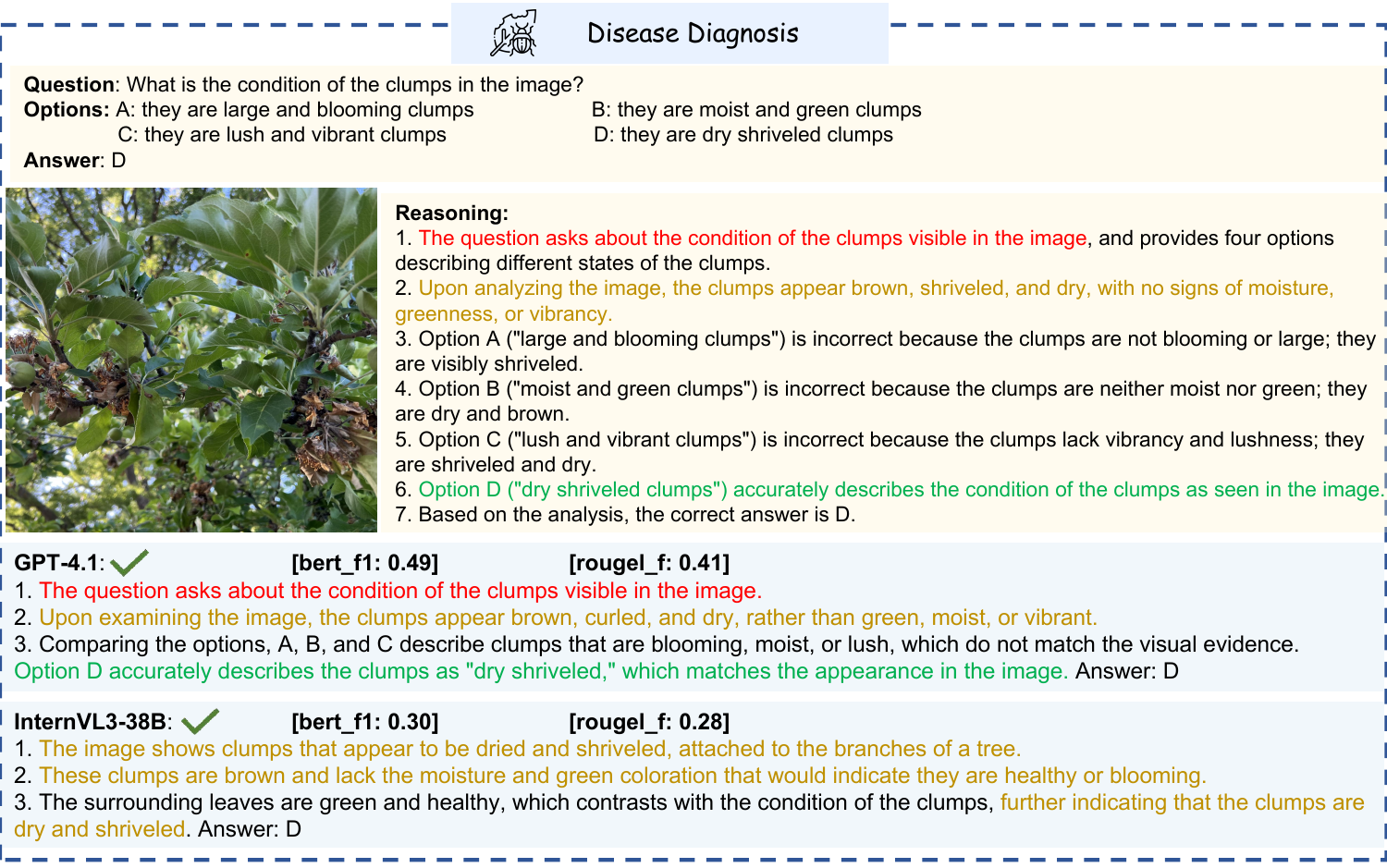}
    \vspace{-1.5em}
    \caption{A case of the Disease Diagnosis task with responses from GPT-4.1 and InternVL3-38B. \textcolor{red}{Red text} denotes question extraction, \textcolor{mybrown}{brown text} represents key analysis, and \textcolor{mygreen}{green text} indicates answer summary.}
    \vspace{-1.25em}
    \label{fig:Disease Diagnosis case}
\end{figure*}

\begin{figure*}[ht]
    \centering
    \includegraphics[width=\linewidth]{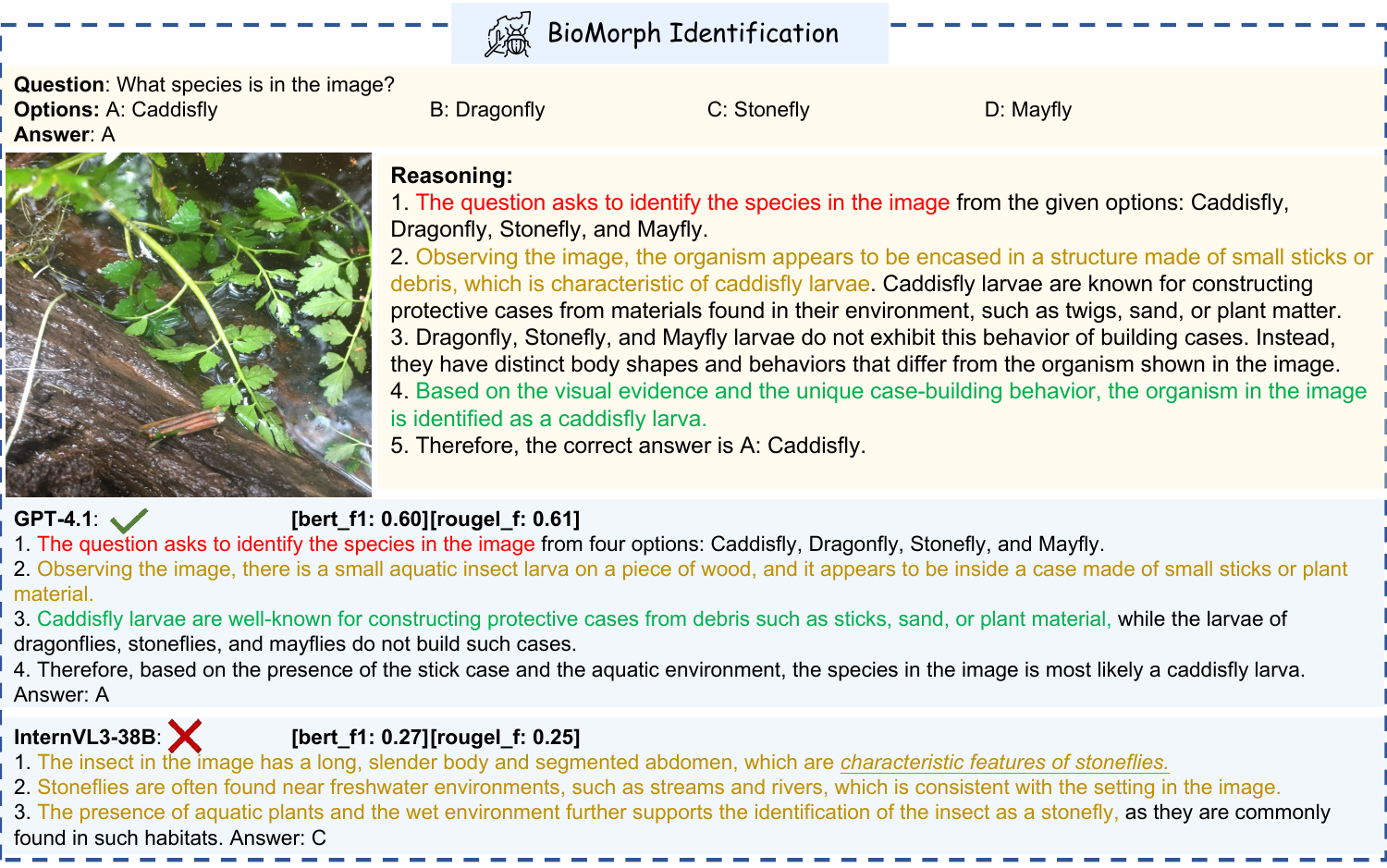}
    \vspace{-1.5em}
    \caption{A case of the BioMorph Identification task with responses from GPT-4.1 and InternVL3-38B. \textcolor{red}{Red text} denotes question extraction, \textcolor{mybrown}{brown text} represents key analysis, and \textcolor{mygreen}{green text} indicates answer summary.
    \underline{\textit{Underlined text}} highlights erroneous reasoning:InternVL3-38B misidentified the insect’s key features, ignoring the caddisfly’s characteristic case-building behavior and incorrectly associating the specimen with stonefly traits.}
    \vspace{-1.25em}
    \label{fig:BioMorph Identification case}
\end{figure*}

\begin{figure*}[ht]
    \centering
    \includegraphics[width=\linewidth]{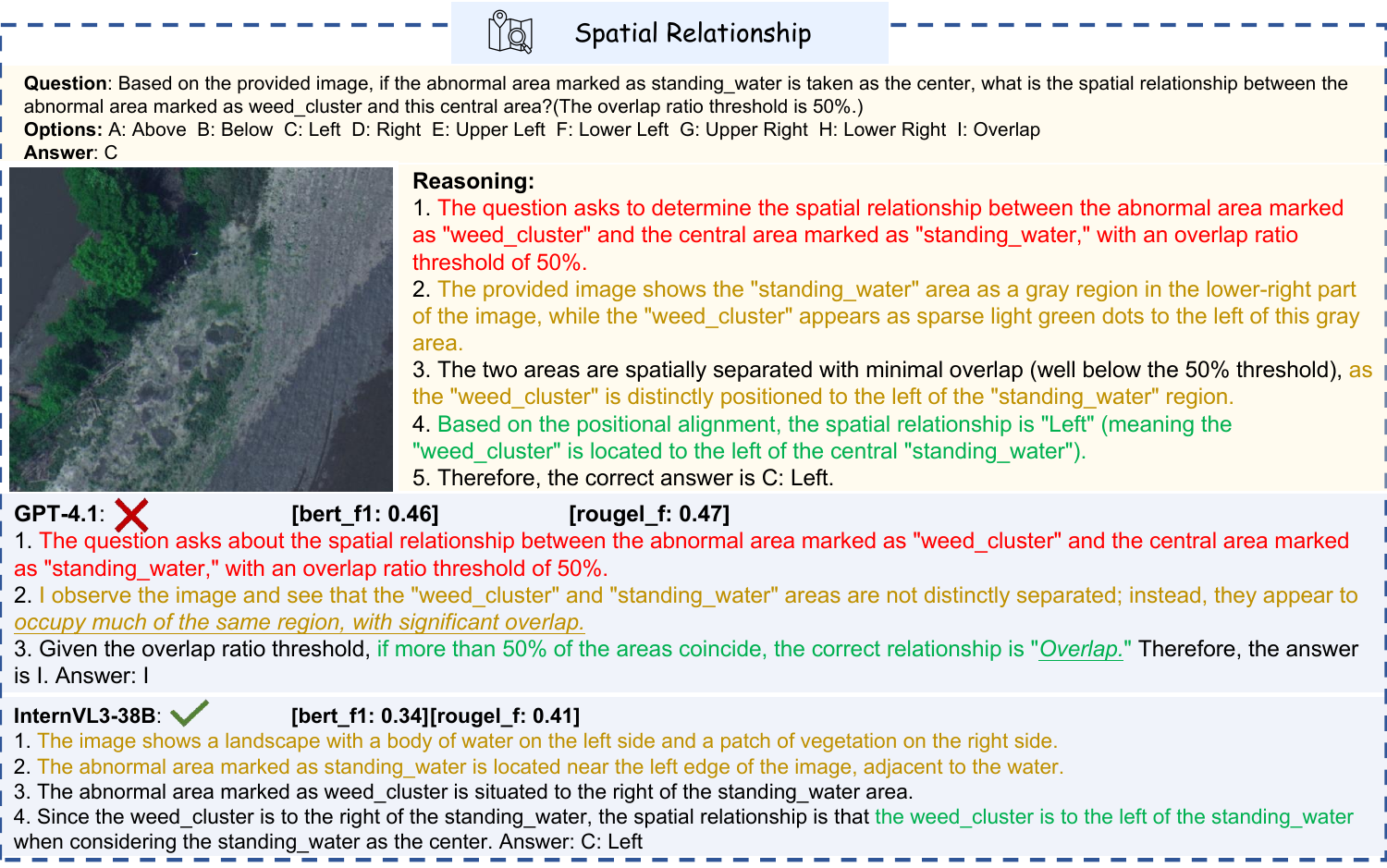}
    \vspace{-1.5em}
    \caption{A case of the Spatial Relationship task with responses from GPT-4.1 and InternVL3-38B. \textcolor{red}{Red text} denotes question extraction, \textcolor{mybrown}{brown text} represents key analysis, and \textcolor{mygreen}{green text} indicates answer summary. 
    \underline{\textit{Underlined text}} highlights erroneous reasoning:GPT-4.1’s reasoning misjudged the two regions as having significant overlap, ignoring their clear spatial separation, which led to the wrong "Overlap" conclusion.
 }
    \vspace{-1.25em}
    \label{fig:Spatial Relationship case}
\end{figure*}

\begin{figure*}[ht]
    \centering
    \includegraphics[width=\linewidth]{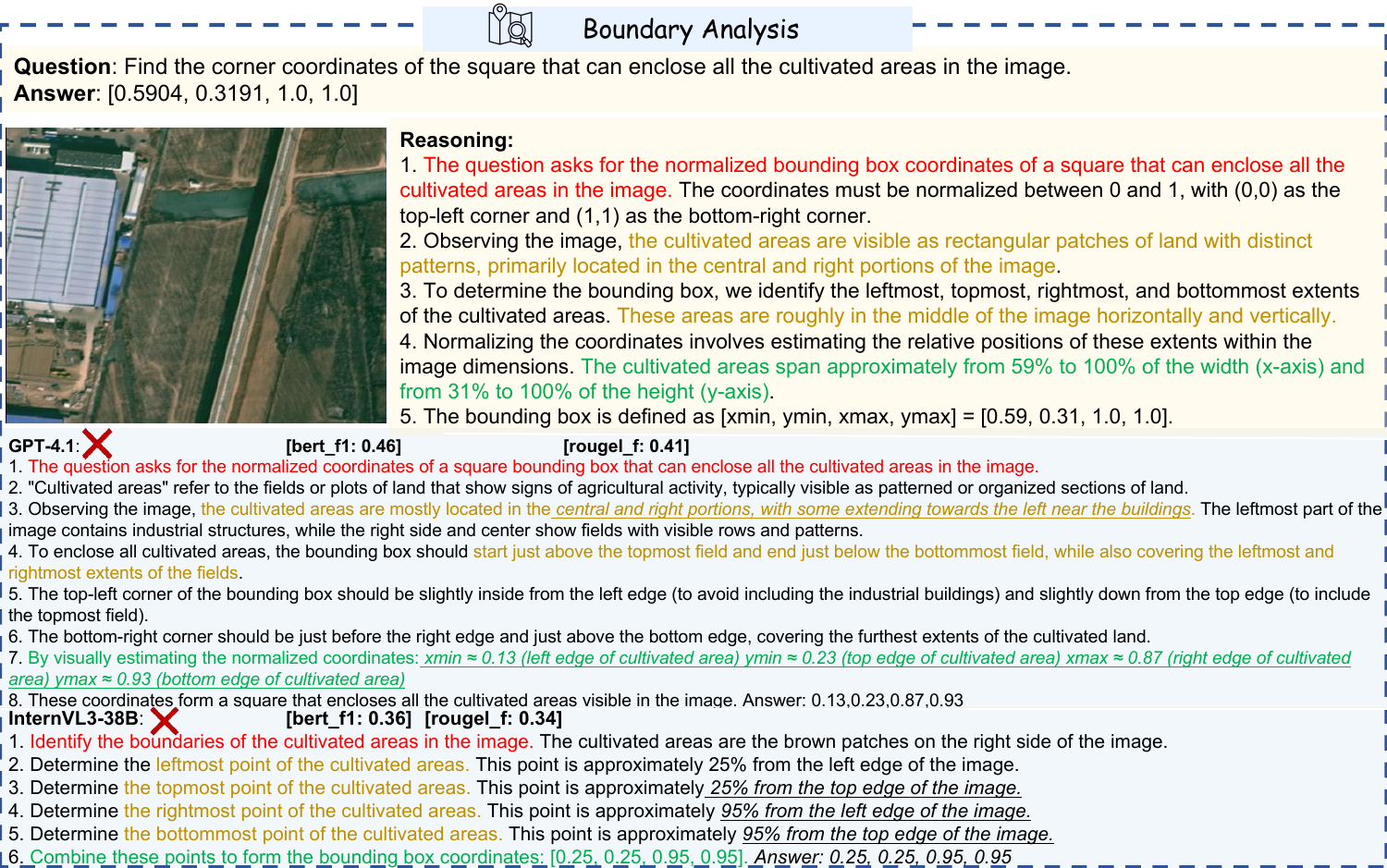}
    \vspace{-1.5em}
    \caption{A case of the Boundary Analysis task with responses from GPT-4.1 and InternVL3-38B. \textcolor{red}{Red text} denotes question extraction, \textcolor{mybrown}{brown text} represents key analysis, and \textcolor{mygreen}{green text} indicates answer summary. Both models failed to produce the correct bounding box coordinates. 
    \underline{\textit{Underlined text}} highlights erroneous reasoning:GPT-4.1 incorrectly identified the leftmost boundary of the cultivated areas, while InternVL3-38B misestimated all four corner coordinates of the bounding box, resulting in an incorrect final box.
 }
    \vspace{-1.25em}
    \label{fig:Boundary Analysis case}
\end{figure*}

\begin{figure*}[ht]
    \centering
    \includegraphics[width=\linewidth]{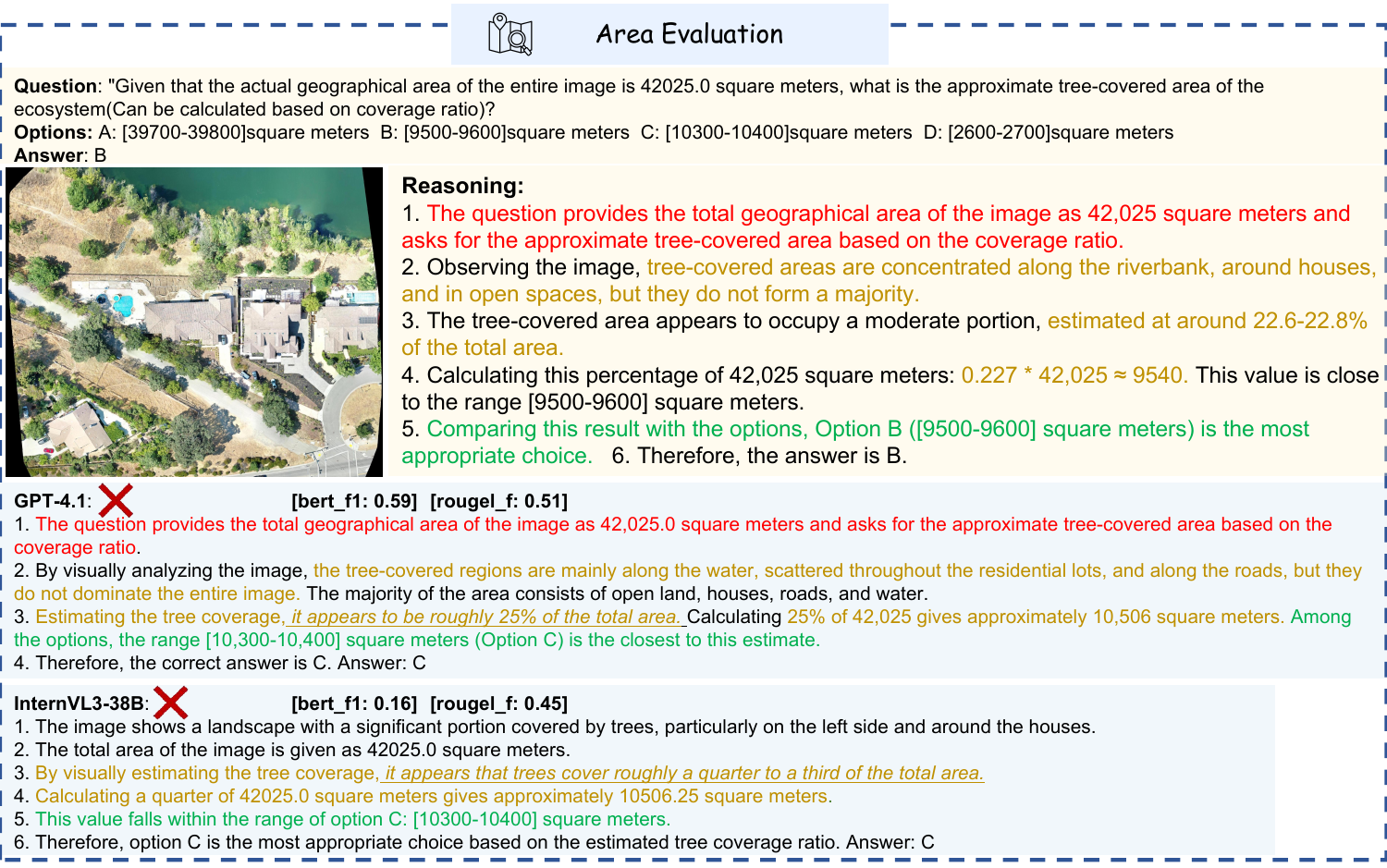}
    \vspace{-1.5em}
    \caption{A case of the Area Evaluation task with responses from GPT-4.1 and InternVL3-38B. \textcolor{red}{Red text} denotes question extraction, \textcolor{mybrown}{brown text} represents key analysis, and \textcolor{mygreen}{green text} indicates answer summary.
    \underline{\textit{Underlined text}} highlights erroneous reasoning:GPT-4.1 misestimated the tree coverage ratio and gave an incorrect answer. InternVL-3-38B overestimated the tree coverage ratio and selected the wrong option.}
    \vspace{-1.25em}
    \label{fig:Area Evaluation case}
\end{figure*}

\begin{figure*}[ht]
    \centering
    \includegraphics[width=\linewidth]{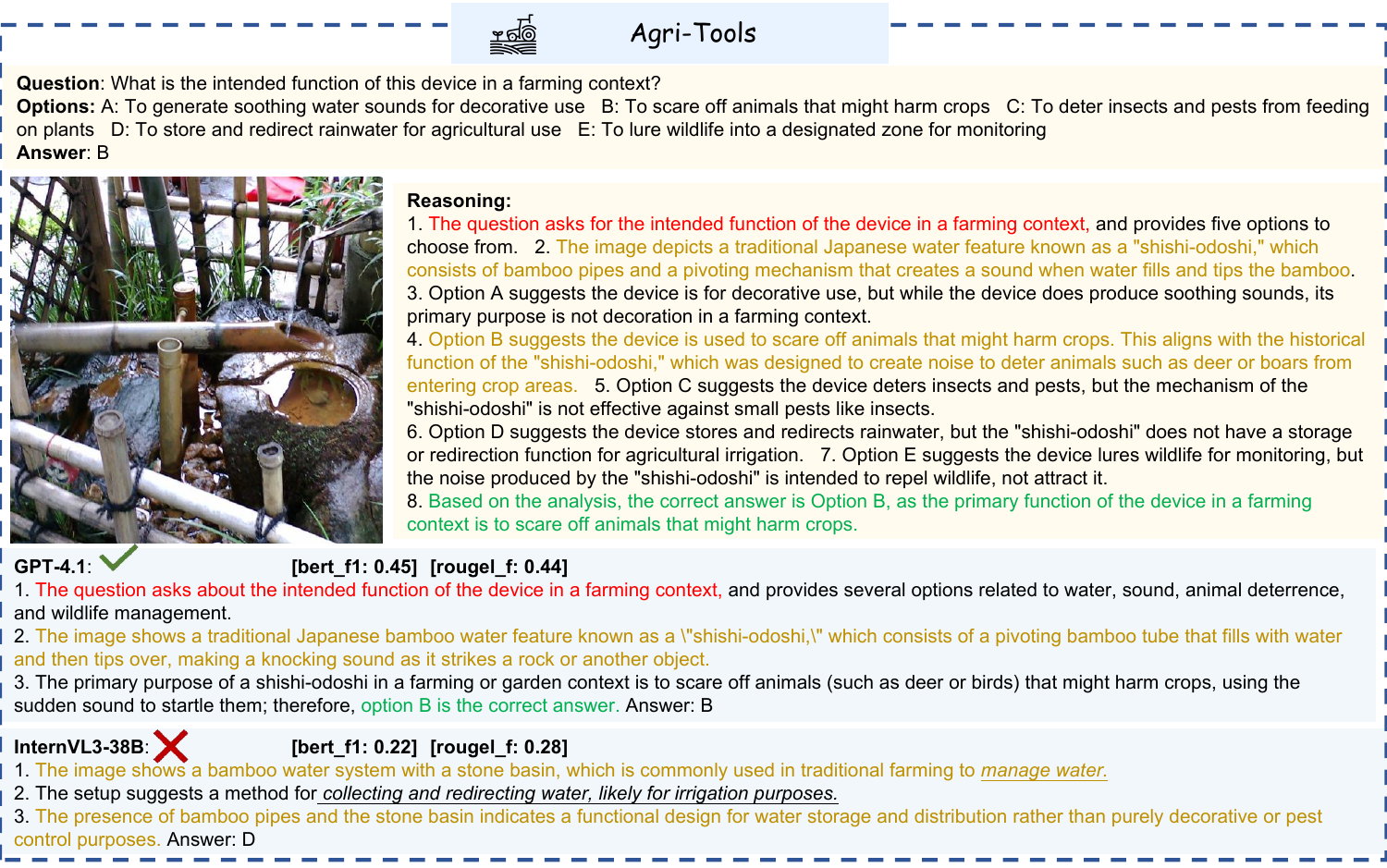}
    \vspace{-1.5em}
    \caption{A case of the Agri-Tools task with responses from GPT-4.1 and InternVL3-38B. \textcolor{red}{Red text} denotes question extraction, \textcolor{mybrown}{brown text} represents key analysis, and \textcolor{mygreen}{green text} indicates answer summary. 
    \underline{\textit{Underlined text}} highlights erroneous reasoning:InternVL3-38B misinterpreted the shishi-odoshi’s core function, focusing only on its water storage/redirection feature while ignoring its primary purpose of scaring animals.
 }
    \vspace{-1.25em}
    \label{fig:Agri-Tools case}
\end{figure*}

\begin{figure*}[ht]
    \centering
    \includegraphics[width=\linewidth]{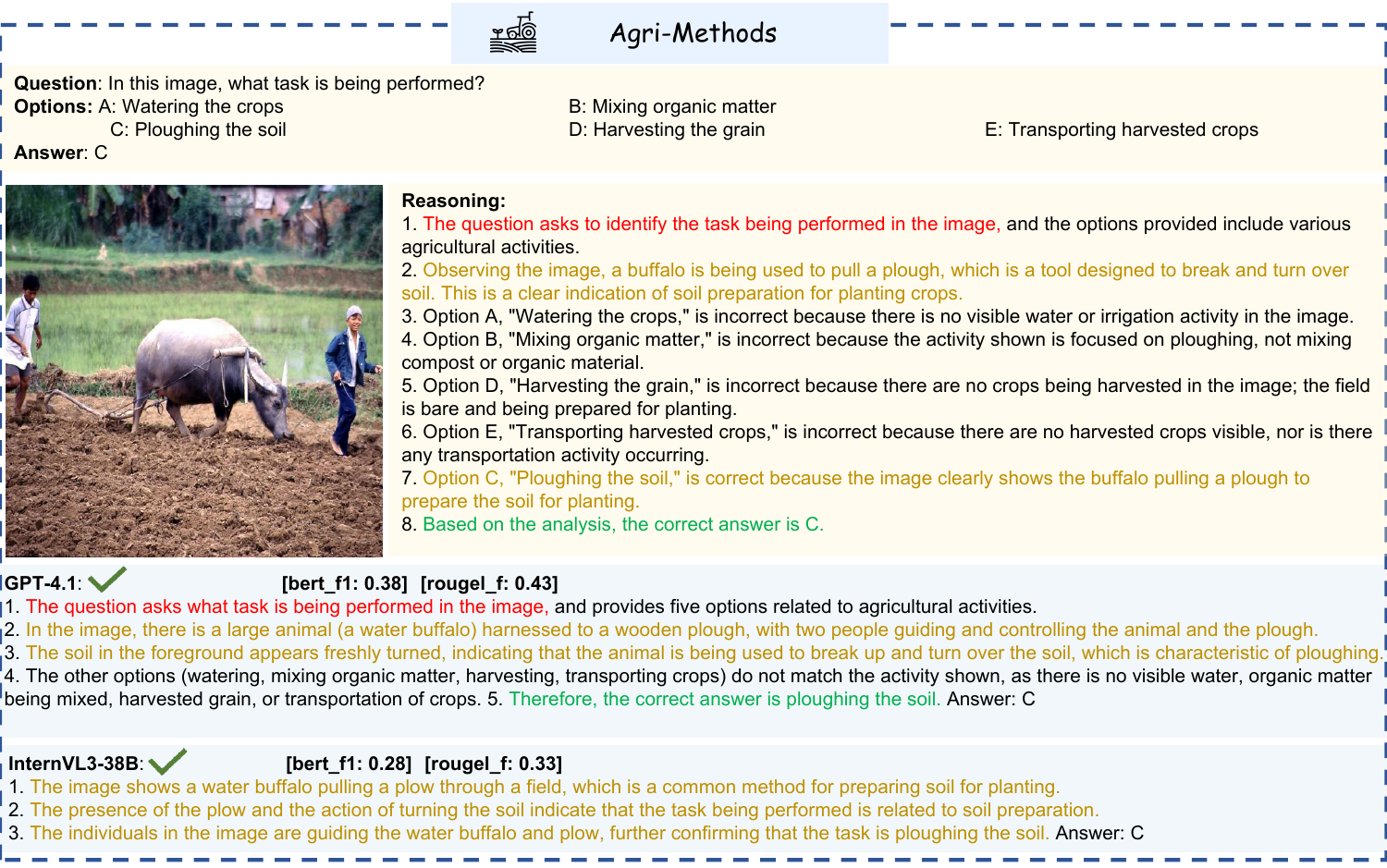}
    \vspace{-1.5em}
    \caption{A case of the Agri-Methods task with responses from GPT-4.1 and InternVL3-38B. \textcolor{red}{Red text} denotes question extraction, \textcolor{mybrown}{brown text} represents key analysis, and \textcolor{mygreen}{green text} indicates answer summary.}
    \vspace{-1.25em}
    \label{fig:Agri-Methods case}
\end{figure*}

\begin{figure*}[ht]
    \centering
    \includegraphics[width=\linewidth]{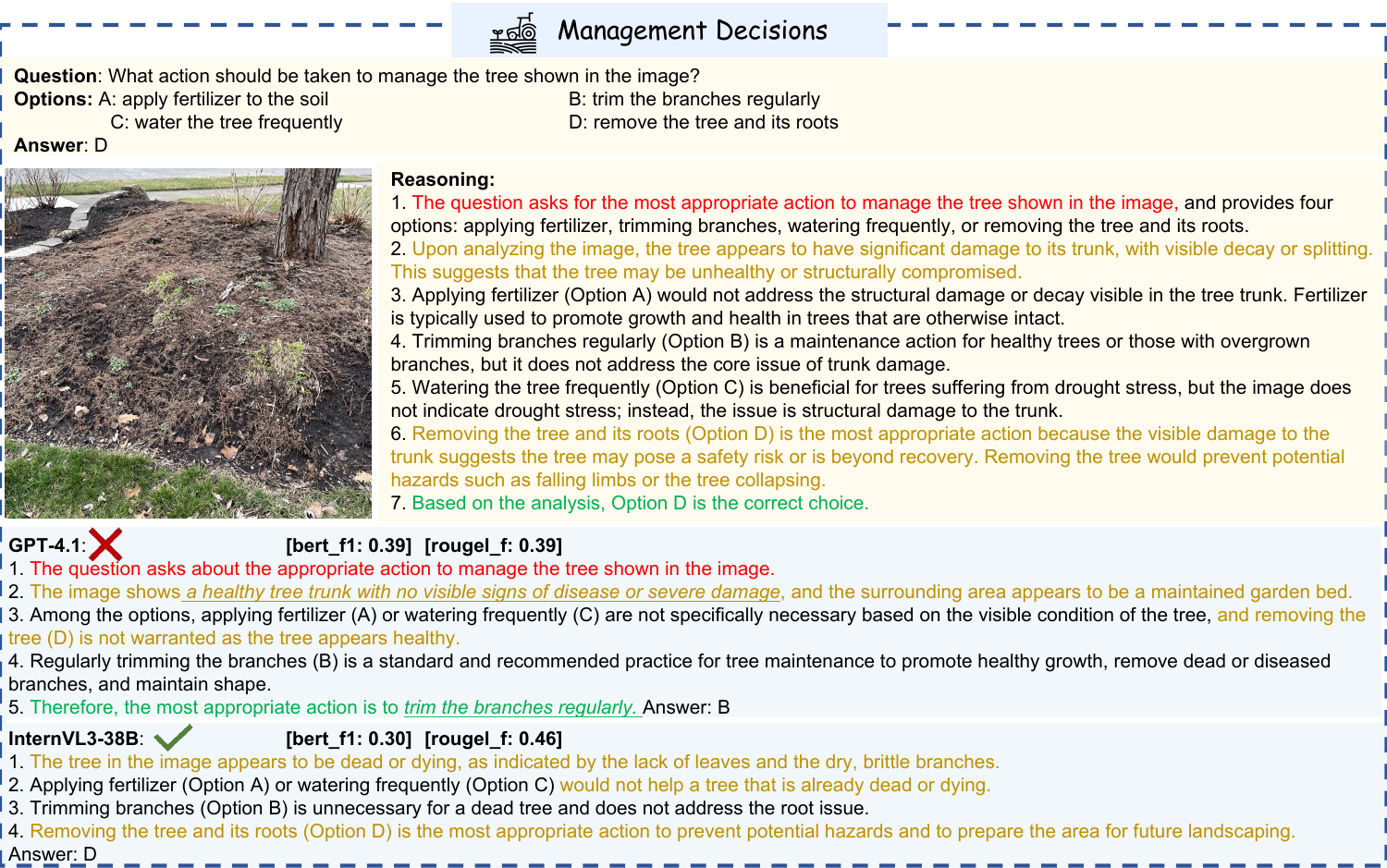}
    \vspace{-1.5em}
    \caption{A case of the Management Decisions task with responses from GPT-4.1 and InternVL3-38B. \textcolor{red}{Red text} denotes question extraction, \textcolor{mybrown}{brown text} represents key analysis, and \textcolor{mygreen}{green text} indicates answer summary. 
    \underline{\textit{Underlined text}} highlights erroneous reasoning:GPT-4.1 misjudged the tree’s condition as healthy, ignoring visible trunk damage and decay, and incorrectly recommended routine pruning instead of addressing the hazard.}
    \vspace{-1.25em}
    \label{fig:Management Decisions case}
\end{figure*}

In this section, we present a series of task-specific question examples from AgroCoT, paired with responses generated by GPT-4.1 and InternVL3-38B (refer to Figure~\ref{fig:Plant Management case} through Figure~\ref{fig:Management Decisions case} for details). For clarity, text colors in the figures are assigned specific meanings: \textcolor{red}{red} denotes question extraction, \textcolor{mybrown}{brown} represents key analysis, \textcolor{mygreen}{green} indicates answer summary, and \underline{\textit{underlined text}} highlights erroneous reasoning.

Although GPT-5 gets highest accuracy score, GPT-4.1 stands out as the top-performing proprietary model in our AgroCoT evaluations. On the open-source side, InternVL3-38B emerges as the strongest model in our assessment, achieving the highest overall accuracy among publicly accessible VLMs. By comparing these two systems, we aim to illustrate both the upper performance threshold of modern VLMs and the advancement of open-source alternatives. These case examples span the full range of AgroCoT’s task dimensions: each case includes a visual input (three views: satellite, aerial, or ground-level imagery), a task-aligned query, and the corresponding reasoning and answers generated by the models.

Our analysis of these examples reveals that GPT-4.1 exhibits more structured reasoning processes—though its outputs lean toward greater verbosity—paired with broader knowledge coverage, which contributes to higher answer accuracy (especially in tasks requiring complex reasoning or spatial interpretation). In contrast, InternVL3-38B delivers more concise, efficient reasoning and performs competently in visual recognition and straightforward classification tasks. However, it occasionally shows gaps in knowledge reserve and struggles with nuanced spatial relationships or logical inference, resulting in slightly lower overall answer accuracy. These distinctions reflect the varying levels of generalization capability and domain adaptability across contemporary VLMs.

To ensure that models effectively utilize visual information during reasoning, the AgroCoT dataset's reasoning chains explicitly integrate spatial localization mechanisms. As shown in Figure~\ref{fig:Organism Counting case}, each reasoning chain includes references to specific image regions in key steps, guiding the model to accurately locate relevant visual evidence when answering questions. This design not only enhances the interpretability of the reasoning process but also encourages the model to establish a tighter interaction between the visual and textual modalities, preventing vague reasoning disconnected from the image content. The example in Figure~\ref{fig:Organism Counting case} further demonstrates that, through this structured spatial guidance, the model can more accurately focus on the correct visual areas, thereby improving the reliability of the answer.

\section{Broader Impacts}
\label{app:broader_impacts}
AgroCoT aims to promote the development of vision-language models that can better understand agricultural scenes and perform domain-grounded reasoning. By providing high-quality agricultural VQA samples with chain-of-thought annotations, the benchmark can support research on crop recognition, pest and disease diagnosis, field-scene understanding, remote-sensing monitoring, and agricultural decision support. In particular, it may help lower the barrier for developing intelligent agricultural assistants and improve access to agricultural knowledge in scenarios where expert resources are limited.

At the same time, AgroCoT should be used with caution. The benchmark is designed for research and evaluation rather than direct deployment in high-stakes agricultural decision-making. Models evaluated on AgroCoT may still produce incorrect or overconfident predictions, especially for rare crops, unseen regions, complex diseases, or management decisions that require additional information such as soil condition, weather, season, pesticide history, and local agronomic practice. Misuse of such models may lead to inappropriate farming recommendations, economic loss, or environmental risks. Therefore, practical applications should involve expert verification, region-specific validation, and clear communication of model uncertainty. We hope AgroCoT can serve as a useful step toward more reliable, transparent, and agriculture-aware multimodal AI systems.

\section{Limitations and Future Work}
\label{app:limitations_and_future}
Although AgroCoT provides a multi-dimensional perspective for evaluating the reasoning capabilities of vision-language models in the agricultural domain, this study has certain limitations. Currently, we have not systematically explored the extent to which the introduction of Chain-of-Thought (CoT) reasoning directly improves the accuracy of model answers. Specifically, the causal contribution of CoT reasoning to the final task performance has yet to be quantified. Additionally, in the highly specialized field of agriculture, further research is needed to design more effective CoT guidance strategies that can better help vision-language models integrate domain knowledge and visual evidence, while addressing the reasoning challenges present in complex agricultural scenarios.

While AgroCoT has been carefully curated and manually refined to address agricultural VQA tasks, the current dataset scale remains insufficient given the rapid evolution of modern VLMs. As model capacities continue to grow, larger and more diverse training data become essential to fully exploit their reasoning potential. Therefore, AgroCoT will be progressively expanded by incorporating additional high-quality samples covering broader crop varieties, cultivation practices, environmental conditions, and region-specific scenarios. 

\section{Ethical Considerations}
\label{app:ethical_considerations}
The AgroCoT dataset is built upon publicly available open-source datasets, with the exception of a small amount of privately created data. All other data is sourced from published academic resources and is used in strict compliance with the corresponding dataset's copyright statements and licensing terms. We commit to releasing the AgroCoT dataset as an open-source project, aimed at advancing research in the field of agricultural artificial intelligence and promoting reproducible evaluation and model iteration.

The 20 annotators complete training covering agricultural knowledge, annotation protocols, and ethical norms. Participation is voluntary, participants’ right to information is guaranteed, and rest breaks are provided during annotation. All annotators receive specialized ethics training and take scheduled rest breaks to prevent contamination of the dataset during the refinement of CoTs. After annotation is completed, a dedicated ethics review of CoT content is conducted to identify and address any potential issues. 

\clearpage
\newpage
\section*{NeurIPS Paper Checklist}

\begin{enumerate}

\item {\bf Claims}
    \item[] Question: Do the main claims made in the abstract and introduction accurately reflect the paper's contributions and scope?
    \item[] Answer: \answerYes{} 
    \item[] Justification: The abstract and Section~\ref{sec:intro} and~\ref{sec:AgroCoT} state the contributions: a 4,759-sample CoT dataset covering five agricultural dimensions (OD, QA, DM, SU, EM) and evaluation of 30 VLMs. These are supported by results in Section~\ref{sec:eval}.

\item {\bf Limitations}
    \item[] Question: Does the paper discuss the limitations of the work performed by the authors?
    \item[] Answer: \answerYes{} 
    \item[] Justification: Appendix~\ref{app:limitations_and_future} discusses limitations including insufficient dataset scale and that the causal contribution of CoT to answer accuracy has not been quantified.

\item {\bf Theory assumptions and proofs}
    \item[] Question: For each theoretical result, does the paper provide the full set of assumptions and a complete (and correct) proof?
    \item[] Answer: \answerNA{} 
    \item[] Justification: The paper does not present theoretical results.

    \item {\bf Experimental result reproducibility}
    \item[] Question: Does the paper fully disclose all the information needed to reproduce the main experimental results of the paper to the extent that it affects the main claims and/or conclusions of the paper (regardless of whether the code and data are provided or not)?
    \item[] Answer: \answerYes{} 
    \item[] Justification: Section~\ref{sec:metric} describes the evaluation protocol. Appendix~\ref{app:experiment_details} provides model configurations and decoding parameters. The dataset URL is provided in the abstract.

\item {\bf Open access to data and code}
    \item[] Question: Does the paper provide open access to the data and code, with sufficient instructions to faithfully reproduce the main experimental results, as described in supplemental material?
    \item[] Answer: \answerYes{} 
    \item[] Justification: The abstract states the dataset is available at the provided Hugging Face URL. Appendix~\ref{app:ethical_considerations} commits to open-source release.

\item {\bf Experimental setting/details}
    \item[] Question: Does the paper specify all the training and test details (e.g., data splits, hyperparameters, how they were chosen, type of optimizer) necessary to understand the results?
    \item[] Answer: \answerYes{} 
    \item[] Justification: Section~\ref{subsec:setup} and Appendix~\ref{app:experiment_details} specify the evaluation metrics and model configurations.

\item {\bf Experiment statistical significance}
    \item[] Question: Does the paper report error bars suitably and correctly defined or other appropriate information about the statistical significance of the experiments?
    \item[] Answer: \answerYes{} 
    \item[] Justification: Table~\ref{tab:eval_results} and Table~\ref{tab:eval_results2} state in their captions that results are reported as mean (standard deviation) over three independent runs.

\item {\bf Experiments compute resources}
    \item[] Question: For each experiment, does the paper provide sufficient information on the computer resources (type of compute workers, memory, time of execution) needed to reproduce the experiments?
    \item[] Answer: \answerYes{} 
    \item[] Justification: Section~\ref{subsec:setup} and Appendix~\ref{app:experiment_details} specifies that open-source models were run on NVIDIA A800 GPUs and proprietary models via official APIs.
    
\item {\bf Code of ethics}
    \item[] Question: Does the research conducted in the paper conform, in every respect, with the NeurIPS Code of Ethics \url{https://neurips.cc/public/EthicsGuidelines}?
    \item[] Answer: \answerYes{} 
    \item[] Justification: Appendix~\ref{app:ethical_considerations} discusses dataset licensing, annotator training, voluntary participation, and ethics review.

\item {\bf Broader impacts}
    \item[] Question: Does the paper discuss both potential positive societal impacts and negative societal impacts of the work performed?
    \item[] Answer: \answerYes{} 
    \item[] Justification: Appendix~\ref{app:broader_impacts} discusses positive societal impacts and negative societal impacts.
    
\item {\bf Safeguards}
    \item[] Question: Does the paper describe safeguards that have been put in place for responsible release of data or models that have a high risk for misuse (e.g., pre-trained language models, image generators, or scraped datasets)?
    \item[] Answer: \answerYes{} 
    \item[] Justification: Appendix~\ref{app:broader_impacts} and~\ref{app:ethical_considerations} describes safeguards including annotator ethics training, scheduled rest breaks, and ethics review of CoT content.

\item {\bf Licenses for existing assets}
    \item[] Question: Are the creators or original owners of assets (e.g., code, data, models), used in the paper, properly credited and are the license and terms of use explicitly mentioned and properly respected?
    \item[] Answer: \answerYes{} 
    \item[] Justification: Appendix~\ref{app:data_collection} and~\ref{app:ethical_considerations} states that all data is used in compliance with the corresponding dataset's copyright statements and licensing terms.

\item {\bf New assets}
    \item[] Question: Are new assets introduced in the paper well documented and is the documentation provided alongside the assets?
    \item[] Answer: \answerYes{} 
    \item[] Justification: Section~\ref{subsec:construction} and Appendix~\ref{app:data_collection} document the dataset. The abstract provides the Hugging Face link.

\item {\bf Crowdsourcing and research with human subjects}
    \item[] Question: For crowdsourcing experiments and research with human subjects, does the paper include the full text of instructions given to participants and screenshots, if applicable, as well as details about compensation (if any)? 
    \item[] Answer: \answerYes{} 
    \item[] Justification: Appendix~\ref{app:human_refinement} and~\ref{app:human_evaluation} mentions annotators but does not provide compensation details.

\item {\bf Institutional review board (IRB) approvals or equivalent for research with human subjects}
    \item[] Question: Does the paper describe potential risks incurred by study participants, whether such risks were disclosed to the subjects, and whether Institutional Review Board (IRB) approvals (or an equivalent approval/review based on the requirements of your country or institution) were obtained?
    \item[] Answer: \answerNo{} 
    \item[] Justification: The paper does not mention IRB approval.

\item {\bf Declaration of LLM usage}
    \item[] Question: Does the paper describe the usage of LLMs if it is an important, original, or non-standard component of the core methods in this research? Note that if the LLM is used only for writing, editing, or formatting purposes and does \emph{not} impact the core methodology, scientific rigor, or originality of the research, declaration is not required.
    \item[] Answer: \answerYes{} 
    \item[] Justification: Section~\ref{subsec:construction} explicitly describe the use of GPT-4o for initial CoT generation.

\end{enumerate}

\end{document}